%% file: main.tex
\documentclass[10pt,twocolumn,letterpaper]{article}
\usepackage{cvpr}
\usepackage{times}
\usepackage{amsfonts}
\usepackage{amsmath}
\usepackage{amssymb}
\usepackage{array}
\usepackage{booktabs}
\usepackage{caption}
\usepackage{graphicx}
\usepackage{microtype}
\usepackage{nicefrac}
\usepackage{soul}
\usepackage{subcaption}
\usepackage{url}
\usepackage{xcolor}
\usepackage[pagebackref=true,breaklinks=true,colorlinks,bookmarks=false]{hyperref}
\usepackage[capitalise]{cleveref}

\newcommand{\cmt}[1]{\ignorespaces}

\makeatletter
\renewcommand{\paragraph}{%
 \@startsection{paragraph}{4}%
 {\z@}{0.5em}{-1em}%
 {\normalfont\normalsize\bfseries}%
}
\makeatother

\newif\ifarxiv
\arxivtrue

\cvprfinalcopy
\def\cvprPaperID{007}

\title{Unsupervised Learning of Probably Symmetric Deformable 3D Objects \\ from Images in the Wild}

\author{Shangzhe Wu \qquad
Christian Rupprecht \qquad
Andrea Vedaldi \vspace{0.5em} \\
Visual Geometry Group, University of Oxford\\
{\tt\small \{szwu, chrisr, vedaldi\}@robots.ox.ac.uk}}

\begin{document}
\twocolumn[\maketitle\vspace{-3em}\input{fig-teaser}\bigbreak]

\input{0_abstract}
\input{1_intro}
\input{2_related}
\input{3_method}

\input{4_experiments}
\input{5_conclusions}
\input{6_acknowledge}

{\small\bibliographystyle{ieee_fullname}\bibliography{shortstrings,refs}}

\newpage
\input{7_appendix}

\end{document}

%% file: fig-teaser.tex
\begin{center}
  \includegraphics[width=1\linewidth]{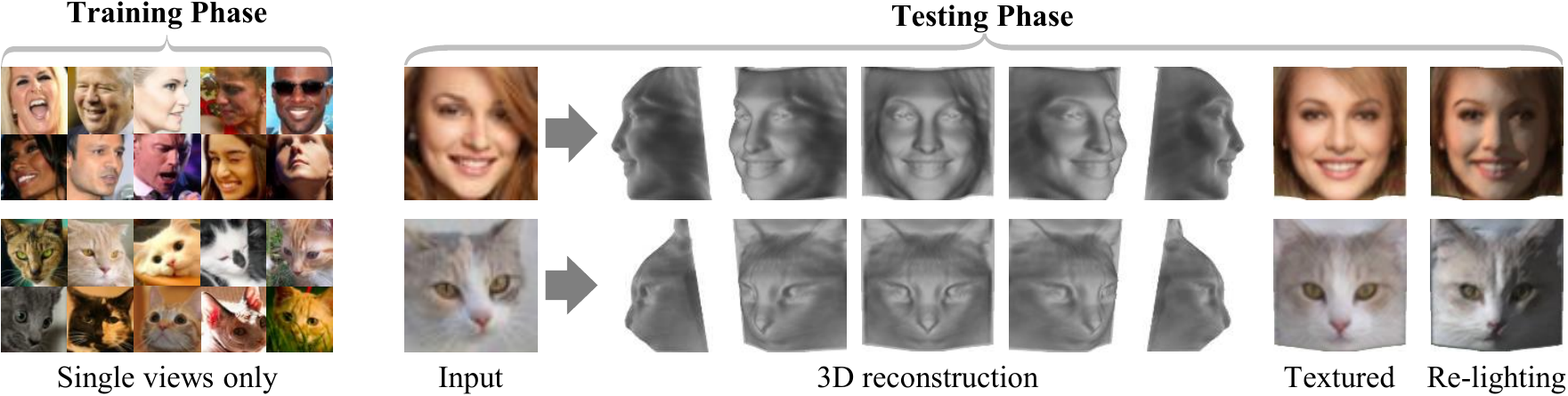}
\end{center}\vspace{-1.5em}
\captionof{figure}{\textbf{Unsupervised learning of 3D deformable objects from in-the-wild images.}
Left: Training uses \emph{only} single views of the object category with \emph{no} additional supervision at all (\ie no ground-truth 3D information, multiple views, or any prior model of the object).
Right: Once trained, our model reconstructs the 3D pose, shape, albedo and illumination of a deformable object instance from a single image with excellent fidelity. Code and demo at {\footnotesize\url{https://github.com/elliottwu/unsup3d}}.}
\label{fig:teaser}

%% file: 0_abstract.tex
\begin{abstract}
We propose a method to learn 3D deformable object categories from raw single-view images, without external supervision.
The method is based on an autoencoder that factors each input image into depth, albedo, viewpoint and illumination.
In order to disentangle these components without supervision, we use the fact that many object categories have, at least in principle, a symmetric structure.
We show that reasoning about illumination allows us to exploit the underlying object symmetry even if the appearance is not symmetric due to shading.
Furthermore, we model objects that are probably, but not certainly, symmetric by predicting a symmetry probability map, learned end-to-end with the other components of the model.
Our experiments show that this method can recover very accurately the 3D shape of human faces, cat faces and cars from single-view images, without any supervision or a prior shape model.
On benchmarks, we demonstrate superior accuracy compared to another method that uses supervision at the level of 2D image correspondences. 
\end{abstract}

%% file: 1_intro.tex
\section{Introduction}\label{s:intro}

Understanding the 3D structure of images is key in many computer vision applications.
Futhermore, while many deep networks appear to understand images as 2D textures~\cite{geirhos19imagenet-trained}, 3D modelling can explain away much of the variability of natural images and potentially improve image understanding in general.
Motivated by these facts, we consider the problem of learning 3D models for deformable object categories.

We study this problem under two challenging conditions.
The \emph{first} condition is that no 2D or 3D ground truth information (such as keypoints, segmentation, depth maps, or prior knowledge of a 3D model) is available.
Learning without external supervisions removes the bottleneck of collecting image annotations, which is often a major obstacle to deploying deep learning for new applications.
The \emph{second} condition is that the algorithm must use an unconstrained collection of single-view images --- in particular, it should not require multiple views of the same instance.
Learning from single-view images is useful because in many applications, and especially for deformable objects, we solely have a source of still images to work with.
Consequently, our learning algorithm ingests a number of single-view images of a deformable object category and produces as output a deep network that can estimate the 3D shape of any instance given a single image of it (\cref{fig:teaser}).

We formulate this as an autoencoder that internally decomposes the image into albedo, depth, illumination and viewpoint, \emph{without direct supervision for any of these factors}.
However, without further assumptions, decomposing images into these four factors is ill-posed.
In search of minimal assumptions to achieve this, we note that many object categories are \emph{symmetric} (\eg almost all animals and many handcrafted objects).
Assuming an object is perfectly symmetric, one can obtain a virtual second view of it by simply mirroring the image.
In fact, if correspondences between the pair of mirrored images were available, 3D reconstruction could be achieved by stereo reconstruction~\cite{Mukherjee94,francois03mirrorsymmetry,thrun05symmetry,sinha12symmetry,gao17exploiting}.
Motivated by this, we seek to leverage symmetry as a geometric cue to constrain the decomposition.

However, specific object instances are in practice never fully symmetric, neither in shape nor appearance.
Shape is non-symmetric due to variations in pose or other details (\eg hair style or expressions on a human face), and albedo can also be non-symmetric (\eg asymmetric texture of cat faces).
Even when both shape and albedo are symmetric, the appearance may still not be, due to asymmetric illumination.

We address this issue in two ways.
First, we explicitly model illumination to exploit the underlying symmetry and show that, by doing so, the model can exploit illumination as an additional cue for recovering the shape.
Second, we augment the model to reason about potential lack of symmetry in the objects.
To do this, the model predicts, along with the other factors, a dense map containing the probability that a given pixel has a symmetric counterpart in the image.

We combine these elements in an end-to-end learning formulation, where all components, including the confidence maps, are learned from raw RGB data only.
We also show that symmetry can be enforced by flipping internal representations, which is particularly useful for reasoning about symmetries probabilistically.

We demonstrate our method on several datasets, including human faces, cat faces and cars.
We provide a thorough ablation study using a synthetic face dataset to obtain the necessary 3D ground truth.
On real images, we achieve higher fidelity reconstruction results compared to other methods~\cite{sahasrabudhe19lifting,Szabo19} that do not rely on 2D or 3D ground truth information, nor prior knowledge of a 3D model of the instance or class.
In addition, we also outperform a recent state-of-the-art method~\cite{Moniz2018} that uses keypoint supervision for 3D reconstruction on real faces, while our method uses no external supervision at all.
Finally, we demonstrate that our trained face model generalizes to non-natural images such as face paintings and cartoon drawings without fine-tuning.

%% file: 2_related.tex
\section{Related Work}\label{s:related_work}

\input{tab-related}

In order to assess our contribution in relation to the vast literature on image-based 3D reconstruction, it is important to consider three aspects of each approach:
which information is used, which assumptions are made, and what the output is.
Below and in~\cref{t:related} we compare our contribution to prior works based on these factors.

Our method uses single-view images of an object category as training data, assumes that the objects belong to a specific class (\eg human faces) which is weakly symmetric, and outputs a monocular predictor capable of decomposing any image of the category into shape, albedo, illumination, viewpoint and symmetry probability.

\paragraph{Structure from Motion.}
Traditional methods such as Structure from Motion (SfM)~\cite{Faugeras01geometry} can reconstruct the 3D structure of individual rigid scenes given as input multiple views of each scene and 2D keypoint matches between the views.
This can be extended in two ways.
First, \emph{monocular reconstruction methods} can perform dense 3D reconstruction from a single image without 2D keypoints~\cite{zhou2017unsupervised,ummenhofer2017demon,monodepth17}.
However, they require multiple views~\cite{monodepth17} or videos of rigid scenes for training~\cite{zhou2017unsupervised}.
Second, \emph{Non-Rigid SfM} (NRSfM) approaches~\cite{bregler00recovering,novotny19c3dpo} can learn to reconstruct deformable objects by allowing 3D points to deform in a limited manner between views, but require supervision in terms of annotated 2D keypoints for both training and testing.
Hence, neither family of SfM approaches can learn to reconstruct deformable objects from raw pixels of a single view.

\paragraph{Shape from X.}
Many other monocular cues have been used as alternatives or supplements to SfM for recovering shape from images, such as shading~\cite{horn89shape, Zhang99Sfshading}, silhouettes~\cite{Koenderink84Sfsilhouette}, texture~\cite{Witkin81Sftexture}, symmetry~\cite{Mukherjee94,francois03mirrorsymmetry} etc.
In particular, our work is inspired from \emph{shape from symmetry} and \emph{shape from shading}.
Shape from symmetry~\cite{Mukherjee94,francois03mirrorsymmetry,thrun05symmetry,sinha12symmetry} reconstructs symmetric objects from a single image by using the mirrored image as a virtual second view, provided that symmetric correspondences are available. {}\cite{sinha12symmetry} also shows that it is possible to detect symmetries and correspondences using descriptors.
Shape from shading~\cite{horn89shape, Zhang99Sfshading} assumes a shading model such as Lambertian reflectance, and reconstructs the surface by exploiting the non-uniform illumination.

\paragraph{Category-specific reconstruction.}
Learning-based methods have recently been leveraged to reconstruct objects from a single view, either in the form of a raw image or 2D keypoints (see also~\cref{t:related}).
While this task is ill-posed, it has been shown to be solvable by learning a suitable object prior from the training data~\cite{bfm09, wu16learning, achlioptas17learning, ranjan18generating}.
A variety of supervisory signals have been proposed to learn such priors.
Besides using 3D ground truth directly, authors have considered using videos~\cite{Agrawal15, zhou2017unsupervised, Novotny17b, Wang_2018_CVPR} and stereo pairs~\cite{monodepth17, Luo2018SVS}.
Other approaches have used single views with 2D keypoint annotations~\cite{Kanazawa18cmr, Moniz2018, Suwajanakorn2018, Chen19pose} or object masks~\cite{Kanazawa18cmr,DIBR19}.
For objects such as human bodies and human faces, some methods~\cite{kanazawa18end-to-end,gerig18morphable,wang2019adversarial,gecer19ganfit} have learn to reconstruct from raw images, but starting from the knowledge of a predefined shape model such as SMPL~\cite{loper2015smpl} or Basel~\cite{bfm09}.
These prior models are constructed using specialized hardware and/or other forms of supervision, which are often difficult to obtain for deformable objects in the wild, such as animals, and also limited in details of the shape.

Only recently have authors attempted to learn the geometry of object categories from raw, monocular views \emph{only}.
Thewlis~\etal~\cite{Thewlis17b,Thewlis18} uses equivariance to learn dense landmarks, which recovers the 2D geometry of the objects.
DAE~\cite{Shu2018} learns to predict a deformation field through heavily constraining an autoencoder with a small bottleneck embedding and lift that to 3D in~\cite{sahasrabudhe19lifting} --- in post processing, they further decompose the reconstruction in albedo and shading, obtaining an output similar to ours.

Adversarial learning has been proposed as a way of hallucinating new views of an object.
Some of these methods start from 3D representations~\cite{wu16learning,achlioptas17learning,zhu2018von,ranjan18generating}. Kato~\etal~\cite{kato2019vpl} trains a discriminator on raw images but uses viewpoint as addition supervision.
HoloGAN~\cite{nguyen2019hologan} only uses raw images but does not obtain an explicit 3D reconstruction.
Szabo~\etal~\cite{Szabo19} uses adversarial training to reconstruct 3D meshes of the object, but does not assess their results quantitatively.
Henzler~\etal~\cite{henzler19escaping} also learns from raw images, but only experiments with images that contain the object on a white background, which is akin to supervision with 2D silhouettes.
In \cref{s:exp_sota}, we compare to \cite{sahasrabudhe19lifting,Szabo19} and demonstrate superior reconstruction results with much higher fidelity.

Since our model generates images from an internal 3D representation, one essential component is a differentiable renderer.
However, with a traditional rendering pipeline, gradients across occlusions and boundaries are not defined.
Several soft relaxations have thus been proposed~\cite{Loper2014, kato2018renderer, Liu2018softras}.
Here, we use an implementation\footnote{\scriptsize \url{https://github.com/daniilidis-group/neural_renderer}} of~\cite{kato2018renderer}.

%% file: tab-related.tex
\begin{table}
\setlength{\belowcaptionskip}{-5pt}
  \footnotesize
  \setlength{\tabcolsep}{2pt}
  \begin{tabular}{clll}
  \toprule
  Paper                          & Supervision   & Goals                           & Data \\
  \midrule
  {}\cite{bfm09}     & 3D scans      & 3DMM                            & Face \\
  {}\cite{wu16learning}          & 3DV, I        & Prior on 3DV, predict from I    & ShapeNet, Ikea \\
  {}\cite{achlioptas17learning}  & 3DP           & Prior on 3DP                    & ShapeNet \\
  {}\cite{ranjan18generating}    & 3DM           & Prior on 3DM                    & Face \\
  \midrule
  {}\cite{geng193d-guided}       & 3DMM, 2DKP, I & Refine 3DMM fit to I            & Face \\
  {}\cite{gecer19ganfit}         & 3DMM, 2DKP, I & Fit 3DMM to I+2DKP              & Face \\
  {}\cite{gerig18morphable}      & 3DMM          & Fit 3DMM to 3D scans            & Face \\
  {}\cite{kanazawa18end-to-end}  & 3DMM, 2DKP    & Pred.~3DMM from I               & Humans \\
  {}\cite{Sengupta18sfsnet}      & 3DMM, 2DS+KP  & Pred.~N, A, L from I                & Face \\
  {}\cite{wang2019adversarial}   & 3DMM, I       & Pred.~3DM, VP, T, E~from I      & Face \\
  {}\cite{sanyal19learning}      & 3DMM, 2DKP, I & Fit 3DMM to I                   & Face \\
  \midrule
  {}\cite{gadelha163d-shape}     & 2DS           & Prior on 3DV, pred.~from I      & Model/ScanNet \\
  {}\cite{kato2019vpl}           & I, 2DS, VP    & Prior on 3DV                    & ScanNet, PAS3D \\
  {}\cite{Kanazawa18cmr}         & I, 2DS+KP     & Pred.~3DM, T, VP from I         & Birds \\
  {}\cite{DIBR19}                & I, 2DS        & Pred.~3DM, T, L, VP from I      & ShapeNet, Birds \\
  {}\cite{henzler19escaping}     & I, 2DS        & Pred.~3DV, VP from I            & ShapeNet, others \\
  \midrule
  \midrule
  {}\cite{Szabo19}               & I             & Prior on 3DM, T, I                  & Face \\
  {}\cite{sahasrabudhe19lifting} & I             & Pred.~3DM, VP, T$^\dagger$ from I & Face \\
  {}\cite{henderson19learning}   & I             & Pred.~V, L, VP from I           & ShapeNet \\
  Ours                           & I             & Pred.~D, L, A, VP from I        & Face, others \\
  \bottomrule
  \end{tabular}
  \caption{Comparison with selected prior work: supervision, goals, and data.
  I\@: image,
  3DMM\@: 3D morphable model,
  2DKP\@: 2D keypoints,
  2DS\@: 2D silhouette,
  3DP\@: 3D points,
  VP\@: viewpoint,
  E\@: expression,
  3DM\@: 3D mesh,
  3DV\@: 3D volume,
  D\@: depth,
  N\@: normals,
  A\@: albedo,
  T\@: texture,
  L\@: light.
  {}$^\dagger$ can also recover A and L in post-processing.
  }\label{t:related}
  \end{table}

%% file: 3_method.tex
\section{Method}\label{s:method}

\input{fig-model}

Given an unconstrained collection of images of an object category, such as human faces, our goal is to learn a model $\Phi$ that receives as input an image of an object instance and produces as output a decomposition of it into 3D shape, albedo, illumination and viewpoint, as illustrated in~\cref{fig:pipeline}.

As we have only raw images to learn from, the learning objective is reconstructive: namely, the model is trained so that the combination of the four factors gives back the input image.
This results in an autoencoding pipeline where the factors have, due to the way they are recomposed, an explicit photo-geometric meaning.

In order to learn such a decomposition without supervision for any of the components, we use the fact that many object categories are \emph{bilaterally symmetric}.
However, the appearance of object instances is never perfectly symmetric.
Asymmetries arise from shape deformation, asymmetric albedo and asymmetric illumination.
We take two measures to account for these asymmetries.
First, we explicitly model asymmetric illumination.
Second, our model also estimates, for each pixel in the input image, a confidence score that explains the probability of the pixel having a symmetric counterpart in the image (see \textbf{conf} $\sigma, \sigma'$ in \cref{fig:pipeline}).

The following sections describe how this is done, looking first at the photo-geometric autoencoder (\cref{s:decomposition}), then at how symmetries are modelled (\cref{s:prob}), followed by details of the image formation (\cref{s:formation}) and the supplementary perceptual loss (\cref{s:perc_loss}).

\subsection{Photo-geometric autoencoding}\label{s:decomposition}

An image $\mathbf{I}$ is a function $\Omega\rightarrow\mathbb{R}^3$ defined on a grid $\Omega = \{0, \ldots, W-1\} \times \{0, \ldots, H-1\}$, or, equivalently, a tensor in $\mathbb{R}^{3\times W\times H}$.
We assume that the image is roughly centered on an instance of the object of interest.
The goal is to learn a function $\Phi$, implemented as a neural network, that maps the image $\mathbf{I}$ to four factors $(d, a, w, l)$ comprising a \emph{depth map} $d : \Omega \rightarrow \mathbb{R}_+$, an \emph{albedo image} $a : \Omega \rightarrow \mathbb{R}^3$, a global \emph{light direction} $l \in \mathbb{S}^2$, and a \emph{viewpoint} $w \in \mathbb{R}^6$ so that the image can be reconstructed from them.

The image $\mathbf{I}$ is reconstructed from the four factors in two steps, \emph{lighting} $\Lambda$ and \emph{reprojection} $\Pi$, as follows:
\begin{equation}\label{e:generator}
\hat{\mathbf{I}} = \Pi\left(\Lambda(a, d, l), d, w\right).
\end{equation}
The lighting function $\Lambda$ generates a version of the object based on the depth map $d$, the light direction $l$ and the albedo $a$ as seen from a canonical viewpoint $w=0$.
The viewpoint $w$ represents the transformation between the canonical view and the viewpoint of the actual input image $\mathbf{I}$.
Then, the reprojection function $\Pi$ simulates the effect of a viewpoint change and generates the image $\hat{\mathbf{I}}$ given the canonical depth $d$ and the shaded canonical image $\Lambda(a, d, l)$.
Learning uses a reconstruction loss which encourages $\mathbf{I} \approx \hat{\mathbf{I}}$ (\cref{s:prob}).

\paragraph{Discussion.}

The effect of lighting could be incorporated in the albedo $a$ by interpreting the latter as a texture rather than as the object's albedo.
However, there are two good reasons to avoid this.
First, the albedo $a$ is often symmetric even if the illumination causes the corresponding appearance to look asymmetric.
Separating them allows us to more effectively incorporate the symmetry constraint described below.
Second, shading provides an additional cue on the underlying 3D shape~\cite{horn75obtaining, belhumeur99thebasrelief}.
In particular, unlike the recent work of~\cite{Shu2018} where a shading map is predicted independently from shape, our model computes the shading based on the predicted depth, mutually constraining each other.

\subsection{Probably symmetric objects}\label{s:prob}

Leveraging symmetry for 3D reconstruction requires identifying symmetric object points in an image.
Here we do so implicitly, assuming that depth and albedo, which are reconstructed in a canonical frame, are symmetric about a fixed vertical plane.
An important beneficial side effect of this choice is that it helps the model discover a `canonical view' for the object, which is important for reconstruction~\cite{novotny19c3dpo}.

To do this, we consider the operator that flips a map $a\in\mathbb{R}^{C\times W\times H}$ along the horizontal axis\footnote{The choice of axis is arbitrary as long as it is fixed.}:
$
  [\operatorname{flip} a]_{c,u,v} = a_{c,W-1-u,v}.
$
We then require $d \approx \operatorname{flip} d'$ and $a \approx \operatorname{flip} a'$.
While these constraints could be enforced by adding corresponding loss terms to the learning objective, they would be difficult to balance.
Instead, we achieve the same effect indirectly, by obtaining a second reconstruction $\hat{\mathbf{I}}'$ from the flipped depth and albedo:
\begin{equation}\label{e:generator-flipped}
  \mathbf{\hat{I}}' = \Pi\left(\Lambda(a', d', l), d', w\right),
  ~~
  a' =  \operatorname{flip} a,
  ~~
  d' = \operatorname{flip} d.
\end{equation}

Then, we consider two reconstruction losses encouraging $\mathbf{I} \approx \hat{\mathbf{I}}$ and $\mathbf{I} \approx \mathbf{\hat{I}}'$.
Since the two losses are commensurate, they are easy to balance and train jointly.
Most importantly, this approach allows us to easily reason about symmetry probabilistically, as explained next.

The source image $\mathbf{I}$ and the reconstruction $\hat{\mathbf{I}}$ are compared via the loss:
\begin{equation}\label{e:loss-l1}
\mathcal{L} (\hat{\mathbf{I}}, \mathbf{I}, \sigma) =
- \frac{1}{|\Omega|}
  \sum_{uv\in\Omega}
  \ln \frac{1} {\sqrt{2}\sigma_{uv}}
  \exp{-\frac{\sqrt{2} \ell_{1,uv}} {\sigma_{uv}}},
\end{equation}
where $\ell_{1,uv}=|\hat{\mathbf{I}}_{uv} - \mathbf{I}_{uv}|$ is the $L_1$ distance between the intensity of pixels at location $uv$, and $\sigma \in \mathbb{R}^{W\times H}_+$ is a \emph{confidence map}, also estimated by the network $\Phi$ from the image $\mathbf{I}$, which expresses the \emph{aleatoric uncertainty} of the model.
The loss can be interpreted as the negative log-likelihood of a factorized Laplacian distribution on the reconstruction residuals.
Optimizing likelihood causes the model to self-calibrate, learning a meaningful confidence map~\cite{kendall2017uncertainties}.

Modelling uncertainty is generally useful, but in our case is particularly important when we consider the ``symmetric'' reconstruction $\hat{\mathbf{I}}'$,
for which we use the same loss $\mathcal{L}(\hat{\mathbf{I}}', \mathbf{I}, \sigma')$.
Crucially, we use the network to estimate, also from the same input image $\mathbf{I}$, a \emph{second} confidence map $\sigma'$.
This confidence map allows the model to learn which portions of the input image might \emph{not} be symmetric.
For instance, in some cases hair on a human face is not symmetric as shown in~\cref{fig:pipeline}, and $\sigma'$ can assign a higher reconstruction uncertainty to the hair region where the symmetry assumption is not satisfied.
Note that this depends on the \emph{specific} instance under consideration, and is learned by the model itself.

Overall, the learning objective is given by the combination of the two reconstruction errors:
\begin{equation}\label{e:totalconf}
\mathcal{E}(\Phi;\mathbf{I}) =
\mathcal{L}(\hat{\mathbf{I}}, \mathbf{I}, \sigma) +
\lambda_\text{f} \mathcal{L}(\hat{\mathbf{I}}', \mathbf{I}, \sigma'), 
\end{equation}
where $\lambda_\text{f} = 0.5$ is a weighing factor, $(d, a, w, l, \sigma, \sigma') = \Phi(\mathbf{I})$ is the output of the neural network,
 and $\hat{\mathbf{I}}$ and $\hat{\mathbf{I}}'$ are obtained according to~\cref{e:generator,e:generator-flipped}.

\subsection{Image formation model}\label{s:formation}

We now describe the functions $\Pi$ and $\Lambda$ in~\cref{e:generator} in more detail.
The image is formed by a camera looking at a 3D object.
If we denote with $P = (P_x,P_y,P_z)\in\mathbb{R}^3$ a 3D point expressed in the reference frame of the camera, this is mapped to pixel
$p = (u,v,1)$ by the following projection:
\begin{equation}\label{e:camera}
p \propto K P,~~~
K =
\begin{bmatrix}
f & 0 & c_u \\
0 & f & c_v \\
0 & 0 & 1 \\
\end{bmatrix},~~~
\begin{cases}
c_u=\frac{W-1}{2},\\
c_v=\frac{H-1}{2},\\
f = \frac{W-1}{2\tan\frac{\theta_{\text{FOV}}}{2}}.\\
\end{cases}
\end{equation}
This model assumes a perspective camera with \emph{field of view} (FOV) $\theta_\text{FOV}$. We assume a nominal distance of the object from the camera at about $1\mathrm{m}$.
Given that the images are cropped around a particular object, we assume a relatively narrow FOV of $\theta_{\text{FOV}} \approx 10^\circ$.

The depth map $d : \Omega\rightarrow\mathbb{R}_+$ associates a depth value $d_{uv}$  to each pixel $(u,v) \in \Omega$ in the canonical view.
By inverting the camera model~\eqref{e:camera}, we find that this corresponds to the 3D point
$
 P = d_{uv} \cdot K^{-1} p.
$

The viewpoint $w\in\mathbb{R}^6$ represents an Euclidean transformation $(R,T)\in SE(3)$, where $w_{1:3}$ and $w_{4:6}$ are rotation angles and translations along $x$, $y$ and $z$ axes respectively.

The map $(R,T)$ transforms 3D points from the canonical view to the actual view.
Thus a pixel $(u,v)$ in the canonical view is mapped to the pixel $(u', v')$ in the actual view by the warping function $\eta_{d,w}:  (u,v) \mapsto (u',v')$ given by:
\begin{equation}\label{e:forward}
p'  \propto K (d_{uv} \cdot R K^{-1} p + T),
\end{equation}
where
$
p' = (u',v',1).
$

Finally, the reprojection function $\Pi$ takes as input the depth $d$ and the viewpoint change $w$ and applies the resulting warp to the canonical image  $\mathbf{J}$ to obtain the actual image $\mathbf{\hat{I}} = \Pi(\mathbf{J},d,w)$ as
$
\mathbf{\hat{I}}_{u'v'} = \mathbf{J}_{uv},
$
where
$
(u, v) = \eta_{d,w}^{-1}(u', v').
$%
\footnote{Note that this requires to compute the \emph{inverse} of the warp $\eta_{d,w}$, which is detailed \ifarxiv in~\cref{sec:render}. \else in the supplementary material. \fi}

The canonical image $\mathbf{J} = \Lambda(a, d, l)$ is in turn generated as a combination of albedo, normal map and light direction.
To do so, given the depth map $d$, we derive the normal map
$n : \Omega \rightarrow \mathbb{S}^2$ by associating to each pixel $(u,v)$ a vector normal to the underlying 3D surface.
In order to find this vector, we compute the vectors  $t^u_{uv}$ and $ t^v_{uv}$ tangent to the surface along the $u$ and $v$ directions.
For example, the first one is:
$
t^u_{uv} =
d_{u+1,v}\cdot
K^{-1}
(p + e_x)
-d_{u-1,v}
\cdot
K^{-1}
(p - e_x)
$
where $p$ is defined above and $e_x = (1,0,0)$.
Then the normal is obtained by taking the vector product $n_{uv} \propto t^u_{uv} \times t^v_{uv}$.

The normal $n_{uv}$ is multiplied by the light direction $l$ to obtain a value for the directional illumination and the latter is added to the ambient light.
Finally, the result is multiplied by the albedo to obtain the illuminated texture, as follows:
$
\mathbf{J}_{uv} =
 \left(k_s + k_d \max \{ 0, \langle l , n_{uv} \rangle \} \right) \cdot a_{uv}.
$
Here $k_s$ and $k_d$ are the scalar coefficients weighting the ambient and diffuse terms, and are predicted by the model with range between 0 and 1 via rescaling a {\tt tanh} output. The light direction $l = (l_x, l_y, 1)^T/(l_x^2 + l_y^2 + 1)^{0.5}$ is modeled as a spherical sector by predicting $l_x$ and $l_y$ with {\tt tanh}.

\subsection{Perceptual loss}\label{s:perc_loss}

The $L_1$ loss function~\cref{e:loss-l1} is sensitive to small geometric imperfections and tends to result in blurry reconstructions.
We add a \emph{perceptual loss} term to mitigate this problem.
The $k$-th layer of an off-the-shelf image encoder $e$ (VGG16 in our case~\cite{Simonyan15}) predicts a representation $e^{(k)}(\mathbf{I}) \in \mathbb{R}^{C_k\times W_k\times H_k}$ where $\Omega_k = \{0,\dots,W_k-1\}\times \{0,\dots,H_k-1\}$ is the corresponding spatial domain.
Note that this feature encoder does not have to be trained with supervised tasks. Self-supervised encoders can be equally effective as shown in~\cref{tab:ablation}.

Similar to~\cref{e:loss-l1}, assuming a Gaussian distribution, the perceptual loss is given by:
{\small
\begin{equation}\label{e:lpercep}
  \mathcal{L}_\text{p}^{(k)}(\hat{\mathbf{I}}, \mathbf{I}, \sigma^{(k)}) =
  -\frac{1}{|\Omega_k|} \sum_{uv \in \Omega_k}
  \ln \frac{1} {\sqrt{2 \pi (\sigma^{(k)}_{uv})^2 }}
\exp
-
\frac
{(\ell^{(k)}_{uv})^2}
{2 (\sigma^{(k)}_{uv})^2},
\end{equation}}
where $\ell^{(k)}_{uv}=|e^{(k)}_{uv}(\hat{\mathbf{I}}) - e^{(k)}_{uv}(\mathbf{I})|$ for each pixel index $uv$ in the $k$-th layer.
We also compute the loss for $\hat{\mathbf{I}}'$ using $\sigma^{(k)'}$.
$\sigma^{(k)}$ and $\sigma^{(k)'}$ are additional confidence maps predicted by our model.
In practice, we found it is good enough for our purpose to use the features from only one layer {\tt relu3\_3} of VGG16.
We therefore shorten the notation of perceptual loss to $\mathcal{L}_\text{p}$.
With this, the loss function $\mathcal{L}$ in~\cref{e:totalconf} is replaced by $\mathcal{L} + \lambda_\text{p} \mathcal{L}_\text{p}$ with $\lambda_\text{p}=1$.

%% file: fig-model.tex
\begin{figure}[t]
\setlength{\belowcaptionskip}{-5pt}
  \includegraphics[width=\linewidth]{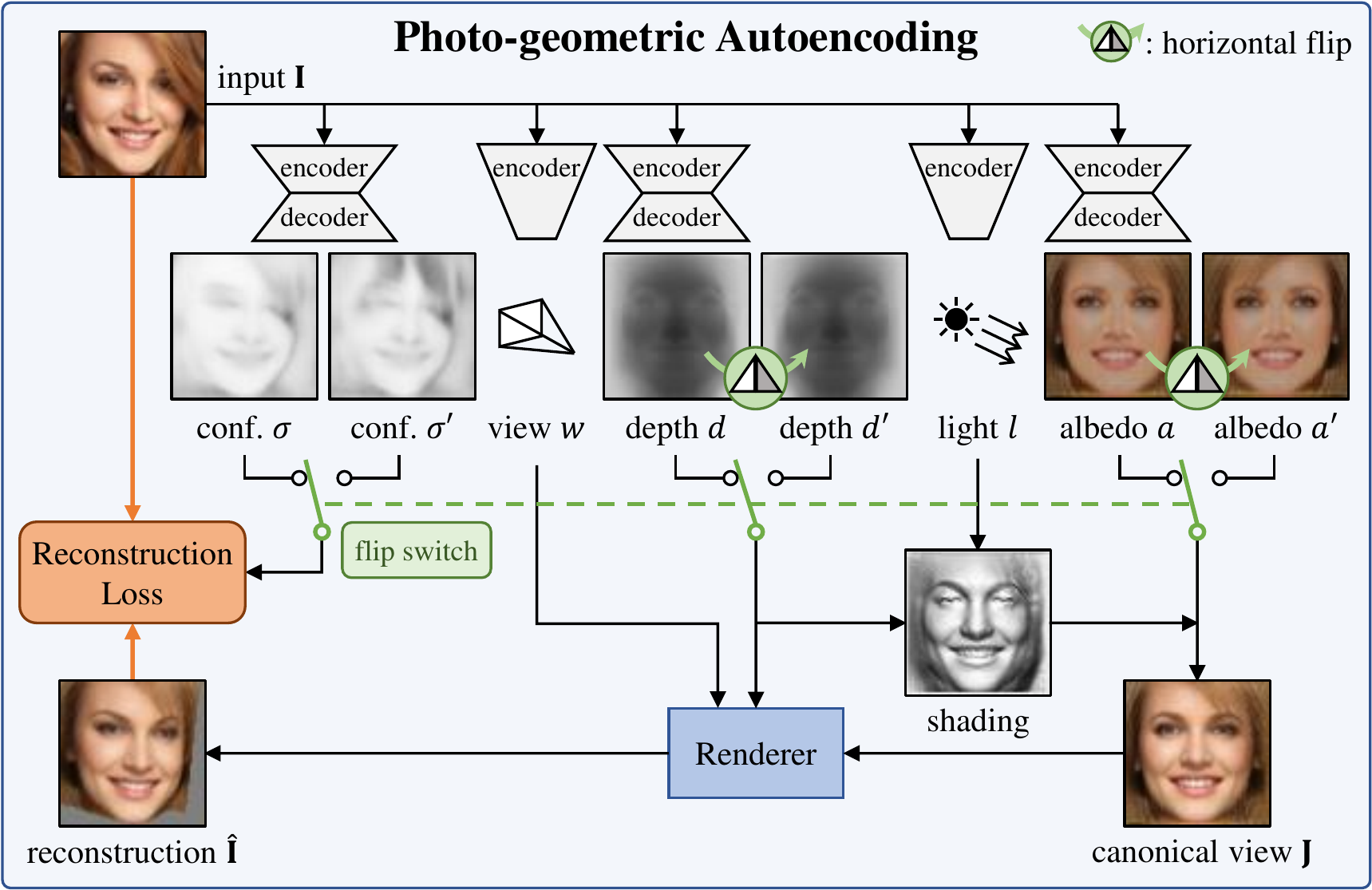}
\caption{\textbf{Photo-geometric autoencoding}. Our network $\Phi$ decomposes an input image $\mathbf{I}$ into depth, albedo, viewpoint and lighting, together with a pair of confidence maps. It is trained to reconstruct the input without external supervision.}\label{fig:pipeline}
\end{figure}

%% file: 4_experiments.tex
\section{Experiments}\label{s:exp}

\subsection{Setup}

\paragraph{Datasets.}

We test our method on three human face datasets: \textbf{CelebA}~\cite{liu2015celeba}, \textbf{3DFAW}~\cite{multipie2010,Jeni2015,ZHANG2014692,Yin2008}
and \textbf{BFM}~\cite{bfm09}.
CelebA is a large scale human face dataset, consisting of over $200$k images of real human faces in the wild annotated with bounding boxes.
3DFAW contains $23$k images with $66$ 3D keypoint annotations, which we use to evaluate our 3D predictions in \cref{s:exp_sota}.
We roughly crop the images around the head region and use the official train/val/test splits.
BFM (Basel Face Model) is a synthetic face model, which we use to assess the quality of the 3D reconstructions (since the in-the-wild datasets lack ground-truth).
We follow the protocol of~\cite{Sengupta18sfsnet} to generate a dataset, sampling shapes, poses, textures, and illumination randomly.
We use images from SUN Database~\cite{Xiao2010SUN} as background and save ground truth depth maps for evaluation.

We also test our method on cat faces and synthetic cars. We use two \textbf{cat datasets}~\cite{zhang2008cat, parkhi12a}.
The first one has $10$k cat images with nine keypoint annotations, and the second one is a collection of dog and cat images, containing $1.2$k cat images with bounding box annotations.
We combine the two datasets and crop the images around the cat heads.
For cars, we render $35$k images of synthetic cars from \textbf{ShapeNet}~\cite{shapenet2015} with random viewpoints and illumination. We randomly split the images by $8$:$1$:$1$ into train, validation and test sets.

\paragraph{Metrics.}

Since the scale of 3D reconstruction from projective cameras is inherently ambiguous~\cite{Faugeras01geometry}, we discount it in the evaluation.
Specifically, given the depth map $d$ predicted by our model in the canonical view, we warp it to a depth map $\bar d$ in the actual view using the predicted viewpoint and compare the latter to the ground-truth depth map $d^*$ using the \textbf{scale-invariant depth error} (SIDE)~\cite{Eigen2014}
$
E_\text{SIDE}(\bar d, d^*) =
(
    \frac{1}{WH} \sum_{uv}\Delta_{uv}^2 -
    (
        \frac{1}{WH}\sum_{uv}\Delta_{uv}
    )^2
)^{\frac{1}{2}}
$
where $\Delta_{uv} = \log \bar d_{uv} - \log d^*_{uv}$. 
We compare only valid depth pixel and erode the foreground mask by one pixel to discount rendering artefacts at object boundaries.
Additionally, we report the \textbf{mean angle deviation} (MAD) between normals computed from ground truth depth and from the predicted depth, measuring how well the surface is captured.

\paragraph{Implementation details.}

The function $(d, a, w, l, \sigma) = \Phi(\mathbf{I})$ that preditcs depth, albedo, viewpoint, lighting, and confidence maps from the image $\mathbf{I}$ is implemented using individual neural networks.
The depth and albedo are generated by encoder-decoder networks, while viewpoint and lighting are regressed using simple encoder networks.
The encoder-decoders do not use skip connections because input and output images are \emph{not} spatially aligned (since the output is in the canonical viewpoint).
All four confidence maps are predicted using the same network, at different decoding layers for the photometric and perceptual losses since these are computed at different resolutions.
The final activation function is {\tt tanh} for depth, albedo, viewpoint and lighting and {\tt softplus} for the confidence maps.
The depth prediction is centered on the mean before {\tt tanh}, as the global distance is estimated as part of the viewpoint.
We do \emph{not} use any special initialization for all predictions, except that two border pixels of the depth maps on both the left and the right are clamped at a maximal depth to avoid boundary issues.

We train using Adam over batches of $64$ input images, resized to $64\times 64$ pixels.
The size of the output depth and albedo is also $64\times 64$. We train for approximately $50$k iterations. For visualization, depth maps are upsampled to $256$.
We include more details \ifarxiv in~\cref{s:train_detail}. \else in the supplementary material. \fi

\subsection{Results}

\input{tab-baseline}
\input{tab-ablation}
\input{tab-asym_perturb}
\input{tab-3dfaw-kpt}

\paragraph{Comparison with baselines.}

\Cref{tab:baseline} uses the BFM dataset to compare the depth reconstruction quality obtained by our method, a fully-supervised baseline and two baselines.
The supervised baseline is a version of our model trained to regress the ground-truth depth maps using an $L_1$ loss.
The trivial baseline predicts a constant uniform depth map, which provides a performance lower-bound.
The third baseline is a constant depth map obtained by averaging all ground-truth depth maps in the test set.
Our method largely outperforms the two constant baselines and approaches the results of supervised training.
Improving over the third baseline (which has access to GT information) confirms that the model learns an \textit{instance specific} 3D representation.

\paragraph{Ablation.}\label{s:ablation}

To understand the influence of the individual parts of the model, we remove them one at a time and evaluate the performance of the ablated model in~\cref{tab:ablation}. Visual results are reported \ifarxiv in~\cref{fig:ablation}. \else in the supplementary material. \fi

In the table, \textbf{row (1)} shows the performance of the full model (the same as in~\cref{tab:baseline}).
\textbf{Row (2)} does not flip the albedo.
Thus, the albedo is not encouraged to be symmetric in the canonical space, which fails to canonicalize the viewpoint of the object and to use cues from symmetry to recover shape.
The performance is as low as the trivial baseline in~\cref{tab:baseline}.
\textbf{Row (3)} does not flip the depth, with a similar effect to row (2).
\textbf{Row (4)} predicts a shading map instead of computing it from depth and light direction.
This also harms performance significantly because shading cannot be used as a cue to recover shape.
\textbf{Row (5)} switches off the perceptual loss, which leads to degraded image quality and hence degraded reconstruction results.
\textbf{Row (6)} replaces the ImageNet pretrained image encoder used in the perceptual loss with one\footnote{We use a RotNet~\cite{gidaris18rotnet} pretrained VGG16 model obtained from \url{https://github.com/facebookresearch/DeeperCluster}.} trained through a self-supervised task~\cite{gidaris18rotnet}, which shows no difference in performance.
Finally, \textbf{row (7)} switches off the confidence maps, using a fixed and uniform value for the confidence --- this reduces losses~\eqref{e:loss-l1} and~\eqref{e:lpercep} to the basic $L_1$ and $L_2$ losses, respectively.
The accuracy does not drop significantly, as faces in BFM are highly symmetric (\eg do not have hair), but its variance increases.
To better understand the effect of the confidence maps, we specifically evaluate on partially asymmetric faces using perturbations.

\paragraph{Asymmetric perturbation.}\label{s:asym_perturb}

\input{fig-asym_perturb}

In order to demonstrate that our uncertainty modelling allows the model to handle asymmetry, we add asymmetric perturbations to BFM\@.
Specifically, we generate random rectangular color patches with $20\%$ to $50\%$ of the image size and blend them onto the images with $\alpha$-values ranging from $0.5$ to $1$, as shown in~\cref{fig:asym_perturb}.
We then train our model with and without confidence on these perturbed images, and report the results in~\cref{tab:asym_perturb}.
Without the confidence maps, the model always predicts a symmetric albedo and geometry reconstruction often fails.
With our confidence estimates, the model is able to reconstruct the asymmetric faces correctly, with very little loss in accuracy compared to the unperturbed case.

\paragraph{Qualitative results.}

\input{fig-face_cat_car}
\input{fig-painting}
\input{fig-symline}

In~\cref{fig:face_cat_car} we show reconstruction results of human faces from CelebA and 3DFAW, cat faces from \cite{zhang2008cat, parkhi12a} and synthetic cars from ShapeNet.
The 3D shapes are recovered with high fidelity. The reconstructed 3D face, for instance, contain fine details of the nose, eyes and mouth even in the presence of extreme facial expression.

To further test generalization, we applied our model trained on the CelebA dataset to a number of paintings and cartoon drawings of faces collected from~\cite{Crowley15} and the Internet.
As shown in \cref{fig:painting}, our method still works well even though it has never seen such images during training.

\paragraph{Symmetry and asymmetry detection.}

Since our model predicts a canonical view of the objects that is symmetric about the vertical center-line of the image, we can easily visualize the symmetry plane, which is otherwise non-trivial to detect from in-the-wild images.
In~\cref{fig:symline}, we warp the center-line of the canonical image to the predicted input viewpoint.
Our method can detect symmetry planes accurately despite the presence of asymmetric texture and lighting effects.
We also overlay the predicted confidence map $\sigma'$ onto the image, confirming that the model assigns low confidence to asymmetric regions in a sample-specific way.

\subsection{Comparison with the state of the art}\label{s:exp_sota}

As shown in~\cref{t:related}, most reconstruction methods in the literature require either image annotations, prior 3D models or both.
When these assumptions are dropped, the task becomes considerably harder, and there is little prior work that is directly comparable.
Of these, {}\cite{henderson19learning} only uses synthetic, texture-less objects from ShapeNet, {}\cite{Szabo19} reconstructs in-the-wild faces but does not report any quantitative results, and {}\cite{sahasrabudhe19lifting} reports quantitative results only on keypoint regression, but not on the 3D reconstruction quality.
We were not able to obtain code or trained models from~\cite{sahasrabudhe19lifting,Szabo19} for a direct quantitative comparison and thus compare qualitatively.

\paragraph{Qualitative comparison.}

\input{fig-compare_paper}

In order to establish a side-by-side comparison, we cropped the examples reported in the papers~\cite{sahasrabudhe19lifting, Szabo19} and compare our results with theirs (\cref{fig:compare_paper}).
Our method produces much higher quality reconstructions than both methods, with fine details of the facial expression, whereas \cite{sahasrabudhe19lifting} recovers 3D shapes poorly and \cite{Szabo19} generates unnatural shapes.
Note that \cite{Szabo19} uses an unconditional GAN that generates high resolution 3D faces from random noise, and cannot recover 3D shapes from images.
The input images for~\cite{Szabo19} in~\cref{fig:compare_paper} were generated by their GAN.

\paragraph{3D keypoint depth evaluation.}

Next, we compare to the DepthNet model of~\cite{Moniz2018}.
This method predicts depth for selected facial keypoints, but uses 2D keypoint annotations as input --- a much easier setup than the one we consider here.
Still, we compare the quality of the reconstruction of these sparse point obtained by DepthNet and our method.
We also compare to the baselines MOFA~\cite{tewari17MoFA} and AIGN~\cite{tung2017} reported in~\cite{Moniz2018}.
For a fair comparison, we use their public code which computes the depth correlation score (between $0$ and $66$) on the frontal faces.
We use the 2D keypoint locations to sample our predicted depth and then evaluate the same metric.
The set of test images from 3DFAW and the preprocessing are identical to~\cite{Moniz2018}.
Since 3DFAW is a small dataset with limited variation, we also report results with CelebA pre-training.

In~\cref{tab:3dfaw_kpt} we report the results from their paper and the slightly improved results we obtained from their publicly-available implementation.
The paper also evaluates a supervised model using a GAN discriminator trained with ground-truth depth information.
While our method does not use any supervision, it still outperforms DepthNet and reaches close-to-supervised performance.

\subsection{Limitations}\label{s:failure}

\input{fig-failure.tex}

While our method is robust in many challenging scenarios (e.g., extreme facial expression, abstract drawing), we do observe failure cases as shown in~\cref{fig:failure}.
During training, we assume a simple Lambertian shading model, ignoring shadows and specularity, which leads to inaccurate reconstructions under extreme lighting conditions (\cref{fig:failure_light}) or highly non-Lambertian surfaces.
Disentangling noisy dark textures and shading (\cref{fig:failure_texture}) is often difficult.
The reconstruction quality is lower for extreme poses (\cref{fig:failure_pose}), partly due to poor supervisory signal from the reconstruction loss of side images. This may be improved by imposing constraints from accurate reconstructions of frontal poses.

%% file: tab-baseline.tex
\begin{table}[t]
\newcommand{\xpm}[1]{{\tiny$\pm#1$}}
\setlength{\tabcolsep}{5pt}
\centering\footnotesize
\begin{tabular}{clccc}
\toprule
  No  &   Baseline                & SIDE ($\times10^{-2}$) $\downarrow$ & MAD (deg.) $\downarrow$  \\ \midrule
  (1) &   Supervised              & $0.410$ \xpm{0.103}                 & $10.78$ \xpm{1.01} \\ \midrule
  (2) &   Const.\ null depth      & $2.723$ \xpm{0.371}                 & $43.34$ \xpm{2.25} \\
  (3) &   Average g.t.\ depth     & $1.990$ \xpm{0.556}                 & $23.26$ \xpm{2.85} \\ \midrule
  (4) &   Ours (unsupervised)     & $0.793$ \xpm{0.140}                 & $16.51$ \xpm{1.56} \\
\bottomrule
\end{tabular}
\setlength{\belowcaptionskip}{-5pt}
\caption{\textbf{Comparison with baselines.} SIDE and MAD errors of our reconstructions on the BFM dataset compared against a fully-supervised and trivial baselines.}\label{tab:baseline}
\end{table}

%% file: tab-ablation.tex
\begin{table}[t]
\footnotesize\newcommand{\xpm}[1]{{\tiny$\pm#1$}}
\setlength{\tabcolsep}{5pt}
\centering\footnotesize
\begin{tabular}{clcc}
\toprule
  No  & Method                         & SIDE~($\times10^{-2}$) $\downarrow$ & MAD~(deg.) $\downarrow$ \\\midrule
  (1) & Ours full                      & $0.793$ \xpm{0.140}    & $16.51$ \xpm{1.56} \\ \midrule
  (2) & w/o albedo flip                & $2.916$ \xpm{0.300}    & $39.04$ \xpm{1.80} \\
  (3) & w/o depth flip                 & $1.139$ \xpm{0.244}    & $27.06$ \xpm{2.33} \\
  (4) & w/o light                      & $2.406$ \xpm{0.676}    & $41.64$ \xpm{8.48} \\
  (5) & w/o perc.~loss                 & $0.931$ \xpm{0.269}    & $17.90$ \xpm{2.31} \\
  (6) & w/ self-sup. perc.~loss        & $0.815$ \xpm{0.145}    & $15.88$ \xpm{1.57} \\
  (7) & w/o confidence                 & $0.829$ \xpm{0.213}    & $16.39$ \xpm{2.12} \\
\bottomrule
\end{tabular}
\setlength{\belowcaptionskip}{-5pt}
\caption{\textbf{Ablation study.} Refer to~\cref{s:ablation} for details.}\label{tab:ablation}
\end{table}

%% file: tab-asym_perturb.tex
\begin{table}[t]
\newcommand{\xpm}[1]{{\tiny$\pm#1$}}
\setlength{\tabcolsep}{5pt}
\centering\footnotesize
\begin{tabular}{lccc}
\toprule
                      & SIDE ($\times10^{-2}$) $\downarrow$ & MAD (deg.) $\downarrow$ \\ \midrule
  No perturb, no conf. & $0.829$ \xpm{0.213}                & $16.39$ \xpm{2.12} \\
  No perturb, conf.    & $0.793$ \xpm{0.140}                & $16.51$ \xpm{1.56} \\ \midrule
  Perturb, no conf.    & $2.141$ \xpm{0.842}                & $26.61$ \xpm{5.39} \\
  Perturb, conf.       & $0.878$ \xpm{0.169}                & $17.14$ \xpm{1.90} \\
\bottomrule
\end{tabular}
\setlength{\belowcaptionskip}{-5pt}
\caption{\textbf{Asymmetric perturbation.}
We add asymmetric perturbations to BFM and show that confidence maps allow the model to reject such noise, while the vanilla model without confidence maps breaks.}\label{tab:asym_perturb}
\end{table}

%% file: tab-3dfaw-kpt.tex
\begin{table}[t]
\newcommand{\xpm}[1]{{\tiny$\pm#1$}}
\setlength{\tabcolsep}{1pt}
\centering\footnotesize
\begin{tabular}{lr}
\toprule
                                                                          & Depth Corr. $\uparrow$ \\ \midrule
Ground truth                                                              & $66$ \\
AIGN~\cite{tung2017} (\textbf{supervised}, from \cite{Moniz2018})         & $50.81$ \\
DepthNetGAN~\cite{Moniz2018} (\textbf{supervised}, from \cite{Moniz2018}) & $58.68$ \\ \midrule \midrule
MOFA~\cite{tewari17MoFA} (\textbf{model-based}, from \cite{Moniz2018})    & $15.97$ \\
DepthNet~\cite{Moniz2018} (from \cite{Moniz2018})                         & $26.32$ \\
DepthNet~\cite{Moniz2018} (from GitHub)                                   & $35.77$ \\ \midrule
Ours                                                                      & $48.98$ \\
Ours (w/ CelebA pre-training)                                             & $54.65$ \\
\bottomrule
\end{tabular}
\setlength{\belowcaptionskip}{-5pt}
\caption{\textbf{3DFAW keypoint depth evaluation.} Depth correlation between ground truth and prediction evaluated at $66$ facial keypoint locations.}\label{tab:3dfaw_kpt}
\end{table}

%% file: fig-asym_perturb.tex
\begin{figure}[t]
\captionsetup[subfigure]{labelformat=empty,aboveskip=0pt}
\begin{subfigure}[b]{\linewidth}\centering
  \caption{perturbed dataset}
  \includegraphics[height=0.7cm]{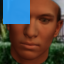}%
  \includegraphics[height=0.7cm]{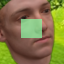}%
  \includegraphics[height=0.7cm]{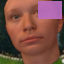}%
  \includegraphics[height=0.7cm]{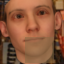}%
  \includegraphics[height=0.7cm]{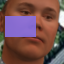}%
  \includegraphics[height=0.7cm]{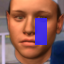}%
  \includegraphics[height=0.7cm]{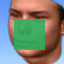}%
  \includegraphics[height=0.7cm]{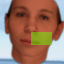}%
  \includegraphics[height=0.7cm]{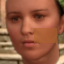}%
  \includegraphics[height=0.7cm]{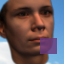}%
  \includegraphics[height=0.7cm]{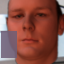}%
  \includegraphics[height=0.7cm]{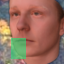}
\end{subfigure}
\\
\begin{subfigure}[b]{.105\linewidth}\centering
  \includegraphics[height=0.85cm]{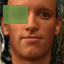}
\end{subfigure}\hfill
\begin{subfigure}[b]{.31\linewidth}\centering
  \includegraphics[height=0.85cm]{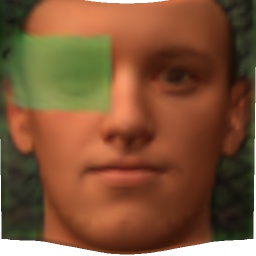}
  \includegraphics[height=0.85cm]{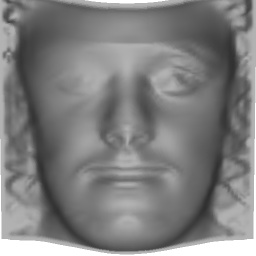}
  \includegraphics[trim={10px 0 40px 0}, clip, height=0.85cm]{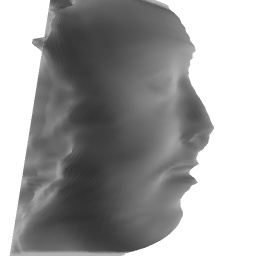}
\end{subfigure}
\begin{subfigure}[b]{.105\linewidth}\centering
  \caption{conf $\sigma$}
  \includegraphics[height=0.85cm]{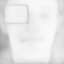}
\end{subfigure}
\begin{subfigure}[b]{.11\linewidth}\centering
  \caption{conf $\sigma'$}
  \includegraphics[height=0.85cm]{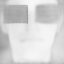}
\end{subfigure}\hfill
\begin{subfigure}[b]{.305\linewidth}\centering
  \includegraphics[height=0.85cm]{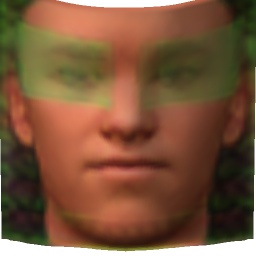}
  \includegraphics[height=0.85cm]{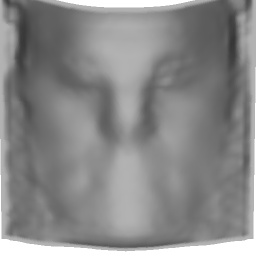}
  \includegraphics[trim={10px 0 52px 0}, clip, height=0.85cm]{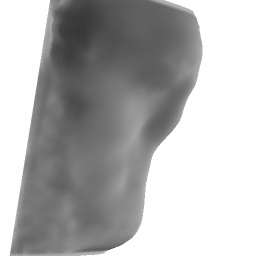}
\end{subfigure}

\vspace{-0.3em}
\setlength{\tabcolsep}{0pt}
\small
\begin{tabular}{>{\centering}p{0.9cm}>{\centering}p{5cm}>{\centering}p{2.4cm}}
input & recon w/ conf & recon w/o conf
\end{tabular}

\vspace{-0.5em}
\setlength{\belowcaptionskip}{-5pt}
\caption{\textbf{Asymmetric perturbation.}
Top: examples of the perturbed dataset. Bottom: reconstructions with and without confidence maps.
Confidence allows the model to correctly reconstruct the 3D shape with the asymmetric texture.
}\label{fig:asym_perturb}
\end{figure}

%% file: fig-face_cat_car.tex
\begin{figure}[t]
\captionsetup[subfigure]{justification=centering,labelformat=empty,aboveskip=2pt}

\begin{subfigure}[b]{.15\linewidth}\centering
  \includegraphics[height=1.15cm]{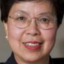}
\end{subfigure}\hfill
\begin{subfigure}[b]{.85\linewidth}\centering
  \includegraphics[trim={0 0 40px 0}, clip, height=1.15cm]{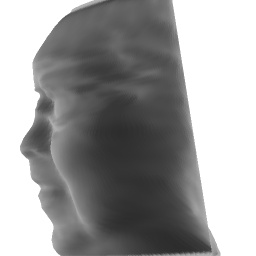}
  \includegraphics[trim={24px 0 0 0}, clip, height=1.15cm]{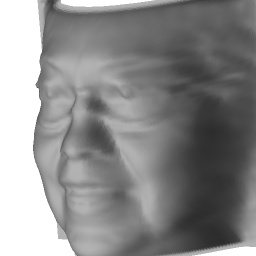}
  \includegraphics[height=1.15cm]{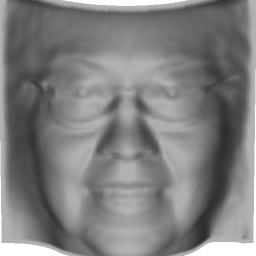}
  \includegraphics[height=1.15cm]{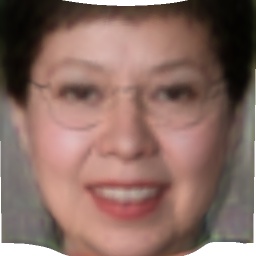}
  \includegraphics[trim={0 0 24px 0}, clip, height=1.15cm]{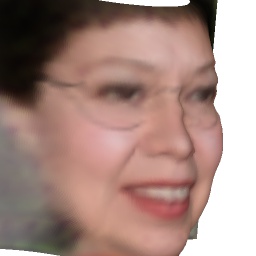}
  \includegraphics[trim={40px 0 0 0}, clip, height=1.15cm]{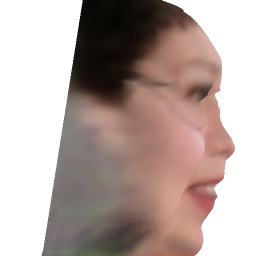}
\end{subfigure}
\\
\begin{subfigure}[b]{.15\linewidth}\centering
  \includegraphics[height=1.15cm]{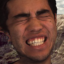}
\end{subfigure}\hfill
\begin{subfigure}[b]{.85\linewidth}\centering
  \includegraphics[trim={0 0 40px 0}, clip, height=1.15cm]{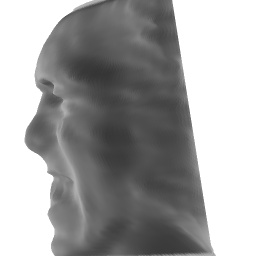}
  \includegraphics[trim={24px 0 0 0}, clip, height=1.15cm]{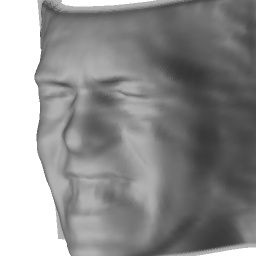}
  \includegraphics[height=1.15cm]{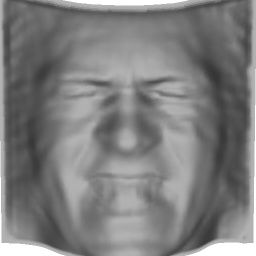}
  \includegraphics[height=1.15cm]{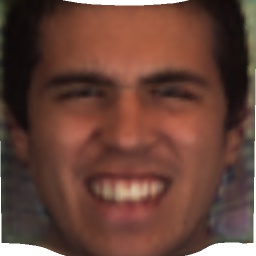}
  \includegraphics[trim={0 0 24px 0}, clip, height=1.15cm]{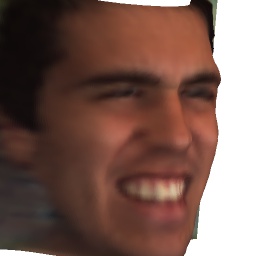}
  \includegraphics[trim={40px 0 0 0}, clip, height=1.15cm]{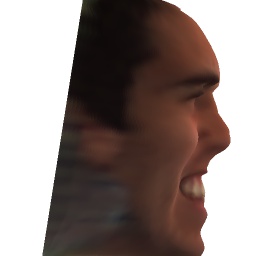}
\end{subfigure}
\\
\begin{subfigure}[b]{.15\linewidth}\centering
  \includegraphics[height=1.15cm]{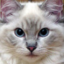}
\end{subfigure}\hfill
\begin{subfigure}[b]{.85\linewidth}\centering
  \includegraphics[trim={0 0 40px 0}, clip, height=1.15cm]{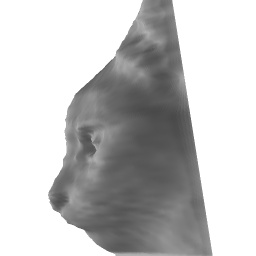}
  \includegraphics[trim={24px 0 0 0}, clip, height=1.15cm]{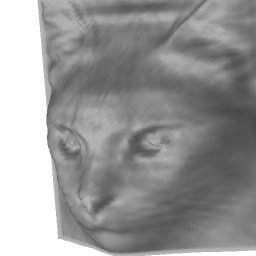}
  \includegraphics[height=1.15cm]{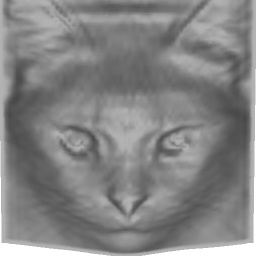}
  \includegraphics[height=1.15cm]{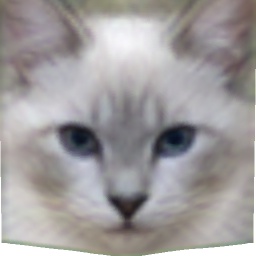}
  \includegraphics[trim={0 0 24px 0}, clip, height=1.15cm]{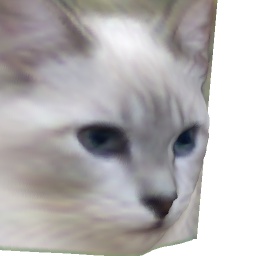}
  \includegraphics[trim={40px 0 0 0}, clip, height=1.15cm]{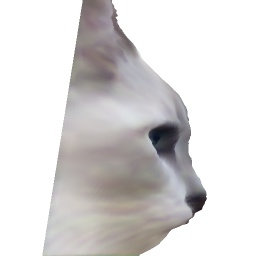}
\end{subfigure}
\\
\begin{subfigure}[b]{.15\linewidth}\centering
  \includegraphics[height=1.15cm]{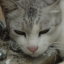}
\end{subfigure}\hfill
\begin{subfigure}[b]{.85\linewidth}\centering
  \includegraphics[trim={0 0 40px 0}, clip, height=1.15cm]{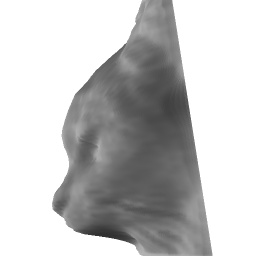}
  \includegraphics[trim={24px 0 0 0}, clip, height=1.15cm]{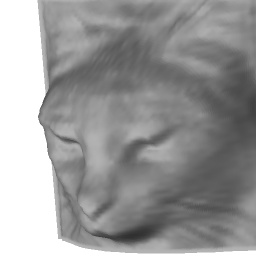}
  \includegraphics[height=1.15cm]{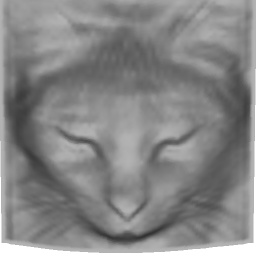}
  \includegraphics[height=1.15cm]{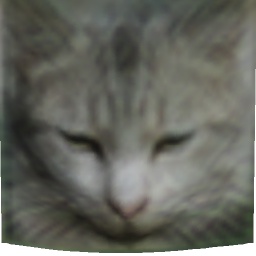}
  \includegraphics[trim={0 0 24px 0}, clip, height=1.15cm]{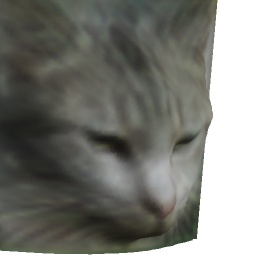}
  \includegraphics[trim={40px 0 0 0}, clip, height=1.15cm]{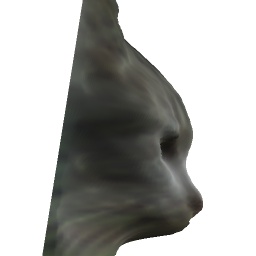}
\end{subfigure}
\\
\begin{subfigure}[b]{.15\linewidth}\centering
  \includegraphics[height=1.15cm]{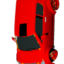}
  \caption{input}
\end{subfigure}\hfill
\begin{subfigure}[b]{.85\linewidth}\centering
  \includegraphics[trim={0 0 40px 0}, clip, height=1.15cm]{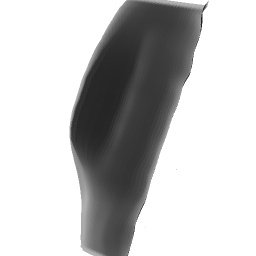}
  \includegraphics[trim={24px 0 0 0}, clip, height=1.15cm]{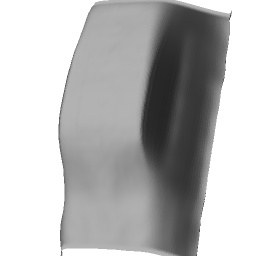}
  \includegraphics[height=1.15cm]{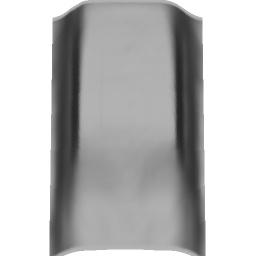}
  \includegraphics[height=1.15cm]{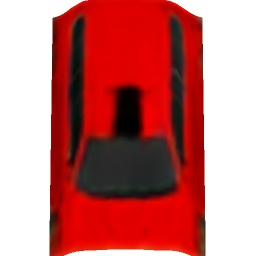}
  \includegraphics[trim={0 0 24px 0}, clip, height=1.15cm]{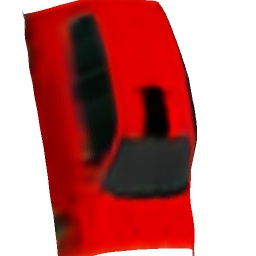}
  \includegraphics[trim={40px 0 0 0}, clip, height=1.15cm]{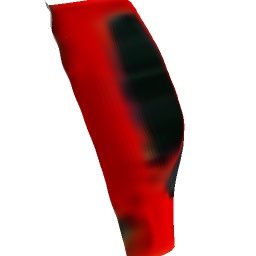}
  \caption{reconstruction}
\end{subfigure}

\vspace{-0.5em}
\setlength{\belowcaptionskip}{-5pt}
\caption{\textbf{Reconstruction of faces, cats and cars.}}\label{fig:face_cat_car}
\end{figure}

%% file: fig-painting.tex
\begin{figure}[t]
\captionsetup[subfigure]{justification=centering,labelformat=empty,aboveskip=2pt}

\begin{subfigure}[b]{.15\linewidth}\centering
    \includegraphics[height=1.15cm]{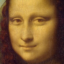}
\end{subfigure}\hfill
\begin{subfigure}[b]{.85\linewidth}\centering
    \includegraphics[trim={0 0 36px 0}, clip, height=1.15cm]{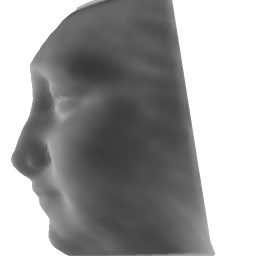}
    \includegraphics[trim={24px 0 0 0}, clip, height=1.15cm]{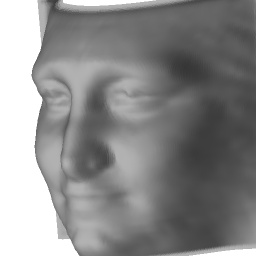}
    \includegraphics[height=1.15cm]{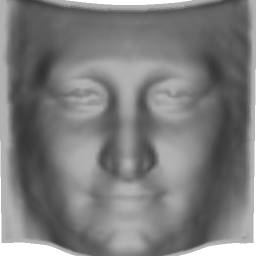}
    \includegraphics[height=1.15cm]{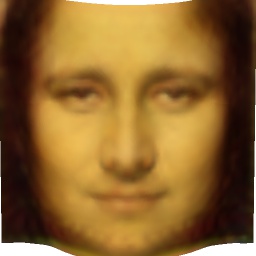}
    \includegraphics[trim={0 0 24px 0}, clip, height=1.15cm]{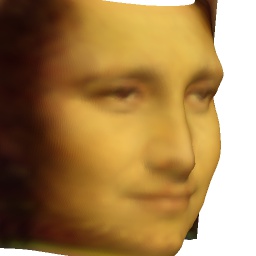}
    \includegraphics[trim={36px 0 0 0}, clip, height=1.15cm]{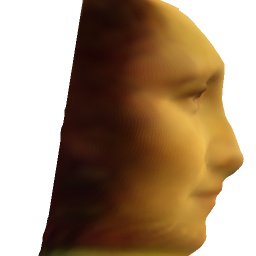}
\end{subfigure}
\\
\begin{subfigure}[b]{.15\linewidth}\centering
    \includegraphics[height=1.15cm]{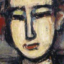}
    \caption{input}
\end{subfigure}\hfill
\begin{subfigure}[b]{.85\linewidth}\centering
    \includegraphics[trim={0 0 36px 0}, clip, height=1.15cm]{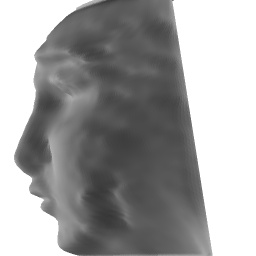}
    \includegraphics[trim={24px 0 0 0}, clip, height=1.15cm]{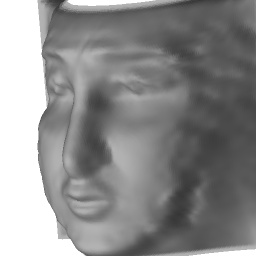}
    \includegraphics[height=1.15cm]{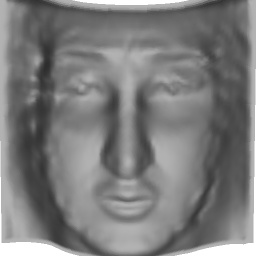}
    \includegraphics[height=1.15cm]{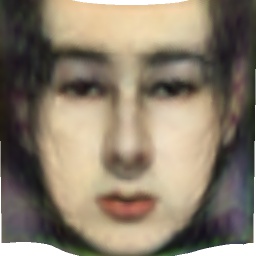}
    \includegraphics[trim={0 0 24px 0}, clip, height=1.15cm]{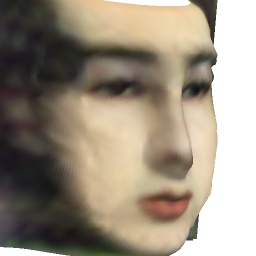}
    \includegraphics[trim={36px 0 0 0}, clip, height=1.15cm]{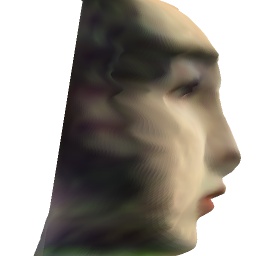}
    \caption{reconstruction}
\end{subfigure}

\vspace{-0.5em}
\setlength{\belowcaptionskip}{-5pt}
\caption{\textbf{Reconstruction of faces in paintings.}}\label{fig:painting}
\end{figure}

%% file: fig-symline.tex
\begin{figure}[t]
\captionsetup[subfigure]{justification=centering,aboveskip=2pt}
\begin{subfigure}[b]{.5\linewidth}\centering
  \includegraphics[height=0.9cm]{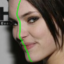}
  \includegraphics[height=0.9cm]{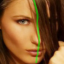}
  \includegraphics[height=0.9cm]{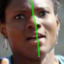}
  \includegraphics[height=0.9cm]{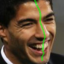}
  \includegraphics[height=0.9cm]{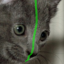}
  \includegraphics[height=0.9cm]{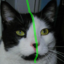}
  \includegraphics[height=0.9cm]{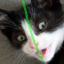}
  \includegraphics[height=0.9cm]{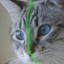}
  \caption{symmetry plane}
\end{subfigure}\hfill
\begin{subfigure}[b]{.5\linewidth}\centering
  \includegraphics[height=0.9cm]{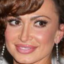}
  \includegraphics[height=0.9cm]{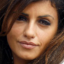}
  \includegraphics[height=0.9cm]{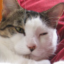}
  \includegraphics[height=0.9cm]{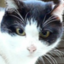}
  \includegraphics[height=0.9cm]{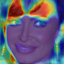}
  \includegraphics[height=0.9cm]{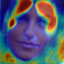}
  \includegraphics[height=0.9cm]{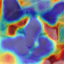}
  \includegraphics[height=0.9cm]{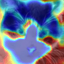}
  \caption{asymmetry visualization}
\end{subfigure}

\vspace{-0.5em}
\setlength{\belowcaptionskip}{-5pt}
\caption{\textbf{Symmetry plane and asymmetry detection.} (a): our model can reconstruct the ``intrinsic'' symmetry plane of an in-the-wild object even though the appearance is highly asymmetric. (b): asymmetries (highlighted in red) are detected and visualized using confidence map $\sigma'$.}\label{fig:symline}
\end{figure}

%% file: fig-compare_paper.tex
\begin{figure}[t]
\captionsetup[subfigure]{justification=centering,labelformat=empty,aboveskip=2pt}
\begin{subfigure}[b]{.13\linewidth}\centering
  \includegraphics[height=1.1cm]{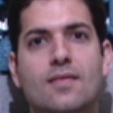}
  \includegraphics[height=1.1cm]{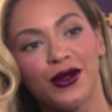}
  \caption{input}
\end{subfigure}\hfill
\begin{subfigure}[b]{.25\linewidth}\centering
  \includegraphics[height=1.1cm]{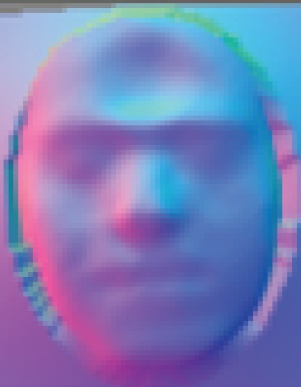}
  \includegraphics[height=1.1cm]{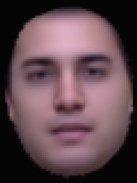}
  \includegraphics[height=1.1cm]{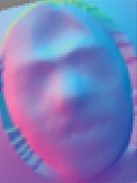}
  \includegraphics[height=1.1cm]{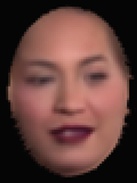}
  \caption{LAE~\cite{sahasrabudhe19lifting}}
\end{subfigure}\hfill
\begin{subfigure}[b]{.47\linewidth}\centering
  \includegraphics[height=1.1cm]{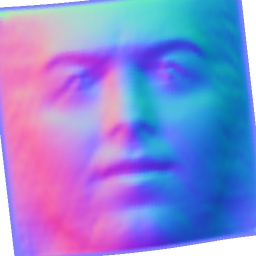}
  \includegraphics[height=1.1cm]{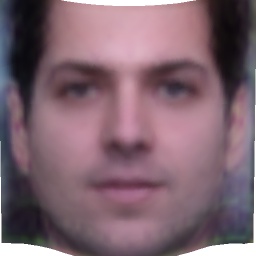}
  \includegraphics[trim={0 0 10mm 0}, clip, height=1.1cm]{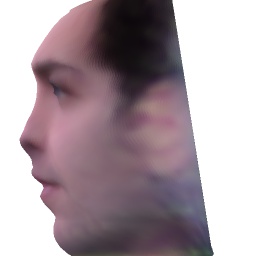}
  \includegraphics[height=1.1cm]{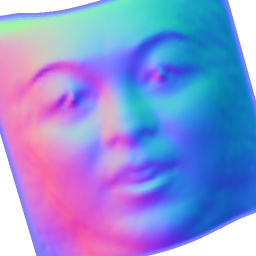}
  \includegraphics[height=1.1cm]{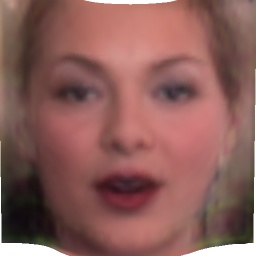}
  \includegraphics[trim={0 0 10mm 0}, clip, height=1.1cm]{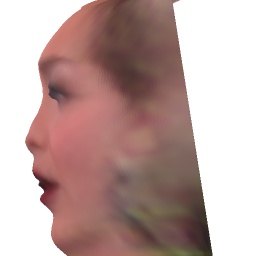}
  \caption{ours}
\end{subfigure}
\\
\begin{subfigure}[b]{.13\linewidth}\centering
  \includegraphics[height=1.1cm]{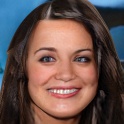}
  \includegraphics[height=1.1cm]{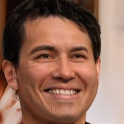}
  \caption{input}
\end{subfigure}\hfill
\begin{subfigure}[b]{.40\linewidth}\centering
  \includegraphics[height=1.1cm]{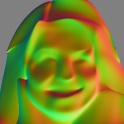}
  \includegraphics[height=1.1cm]{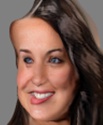}
  \includegraphics[height=1.1cm]{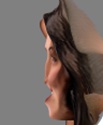}
  \includegraphics[height=1.1cm]{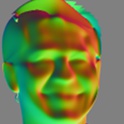}
  \includegraphics[height=1.1cm]{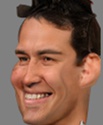}
  \includegraphics[height=1.1cm]{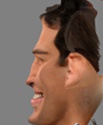}
  \caption{Szab\'{o} \etal~\cite{Szabo19}}
\end{subfigure}\hfill
\begin{subfigure}[b]{.45\linewidth}\centering
  \includegraphics[height=1.1cm]{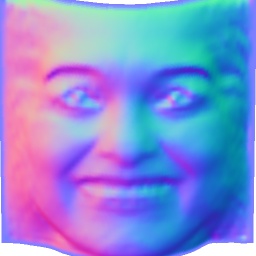}
  \includegraphics[height=1.1cm]{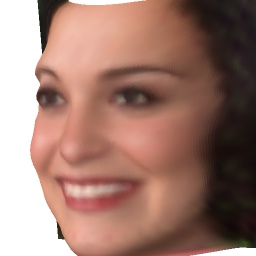}
  \includegraphics[trim={0 0 10mm 0}, clip, height=1.1cm]{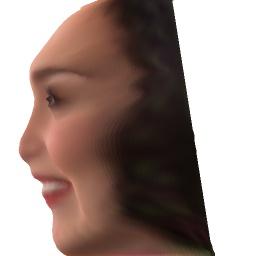}
  \includegraphics[height=1.1cm]{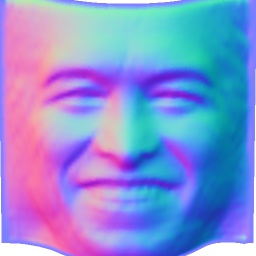}
  \includegraphics[height=1.1cm]{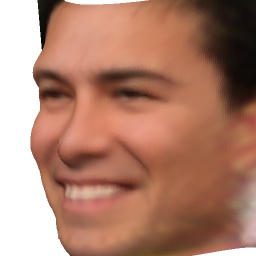}
  \includegraphics[trim={0 0 10mm 0}, clip, height=1.1cm]{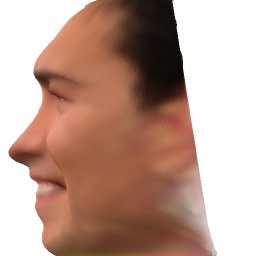}
  \caption{ours}
\end{subfigure}

\vspace{-0.5em}
\setlength{\belowcaptionskip}{-5pt}
\caption{\textbf{Qualitative comparison to SOTA.} Our method recovers much higher quality shapes compared to \cite{sahasrabudhe19lifting,Szabo19}.}\label{fig:compare_paper}
\end{figure}

%% file: fig-failure.tex
\begin{figure}[t]
\captionsetup[subfigure]{justification=centering,labelformat=simple,labelsep=colon,aboveskip=2pt}
\begin{subfigure}[b]{.36\linewidth}\centering
  \includegraphics[height=1cm]{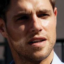}
  \includegraphics[trim={0 0 48px 0}, clip, height=1cm]{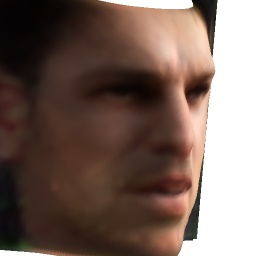}
  \includegraphics[trim={36px 0 12px 0}, clip, height=1cm]{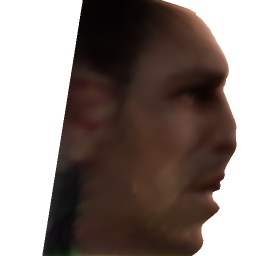}
  \caption{extreme lighting}\label{fig:failure_light}
\end{subfigure}\hfill
\begin{subfigure}[b]{.25\linewidth}\centering
  \includegraphics[height=1cm]{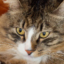}
  \includegraphics[trim={32px 0 32px 0}, clip, height=1cm]{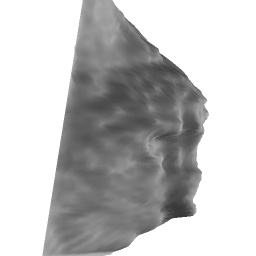}
  \caption{noisy texture}\label{fig:failure_texture}
\end{subfigure}\hfill
\begin{subfigure}[b]{.38\linewidth}\centering
  \includegraphics[height=1cm]{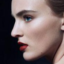}
  \includegraphics[height=1cm]{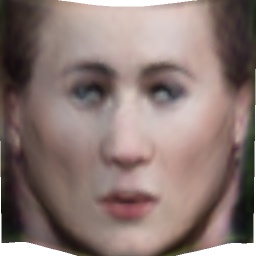}
  \includegraphics[trim={0 0 36px 0}, clip, height=1cm]{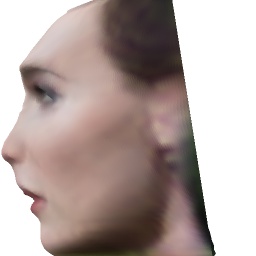}
  \caption{extreme pose}\label{fig:failure_pose}
\end{subfigure}

\vspace{-0.5em}
\setlength{\belowcaptionskip}{-5pt}
\caption{\textbf{Failure cases.} See~\cref{s:failure} for details.}\label{fig:failure}
\end{figure}

%% file: 5_conclusions.tex
\section{Conclusions}\label{s:conc}

We have presented a method that can learn a 3D model of a deformable object category from an unconstrained collection of single-view images of the object category.
The model is able to obtain high-fidelity monocular 3D reconstructions of individual object instances.
This is trained based on a reconstruction loss without any supervision, resembling an autoencoder.
We have shown that symmetry and illumination are strong cues for shape and help the model to converge to a meaningful reconstruction.
Our model outperforms a current state-of-the-art 3D reconstruction method that uses 2D keypoint supervision.
As for future work, the model currently represents 3D shape from a canonical viewpoint using a depth map, which is sufficient for objects such as faces that have a roughly convex shape and a natural canonical viewpoint.
For more complex objects, it may be possible to extend the model to use either multiple canonical views or a different 3D representation, such as a mesh or a voxel map.

%% file: 6_acknowledge.tex
\paragraph{Acknowledgements}
We would like to thank Soumyadip Sengupta for sharing with us the code to generate synthetic face datasets, and Mihir Sahasrabudhe for sending us the reconstruction results of Lifting AutoEncoders. We are also indebted to the members of Visual Geometry Group for insightful discussions. This work is jointly supported by Facebook Research and ERC Horizon 2020 research and innovation programme IDIU 638009.

%% file: 7_appendix.tex
\ifarxiv
\section{Supplementary Material}
\else
\section{Technical Details}
\fi

\subsection{Differentiable rendering layer}\label{sec:render}

As noted \ifarxiv in~\cref{s:formation}, \else in the Section~3.3 in the main paper, \fi the reprojection function $\Pi$ \emph{warps} the canonical image $\mathbf{J}$ to generate the actual image $\mathbf{I}$.
In CNNs, image warping is usually regarded as a simple operation that can be implemented efficiently using a bilinear resampling layer~\cite{jaderberg2015spatial}.
However, this is true only if we can easily send pixels $(u',v')$ in the warped image $\mathbf{I}$ back to pixels $(u,v)$ in the source image $\mathbf{J}$, a process also known as \emph{backward warping}.
Unfortunately, in our case the function $\eta_{d,w}$ obtained \ifarxiv  by~\cref{e:forward} \else by Eq.~(6) in the paper \fi sends pixels in the opposite way.

Implementing a \emph{forward warping} layer is surprisingly delicate.
One way of approaching the problem is to regard this task as a special case of rendering a textured mesh.
The \emph{Neural Mesh Renderer} (NMR) of~\cite{kato2018renderer} is a differentiable renderer of this type.
In our case, the mesh has one vertex per pixel and each group of $2\times 2$ adjacent pixels is tessellated by two triangles.
Empirically, we found the quality of the texture gradients of NMR to be poor in this case, likely caused by high frequency content in the texture image $\mathbf{J}$.

We solve the problem as follows.
First, we use NMR to warp only the depth map $d$, obtaining a version $\bar d$ of the depth map as seen from the input viewpoint.
This has two advantages: backpropagation through NMR is faster and secondly, the gradients are more stable, probably also due to the comparatively smooth nature of the depth map $d$ compared to the texture image $\mathbf{J}$.
Given the depth map $\bar d$, we then use the inverse \ifarxiv of~\cref{e:forward} \else of Eq.~(6) in the paper \fi to find the warp field from the observed viewpoint to the canonical viewpoint, and bilinearly resample the canonical image $\mathbf{J}$ to obtain the reconstruction.

\subsection{Training details}\label{s:train_detail}

We report the training details including all hyper-parameter settings in \cref{tab:param_set}, and detailed network architectures in \cref{tab:arch_depth_albedo,tab:arch_view_light,tab:arch_conf}.
We use standard encoder networks for both viewpoint and lighting predictions, and
encoder-decoder networks for depth, albedo and confidence predictions.
In order to mitigate checkerboard artifacts~\cite{odena2016deconvolution} in the predicted depth and albedo, we add a convolution layer after each deconvolution layer and replace the last deconvolotion layer with nearest-neighbor upsampling, followed by $3$ convolution layers.
Abbreviations of the operators are defined as follows:

\begin{itemize}
	\item $\text{Conv}(c_{in}, c_{out}, k, s, p)$: convolution with $c_{in}$ input channels, $c_{out}$ output channels, kernel size $k$, stride $s$ and padding $p$.
	\item $\text{Deconv}(c_{in}, c_{out}, k, s, p)$: deconvolution~\cite{Zeiler11} with $c_{in}$ input channels, $c_{out}$ output channels, kernel size $k$, stride $s$ and padding $p$.
	\item $\text{Upsample}(s)$: nearest-neighbor upsampling with a scale factor of $s$. 
	\item $\text{GN}(n)$: group normalization~\cite{GroupNorm2018} with $n$ groups.
	\item $\text{LReLU}(\alpha)$: leaky ReLU~\cite{lrelu2013} with a negative slope of $\alpha$.
\end{itemize}

\input{supmat/tab-param-set.tex}
\input{supmat/tab-architecture.tex}

\section{Qualitative Results}\label{s:submat_result}

We provide more qualitative results in the following and 3D animations in the supplementary video\footnote{\scriptsize\url{https://www.youtube.com/watch?v=5rPJyrU-WE4}}.
\cref{fig:ablation} reports the qualitative results of the ablated models \ifarxiv in~\cref{tab:ablation}. \else in Table~3 of the main paper. \fi
\cref{fig:sup-face} shows reconstruction results on human faces from CelebA and 3DFAW\@.
We also show reconstruction results on face paintings and drawings collected from~\cite{Crowley15} and the Internet in \cref{fig:sup-paint,fig:sup-abstract-face}.
\cref{fig:sup-cat,fig:sup-abstract-cat,fig:sup-car} show results on real cat faces from~\cite{zhang2008cat,parkhi12a}, abstract cats collected from the Internet and synthetic cars rendered using ShapeNet.

\paragraph{Re-lighting.}
Since our model predicts the intrinsic components of an image, separating the albedo and illumination, we can easily re-light the objects with different lighting conditions.
In \cref{fig:sup-relight}, we demonstrate results of the intrinsic decomposition and the re-lit faces in the canonical view.

\paragraph{Testing on videos.}
To further assess our model, we apply the model trained on CelebA faces to VoxCeleb~\cite{Chung18a} videos \emph{frame by frame} and include the results in the supplementary video.
Our trained model works surprisingly well, producing consistent, smooth reconstructions across different frames and recovering the details of the facial motions accurately.

\input{supmat/fig-ablation.tex}
\input{supmat/fig-relight.tex}
\input{supmat/fig-face.tex}
\input{supmat/fig-paint.tex}
\input{supmat/fig-abstract-face.tex}
\input{supmat/fig-cat.tex}
\input{supmat/fig-abstract-cat.tex}
\input{supmat/fig-car.tex}

%% file: supmat/tab-param-set.tex
\begin{table}[t]
\footnotesize
\begin{center}
\begin{tabular}{lc}
\toprule
 Parameter & Value/Range \\ \midrule
 Optimizer & Adam \\
 Learning rate & $1\times 10^{-4}$ \\
 Number of epochs & $30$ \\
 Batch size & $64$ \\
 Loss weight $\lambda_{\text{f}}$ & $0.5$ \\
 Loss weight $\lambda_{\text{p}}$ & $1$ \\
 Input image size & $64 \times 64$ \\
 Output image size & $64 \times 64$ \\ \midrule
 Depth map & $(0.9, 1.1)$ \\
 Albedo & $(0, 1)$ \\
 Light coefficient $k_s$ & $(0, 1)$ \\
 Light coefficient $k_d$ & $(0, 1)$ \\
 Light direction $l_x, l_y$ & $(-1, 1)$ \\
 Viewpoint rotation $w_{1:3}$ & $(-60^\circ, 60^\circ)$ \\
 Viewpoint translation $w_{4:6}$ & $(-0.1, 0.1)$ \\
 Field of view (FOV) & $10$ \\
\bottomrule
\end{tabular}
\end{center}
\caption{Training details and hyper-parameter settings.}\label{tab:param_set}
\end{table}

%% file: supmat/tab-architecture.tex
\begin{table}[t]
\footnotesize
\begin{center}
\begin{tabular}{lc}
\toprule
 Encoder & Output size\\ \midrule
 Conv(3, 32, 4, 2, 1) + ReLU & 32\\
 Conv(32, 64, 4, 2, 1) + ReLU & 16\\
 Conv(64, 128, 4, 2, 1) + ReLU & 8\\
 Conv(128, 256, 4, 2, 1) + ReLU & 4\\
 Conv(256, 256, 4, 1, 0) + ReLU & 1\\
 Conv(256, $c_{out}$, 1, 1, 0) + Tanh $\rightarrow output$ & 1\\
\bottomrule
\end{tabular}
\end{center}
\caption{Network architecture for viewpoint and lighting. The output channel size $c_{out}$ is $6$ for viewpoint, corresponding to rotation angles $w_{1:3}$ and translations $w_{4:6}$ in $x$, $y$ and $z$ axes, and $4$ for lighting, corresponding to $k_s$, $k_d$, $l_x$ and $l_y$.}\label{tab:arch_view_light}
\end{table}

\begin{table}[t]
\footnotesize
\begin{center}
\begin{tabular}{lc}
\toprule
 Encoder & Output size \\ \midrule
 Conv(3, 64, 4, 2, 1) + GN(16) + LReLU(0.2) & 32\\
 Conv(64, 128, 4, 2, 1) + GN(32) + LReLU(0.2) & 16\\
 Conv(128, 256, 4, 2, 1) + GN(64) + LReLU(0.2) & 8\\
 Conv(256, 512, 4, 2, 1) + LReLU(0.2) & 4\\
 Conv(512, 256, 4, 1, 0) + ReLU & 1\\ \midrule \midrule
 Decoder & Output size \\ \midrule
 Deconv(256, 512, 4, 1, 0) + ReLU & 4\\
 Conv(512, 512, 3, 1, 1) + ReLU & 4\\
 Deconv(512, 256, 4, 2, 1) + GN(64) + ReLU & 8\\
 Conv(256, 256, 3, 1, 1) + GN(64) + ReLU & 8\\
 Deconv(256, 128, 4, 2, 1) + GN(32) + ReLU & 16\\
 Conv(128, 128, 3, 1, 1) + GN(32) + ReLU & 16\\
 Deconv(128, 64, 4, 2, 1) + GN(16) + ReLU & 32\\
 Conv(64, 64, 3, 1, 1) + GN(16) + ReLU & 32\\
 Upsample(2) & 64\\
 Conv(64, 64, 3, 1, 1) + GN(16) + ReLU & 64\\
 Conv(64, 64, 5, 1, 2) + GN(16) + ReLU & 64\\
 Conv(64, $c_{out}$, 5, 1, 2) + Tanh $\rightarrow output$ & 64\\
\bottomrule
\end{tabular}
\end{center}
\caption{Network architecture for depth and albedo. The output channel size $c_{out}$ is $1$ for depth and $3$ for albedo.}\label{tab:arch_depth_albedo}
\end{table}

\begin{table}[t]
\footnotesize
\begin{center}
\begin{tabular}{lc}
\toprule
 Encoder & Output size \\ \midrule
 Conv(3, 64, 4, 2, 1) + GN(16) + LReLU(0.2) & 32\\
 Conv(64, 128, 4, 2, 1) + GN(32) + LReLU(0.2) & 16\\
 Conv(128, 256, 4, 2, 1) + GN(64) + LReLU(0.2) & 8\\
 Conv(256, 512, 4, 2, 1) + LReLU(0.2) & 4\\
 Conv(512, 128, 4, 1, 0) + ReLU & 1\\ \midrule \midrule
 Decoder & Output size \\ \midrule
 Deconv(128, 512, 4, 1, 0) + ReLU & 4\\
 Deconv(512, 256, 4, 2, 1) + GN(64) + ReLU & 8\\
 Deconv(256, 128, 4, 2, 1) + GN(32) + ReLU & 16\\
 \enskip \rotatebox[origin=c]{180}{$\Lsh$} Conv(128, 2, 3, 1, 1) + SoftPlus $\rightarrow output$ & 16\\
 Deconv(128, 64, 4, 2, 1) + GN(16) + ReLU & 32\\
 Deconv(64, 64, 4, 2, 1) + GN(16) + ReLU & 64\\
 Conv(64, 2, 5, 1, 2) + SoftPlus $\rightarrow output$ & 64\\
\bottomrule
\end{tabular}
\end{center}
\caption{Network architecture for confidence maps. The network outputs two pairs of confidence maps at different spatial resolutions for photometric and perceptual losses.}\label{tab:arch_conf}
\end{table}

%% file: supmat/fig-ablation.tex
\begin{figure*}[t]\centering
  \includegraphics[width=0.875\linewidth]{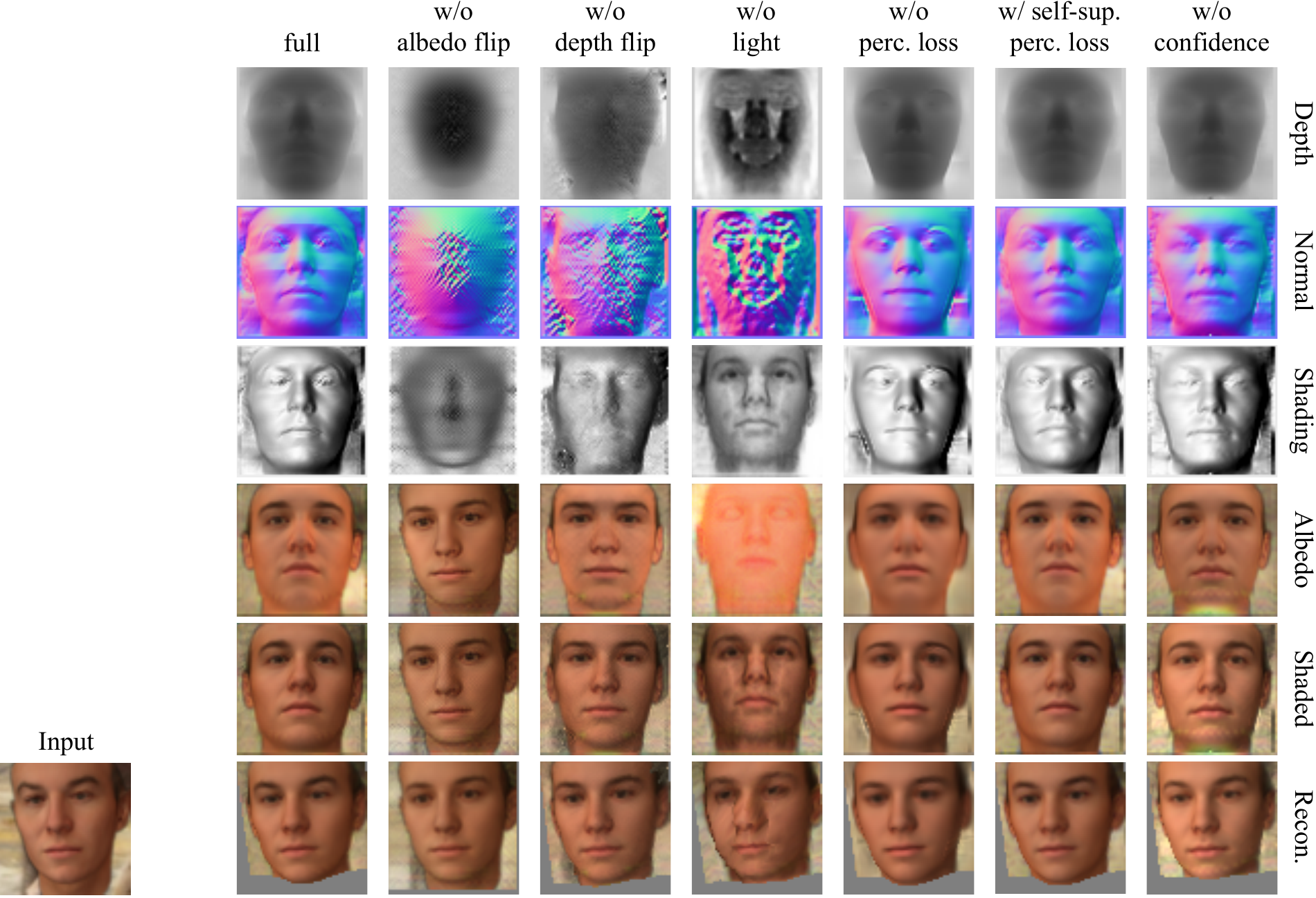}
\caption{\textbf{Qualitative results of the ablated models.}}\label{fig:ablation}
\end{figure*}

%% file: supmat/fig-relight.tex
\begin{figure*}[t]\centering
\captionsetup[subfigure]{justification=centering,labelformat=empty,labelsep=colon,aboveskip=2pt}
\begin{subfigure}[b]{.2\linewidth}\centering
  \includegraphics[height=1.5cm]{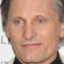}
\end{subfigure}
\begin{subfigure}[b]{.1\linewidth}\centering
  \includegraphics[height=1.5cm]{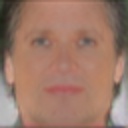}
\end{subfigure}
\begin{subfigure}[b]{.2\linewidth}\centering
  \includegraphics[height=1.5cm]{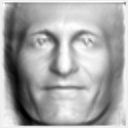}
  \includegraphics[height=1.5cm]{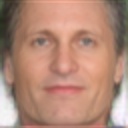}
\end{subfigure}
\begin{subfigure}[b]{.47\linewidth}\centering
  \includegraphics[height=1.5cm]{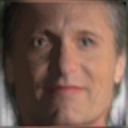}
  \includegraphics[height=1.5cm]{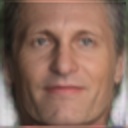}
  \includegraphics[height=1.5cm]{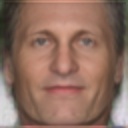}
  \includegraphics[height=1.5cm]{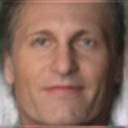}
  \includegraphics[height=1.5cm]{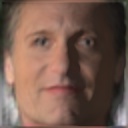}
\end{subfigure}
\\
\begin{subfigure}[b]{.2\linewidth}\centering
  \includegraphics[height=1.5cm]{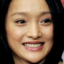}
\end{subfigure}
\begin{subfigure}[b]{.1\linewidth}\centering
  \includegraphics[height=1.5cm]{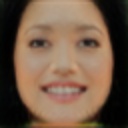}
\end{subfigure}
\begin{subfigure}[b]{.2\linewidth}\centering
  \includegraphics[height=1.5cm]{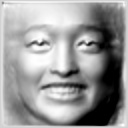}
  \includegraphics[height=1.5cm]{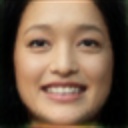}
\end{subfigure}
\begin{subfigure}[b]{.47\linewidth}\centering
  \includegraphics[height=1.5cm]{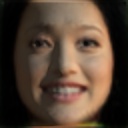}
  \includegraphics[height=1.5cm]{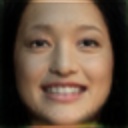}
  \includegraphics[height=1.5cm]{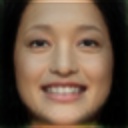}
  \includegraphics[height=1.5cm]{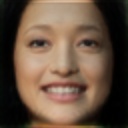}
  \includegraphics[height=1.5cm]{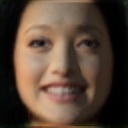}
\end{subfigure}
\\
\begin{subfigure}[b]{.2\linewidth}\centering
  \includegraphics[height=1.5cm]{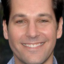}
\end{subfigure}
\begin{subfigure}[b]{.1\linewidth}\centering
  \includegraphics[height=1.5cm]{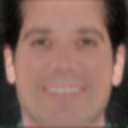}
\end{subfigure}
\begin{subfigure}[b]{.2\linewidth}\centering
  \includegraphics[height=1.5cm]{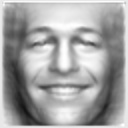}
  \includegraphics[height=1.5cm]{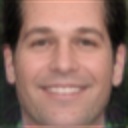}
\end{subfigure}
\begin{subfigure}[b]{.47\linewidth}\centering
  \includegraphics[height=1.5cm]{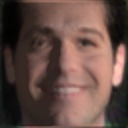}
  \includegraphics[height=1.5cm]{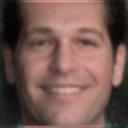}
  \includegraphics[height=1.5cm]{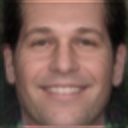}
  \includegraphics[height=1.5cm]{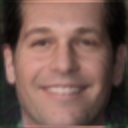}
  \includegraphics[height=1.5cm]{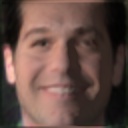}
\end{subfigure}
\\
\begin{subfigure}[b]{.2\linewidth}\centering
  \includegraphics[height=1.5cm]{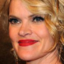}
\end{subfigure}
\begin{subfigure}[b]{.1\linewidth}\centering
  \includegraphics[height=1.5cm]{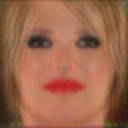}
\end{subfigure}
\begin{subfigure}[b]{.2\linewidth}\centering
  \includegraphics[height=1.5cm]{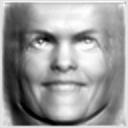}
  \includegraphics[height=1.5cm]{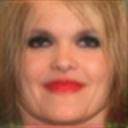}
\end{subfigure}
\begin{subfigure}[b]{.47\linewidth}\centering
  \includegraphics[height=1.5cm]{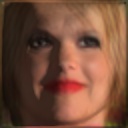}
  \includegraphics[height=1.5cm]{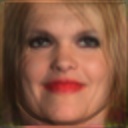}
  \includegraphics[height=1.5cm]{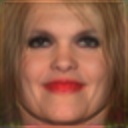}
  \includegraphics[height=1.5cm]{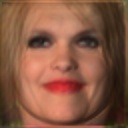}
  \includegraphics[height=1.5cm]{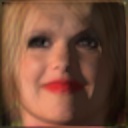}
\end{subfigure}
\\
\begin{subfigure}[b]{.2\linewidth}\centering
  \includegraphics[height=1.5cm]{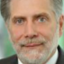}
\end{subfigure}
\begin{subfigure}[b]{.1\linewidth}\centering
  \includegraphics[height=1.5cm]{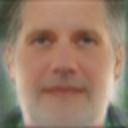}
\end{subfigure}
\begin{subfigure}[b]{.2\linewidth}\centering
  \includegraphics[height=1.5cm]{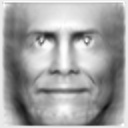}
  \includegraphics[height=1.5cm]{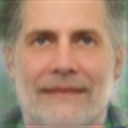}
\end{subfigure}
\begin{subfigure}[b]{.47\linewidth}\centering
  \includegraphics[height=1.5cm]{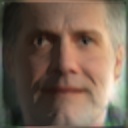}
  \includegraphics[height=1.5cm]{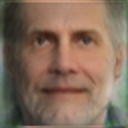}
  \includegraphics[height=1.5cm]{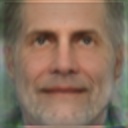}
  \includegraphics[height=1.5cm]{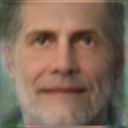}
  \includegraphics[height=1.5cm]{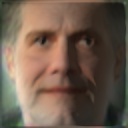}
\end{subfigure}
\\
\begin{subfigure}[b]{.2\linewidth}\centering
  \includegraphics[height=1.5cm]{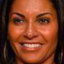}
  \caption{Input}
\end{subfigure}
\begin{subfigure}[b]{.1\linewidth}\centering
  \includegraphics[height=1.5cm]{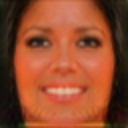}
  \caption{Albedo}
\end{subfigure}
\begin{subfigure}[b]{.2\linewidth}\centering
  \includegraphics[height=1.5cm]{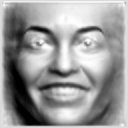}
  \includegraphics[height=1.5cm]{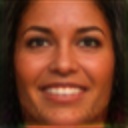}
  \caption{Original light}
\end{subfigure}
\begin{subfigure}[b]{.47\linewidth}\centering
  \includegraphics[height=1.5cm]{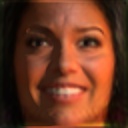}
  \includegraphics[height=1.5cm]{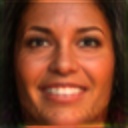}
  \includegraphics[height=1.5cm]{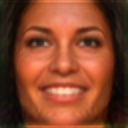}
  \includegraphics[height=1.5cm]{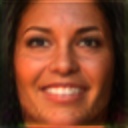}
  \includegraphics[height=1.5cm]{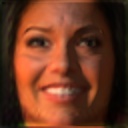}
  \caption{Re-lit}
\end{subfigure}

\caption{\textbf{Re-lighting effects.}}\label{fig:sup-relight}
\end{figure*}

%% file: supmat/fig-face.tex
\begin{figure*}[t]\centering
\captionsetup[subfigure]{justification=centering,labelformat=empty,labelsep=colon,aboveskip=2pt}
\begin{subfigure}[b]{.2\linewidth}\centering
  \includegraphics[height=1.5cm]{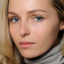}
\end{subfigure}
\begin{subfigure}[b]{.7\linewidth}\centering
  \includegraphics[height=1.5cm]{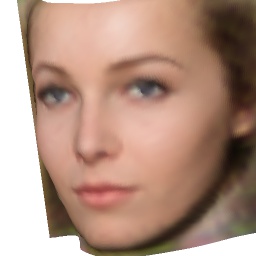}
  \includegraphics[height=1.5cm]{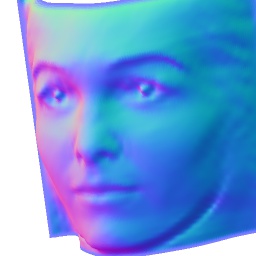}
  \includegraphics[trim={0 0 40px 0}, clip, height=1.5cm]{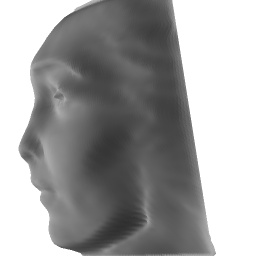}
  \includegraphics[trim={24px 0 0 0}, clip, height=1.5cm]{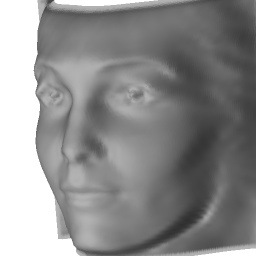}
  \includegraphics[height=1.5cm]{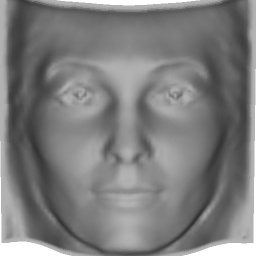}
  \includegraphics[height=1.5cm]{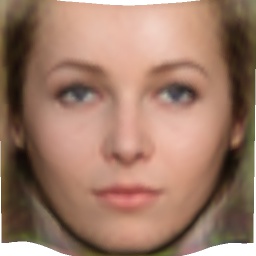}
  \includegraphics[trim={0 0 24px 0}, clip, height=1.5cm]{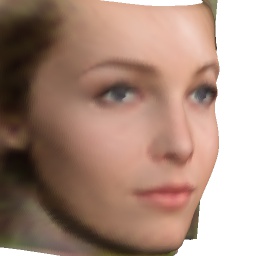}
  \includegraphics[trim={40px 0 0 0}, clip, height=1.5cm]{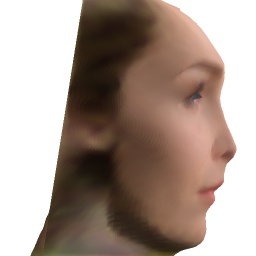}
\end{subfigure}
\\
\begin{subfigure}[b]{.2\linewidth}\centering
  \includegraphics[height=1.5cm]{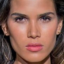}
\end{subfigure}
\begin{subfigure}[b]{.7\linewidth}\centering
  \includegraphics[height=1.5cm]{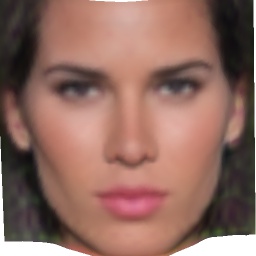}
  \includegraphics[height=1.5cm]{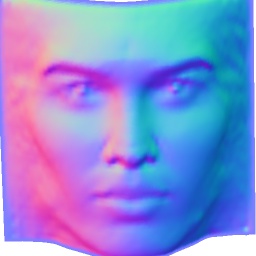}
  \includegraphics[trim={0 0 40px 0}, clip, height=1.5cm]{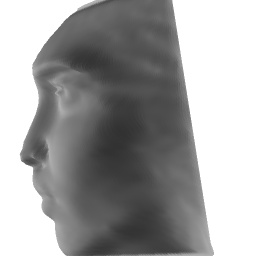}
  \includegraphics[trim={24px 0 0 0}, clip, height=1.5cm]{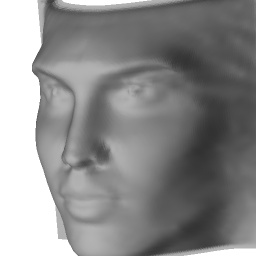}
  \includegraphics[height=1.5cm]{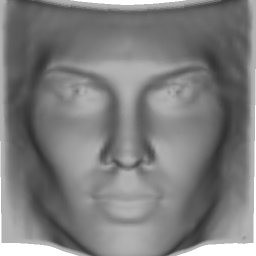}
  \includegraphics[height=1.5cm]{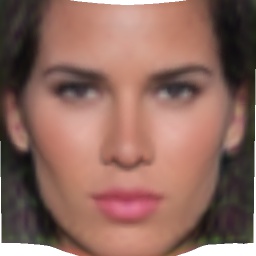}
  \includegraphics[trim={0 0 24px 0}, clip, height=1.5cm]{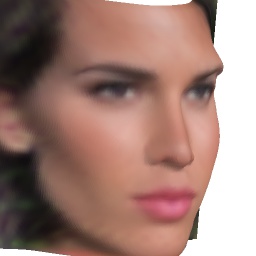}
  \includegraphics[trim={40px 0 0 0}, clip, height=1.5cm]{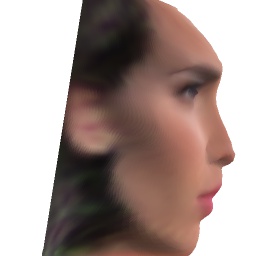}
\end{subfigure}
\\
\begin{subfigure}[b]{.2\linewidth}\centering
  \includegraphics[height=1.5cm]{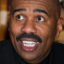}
\end{subfigure}
\begin{subfigure}[b]{.7\linewidth}\centering
  \includegraphics[height=1.5cm]{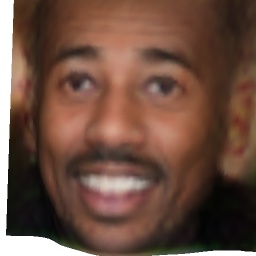}
  \includegraphics[height=1.5cm]{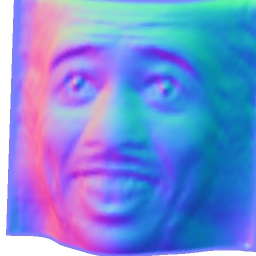}
  \includegraphics[trim={0 0 40px 0}, clip, height=1.5cm]{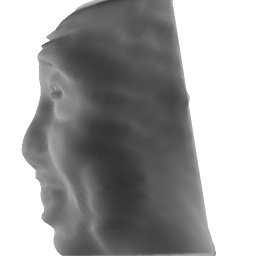}
  \includegraphics[trim={24px 0 0 0}, clip, height=1.5cm]{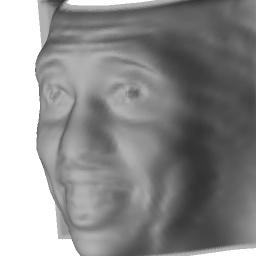}
  \includegraphics[height=1.5cm]{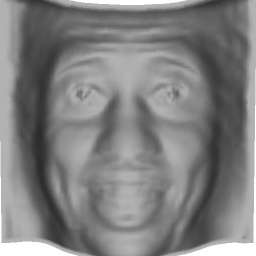}
  \includegraphics[height=1.5cm]{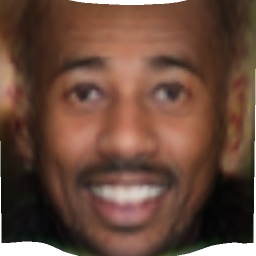}
  \includegraphics[trim={0 0 24px 0}, clip, height=1.5cm]{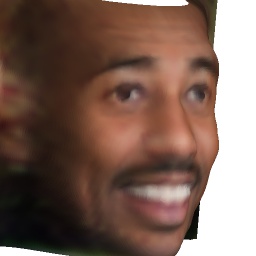}
  \includegraphics[trim={40px 0 0 0}, clip, height=1.5cm]{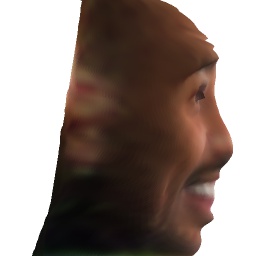}
\end{subfigure}
\\
\begin{subfigure}[b]{.2\linewidth}\centering
  \includegraphics[height=1.5cm]{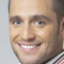}
\end{subfigure}
\begin{subfigure}[b]{.7\linewidth}\centering
  \includegraphics[height=1.5cm]{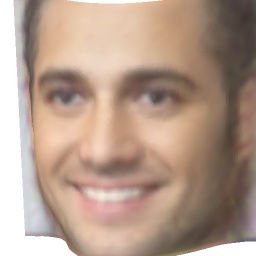}
  \includegraphics[height=1.5cm]{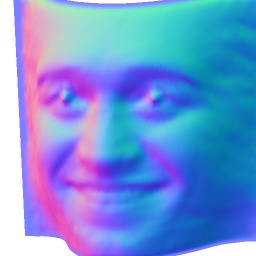}
  \includegraphics[trim={0 0 40px 0}, clip, height=1.5cm]{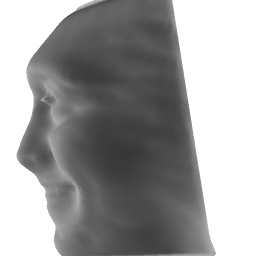}
  \includegraphics[trim={24px 0 0 0}, clip, height=1.5cm]{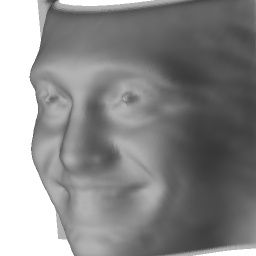}
  \includegraphics[height=1.5cm]{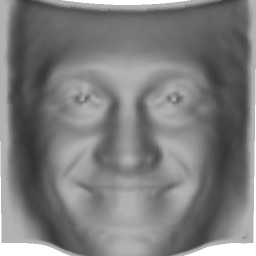}
  \includegraphics[height=1.5cm]{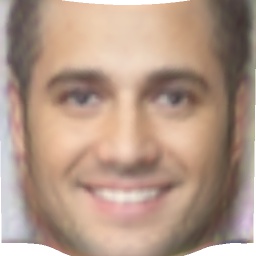}
  \includegraphics[trim={0 0 24px 0}, clip, height=1.5cm]{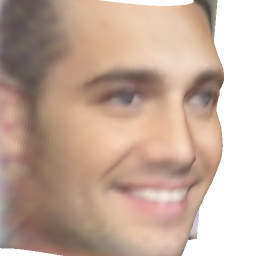}
  \includegraphics[trim={40px 0 0 0}, clip, height=1.5cm]{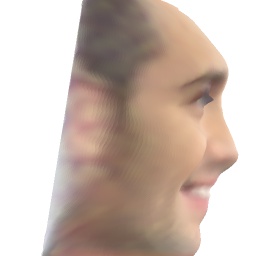}
\end{subfigure}
\\
\begin{subfigure}[b]{.2\linewidth}\centering
  \includegraphics[height=1.5cm]{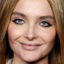}
\end{subfigure}
\begin{subfigure}[b]{.7\linewidth}\centering
  \includegraphics[height=1.5cm]{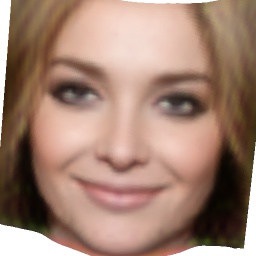}
  \includegraphics[height=1.5cm]{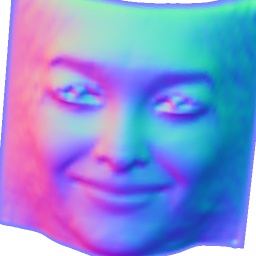}
  \includegraphics[trim={0 0 40px 0}, clip, height=1.5cm]{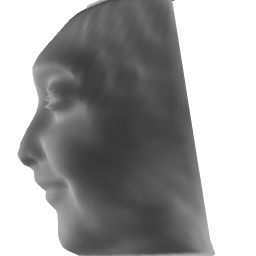}
  \includegraphics[trim={24px 0 0 0}, clip, height=1.5cm]{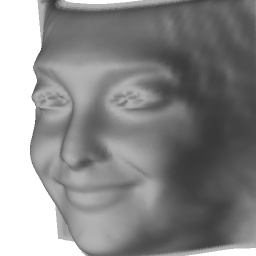}
  \includegraphics[height=1.5cm]{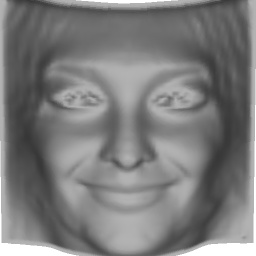}
  \includegraphics[height=1.5cm]{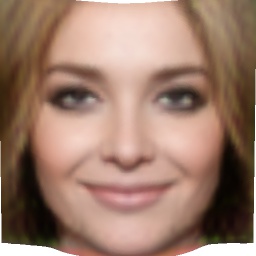}
  \includegraphics[trim={0 0 24px 0}, clip, height=1.5cm]{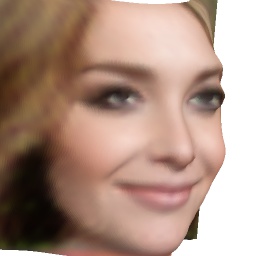}
  \includegraphics[trim={40px 0 0 0}, clip, height=1.5cm]{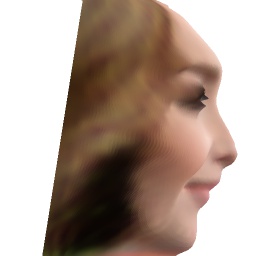}
\end{subfigure}
\\
\begin{subfigure}[b]{.2\linewidth}\centering
  \includegraphics[height=1.5cm]{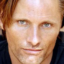}
\end{subfigure}
\begin{subfigure}[b]{.7\linewidth}\centering
  \includegraphics[height=1.5cm]{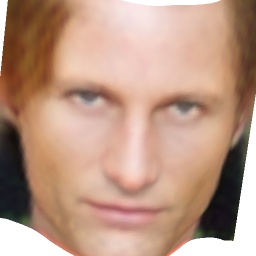}
  \includegraphics[height=1.5cm]{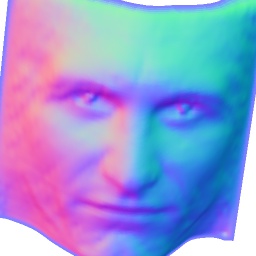}
  \includegraphics[trim={0 0 40px 0}, clip, height=1.5cm]{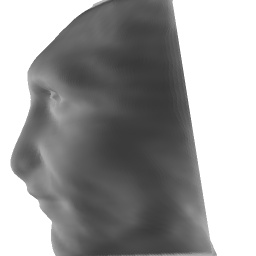}
  \includegraphics[trim={24px 0 0 0}, clip, height=1.5cm]{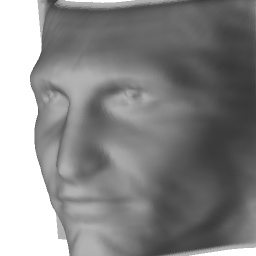}
  \includegraphics[height=1.5cm]{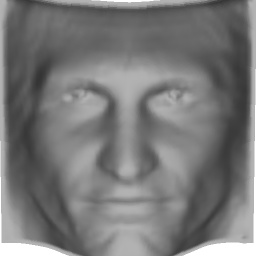}
  \includegraphics[height=1.5cm]{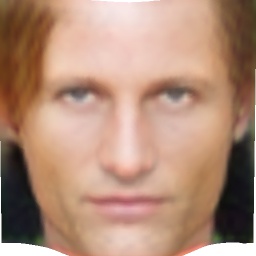}
  \includegraphics[trim={0 0 24px 0}, clip, height=1.5cm]{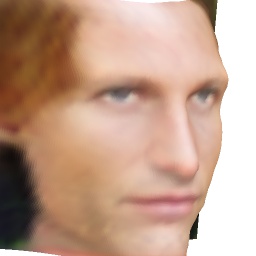}
  \includegraphics[trim={40px 0 0 0}, clip, height=1.5cm]{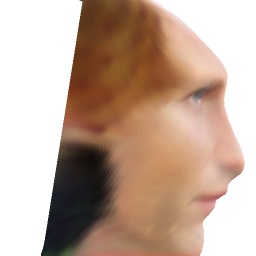}
\end{subfigure}
\\
\begin{subfigure}[b]{.2\linewidth}\centering
  \includegraphics[height=1.5cm]{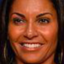}
\end{subfigure}
\begin{subfigure}[b]{.7\linewidth}\centering
  \includegraphics[height=1.5cm]{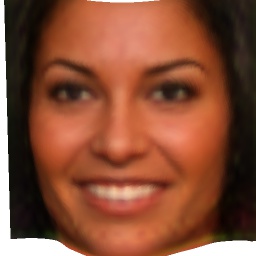}
  \includegraphics[height=1.5cm]{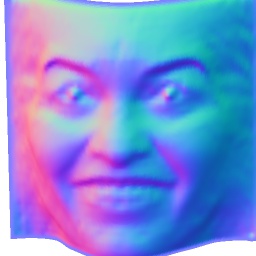}
  \includegraphics[trim={0 0 40px 0}, clip, height=1.5cm]{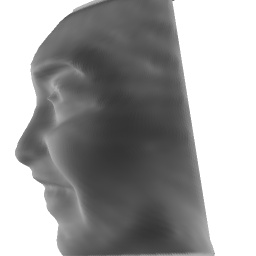}
  \includegraphics[trim={24px 0 0 0}, clip, height=1.5cm]{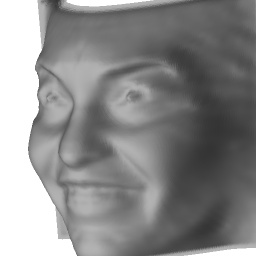}
  \includegraphics[height=1.5cm]{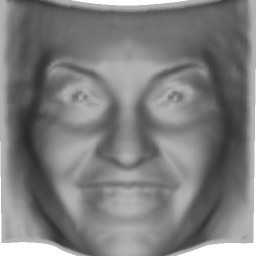}
  \includegraphics[height=1.5cm]{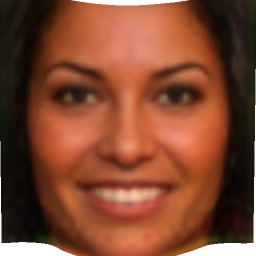}
  \includegraphics[trim={0 0 24px 0}, clip, height=1.5cm]{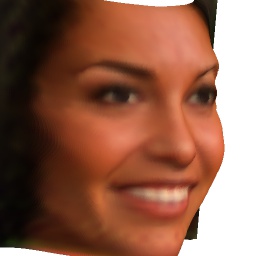}
  \includegraphics[trim={40px 0 0 0}, clip, height=1.5cm]{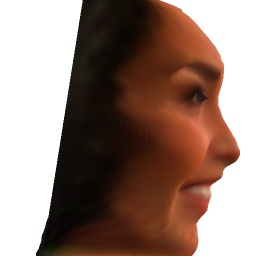}
\end{subfigure}
\\
\begin{subfigure}[b]{.2\linewidth}\centering
  \includegraphics[height=1.5cm]{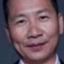}
\end{subfigure}
\begin{subfigure}[b]{.7\linewidth}\centering
  \includegraphics[height=1.5cm]{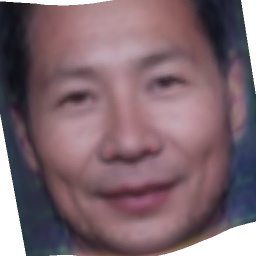}
  \includegraphics[height=1.5cm]{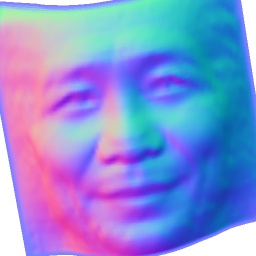}
  \includegraphics[trim={0 0 40px 0}, clip, height=1.5cm]{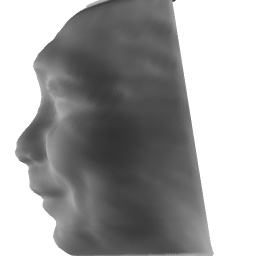}
  \includegraphics[trim={24px 0 0 0}, clip, height=1.5cm]{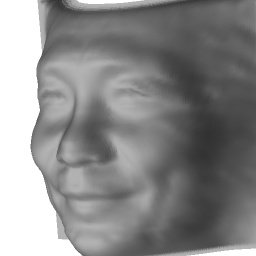}
  \includegraphics[height=1.5cm]{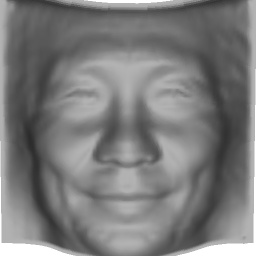}
  \includegraphics[height=1.5cm]{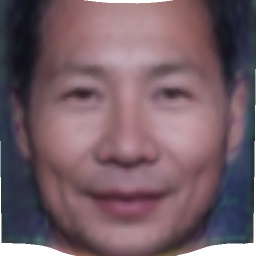}
  \includegraphics[trim={0 0 24px 0}, clip, height=1.5cm]{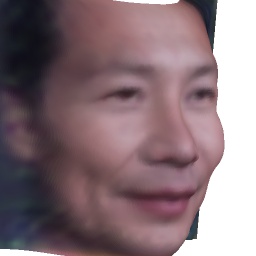}
  \includegraphics[trim={40px 0 0 0}, clip, height=1.5cm]{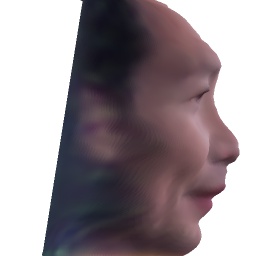}
\end{subfigure}
\\
\begin{subfigure}[b]{.2\linewidth}\centering
  \includegraphics[height=1.5cm]{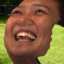}
\end{subfigure}
\begin{subfigure}[b]{.7\linewidth}\centering
  \includegraphics[height=1.5cm]{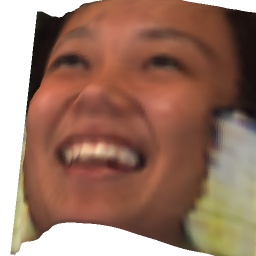}
  \includegraphics[height=1.5cm]{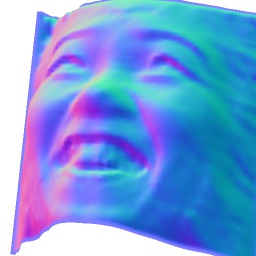}
  \includegraphics[trim={0 0 40px 0}, clip, height=1.5cm]{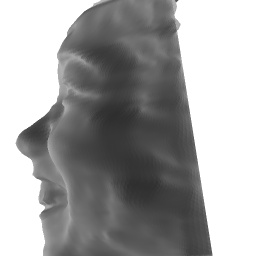}
  \includegraphics[trim={24px 0 0 0}, clip, height=1.5cm]{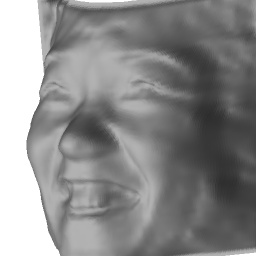}
  \includegraphics[height=1.5cm]{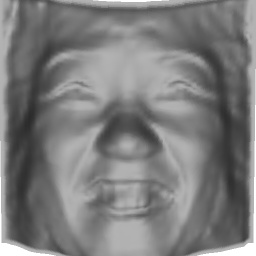}
  \includegraphics[height=1.5cm]{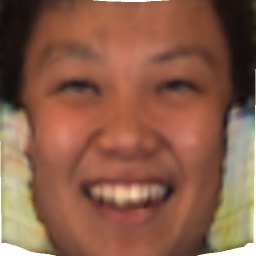}
  \includegraphics[trim={0 0 24px 0}, clip, height=1.5cm]{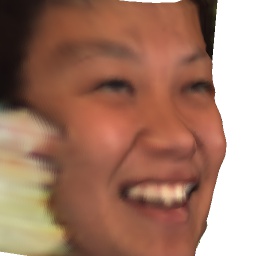}
  \includegraphics[trim={40px 0 0 0}, clip, height=1.5cm]{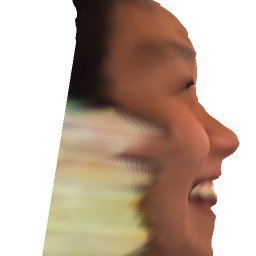}
\end{subfigure}
\\
\begin{subfigure}[b]{.2\linewidth}\centering
  \includegraphics[height=1.5cm]{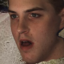}
\end{subfigure}
\begin{subfigure}[b]{.7\linewidth}\centering
  \includegraphics[height=1.5cm]{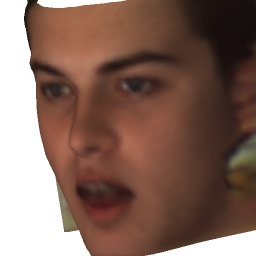}
  \includegraphics[height=1.5cm]{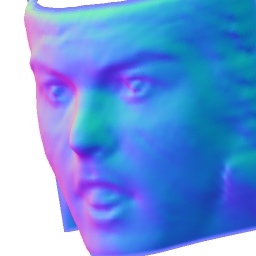}
  \includegraphics[trim={0 0 40px 0}, clip, height=1.5cm]{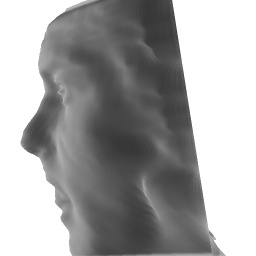}
  \includegraphics[trim={24px 0 0 0}, clip, height=1.5cm]{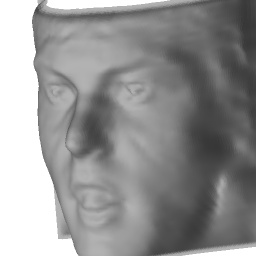}
  \includegraphics[height=1.5cm]{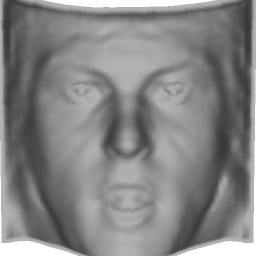}
  \includegraphics[height=1.5cm]{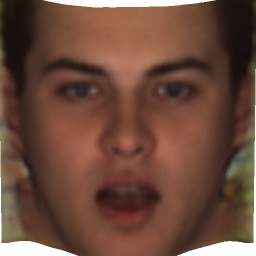}
  \includegraphics[trim={0 0 24px 0}, clip, height=1.5cm]{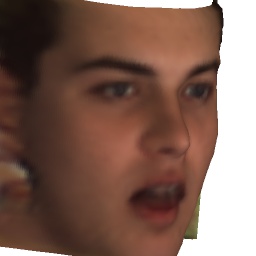}
  \includegraphics[trim={40px 0 0 0}, clip, height=1.5cm]{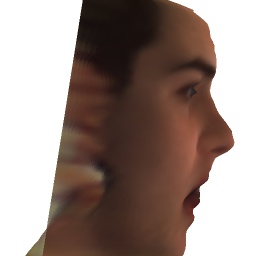}
\end{subfigure}
\\
\begin{subfigure}[b]{.2\linewidth}\centering
  \includegraphics[height=1.5cm]{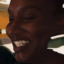}
\end{subfigure}
\begin{subfigure}[b]{.7\linewidth}\centering
  \includegraphics[height=1.5cm]{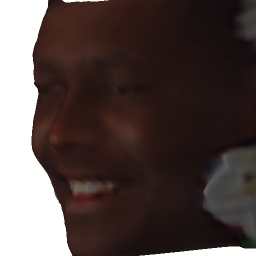}
  \includegraphics[height=1.5cm]{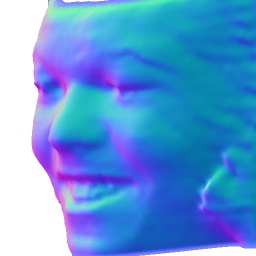}
  \includegraphics[trim={0 0 40px 0}, clip, height=1.5cm]{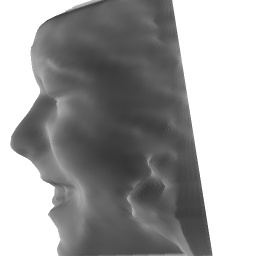}
  \includegraphics[trim={24px 0 0 0}, clip, height=1.5cm]{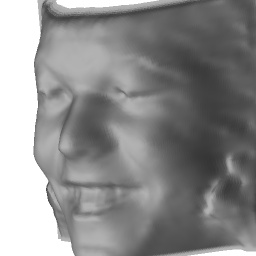}
  \includegraphics[height=1.5cm]{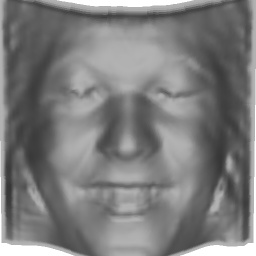}
  \includegraphics[height=1.5cm]{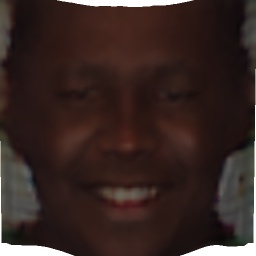}
  \includegraphics[trim={0 0 24px 0}, clip, height=1.5cm]{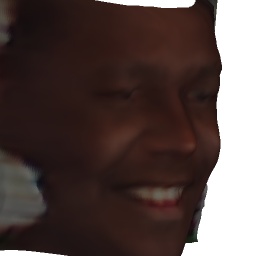}
  \includegraphics[trim={40px 0 0 0}, clip, height=1.5cm]{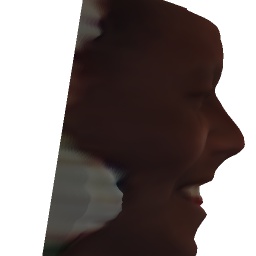}
\end{subfigure}
\\
\begin{subfigure}[b]{.2\linewidth}\centering
  \includegraphics[height=1.5cm]{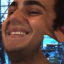}
\end{subfigure}
\begin{subfigure}[b]{.7\linewidth}\centering
  \includegraphics[height=1.5cm]{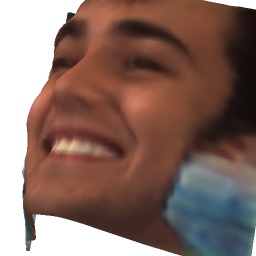}
  \includegraphics[height=1.5cm]{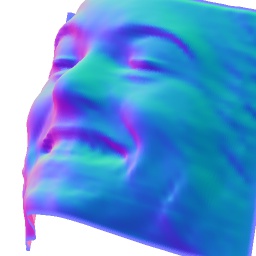}
  \includegraphics[trim={0 0 40px 0}, clip, height=1.5cm]{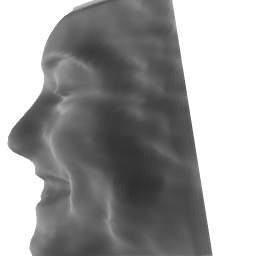}
  \includegraphics[trim={24px 0 0 0}, clip, height=1.5cm]{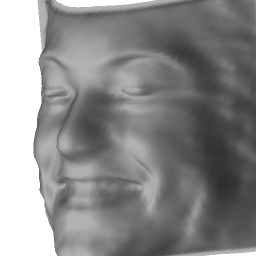}
  \includegraphics[height=1.5cm]{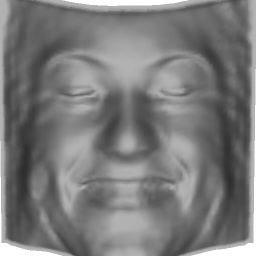}
  \includegraphics[height=1.5cm]{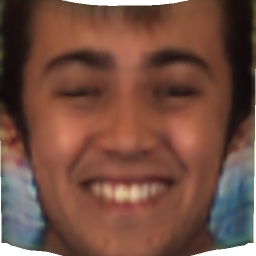}
  \includegraphics[trim={0 0 24px 0}, clip, height=1.5cm]{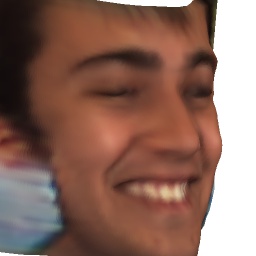}
  \includegraphics[trim={40px 0 0 0}, clip, height=1.5cm]{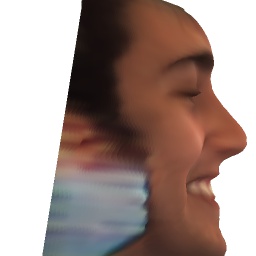}
\end{subfigure}
\\
\begin{subfigure}[b]{.2\linewidth}\centering
  \includegraphics[height=1.5cm]{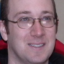}
\end{subfigure}
\begin{subfigure}[b]{.7\linewidth}\centering
  \includegraphics[height=1.5cm]{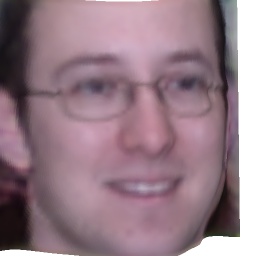}
  \includegraphics[height=1.5cm]{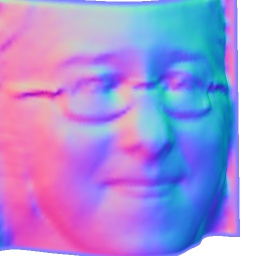}
  \includegraphics[trim={0 0 40px 0}, clip, height=1.5cm]{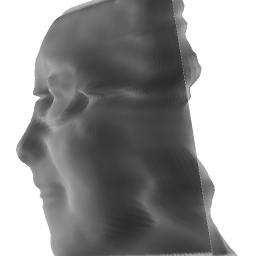}
  \includegraphics[trim={24px 0 0 0}, clip, height=1.5cm]{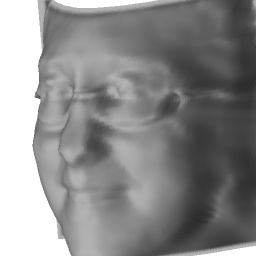}
  \includegraphics[height=1.5cm]{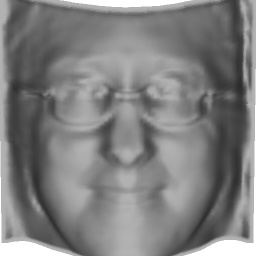}
  \includegraphics[height=1.5cm]{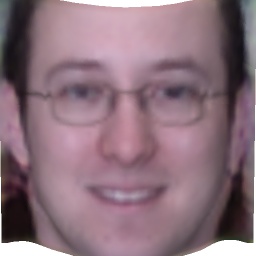}
  \includegraphics[trim={0 0 24px 0}, clip, height=1.5cm]{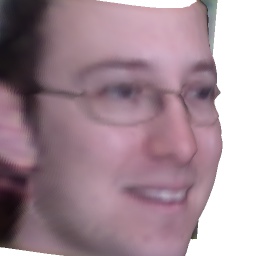}
  \includegraphics[trim={40px 0 0 0}, clip, height=1.5cm]{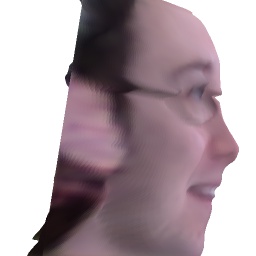}
\end{subfigure}
\\
\begin{subfigure}[b]{.2\linewidth}\centering
  \includegraphics[height=1.5cm]{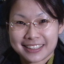}
  \caption{Input}
\end{subfigure}
\begin{subfigure}[b]{.7\linewidth}\centering
  \includegraphics[height=1.5cm]{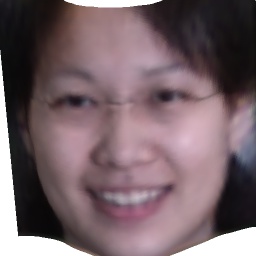}
  \includegraphics[height=1.5cm]{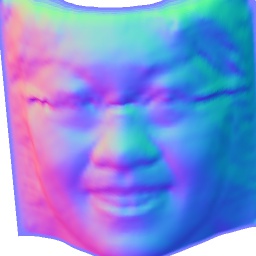}
  \includegraphics[trim={0 0 40px 0}, clip, height=1.5cm]{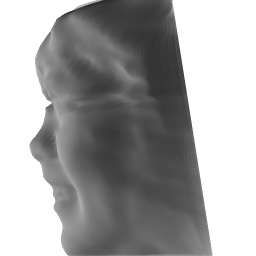}
  \includegraphics[trim={24px 0 0 0}, clip, height=1.5cm]{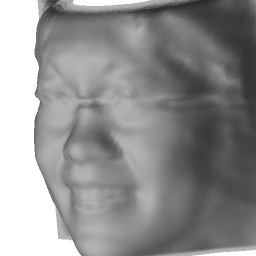}
  \includegraphics[height=1.5cm]{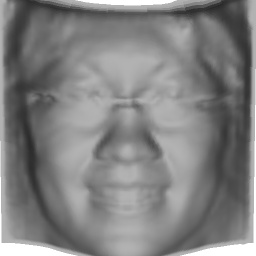}
  \includegraphics[height=1.5cm]{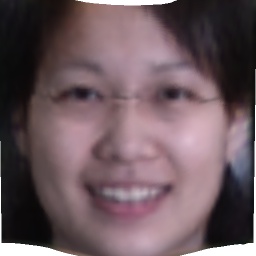}
  \includegraphics[trim={0 0 24px 0}, clip, height=1.5cm]{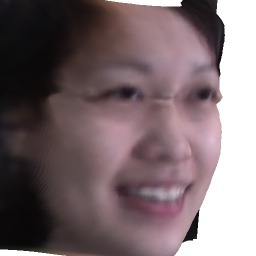}
  \includegraphics[trim={40px 0 0 0}, clip, height=1.5cm]{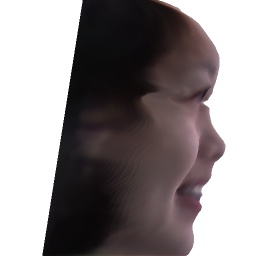}
  \caption{Reconstruction}
\end{subfigure}

\caption{\textbf{Reconstruction of human faces.}}\label{fig:sup-face}
\end{figure*}

%% file: supmat/fig-paint.tex
\begin{figure*}[t]\centering
\captionsetup[subfigure]{justification=centering,labelformat=empty,labelsep=colon,aboveskip=2pt}
\begin{subfigure}[b]{.2\linewidth}\centering
  \includegraphics[height=1.5cm]{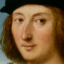}
\end{subfigure}
\begin{subfigure}[b]{.7\linewidth}\centering
  \includegraphics[height=1.5cm]{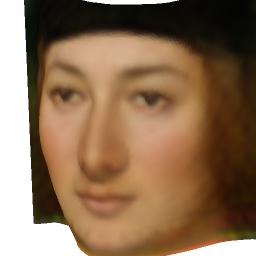}
  \includegraphics[height=1.5cm]{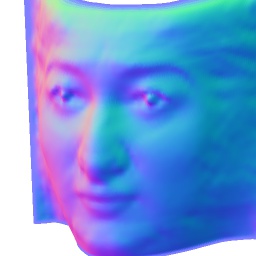}
  \includegraphics[trim={0 0 40px 0}, clip, height=1.5cm]{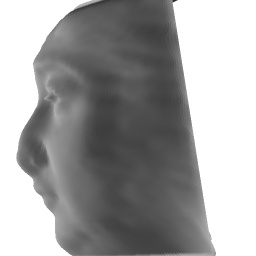}
  \includegraphics[trim={24px 0 0 0}, clip, height=1.5cm]{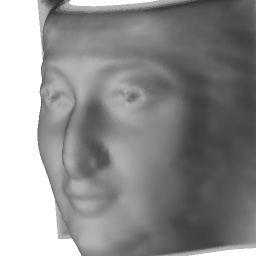}
  \includegraphics[height=1.5cm]{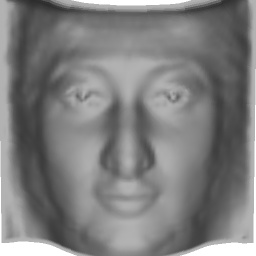}
  \includegraphics[height=1.5cm]{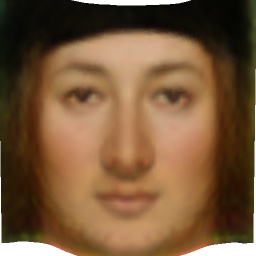}
  \includegraphics[trim={0 0 24px 0}, clip, height=1.5cm]{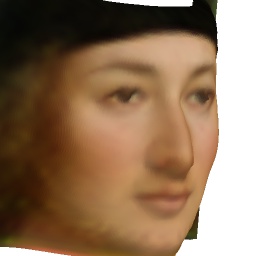}
  \includegraphics[trim={40px 0 0 0}, clip, height=1.5cm]{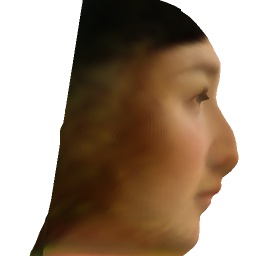}
\end{subfigure}
\\
\begin{subfigure}[b]{.2\linewidth}\centering
  \includegraphics[height=1.5cm]{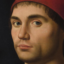}
\end{subfigure}
\begin{subfigure}[b]{.7\linewidth}\centering
  \includegraphics[height=1.5cm]{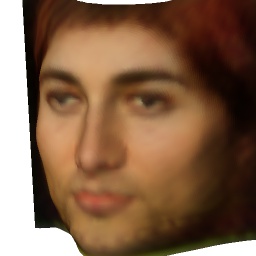}
  \includegraphics[height=1.5cm]{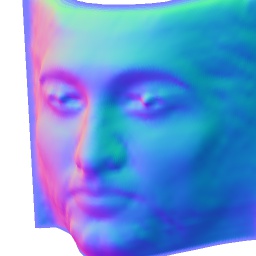}
  \includegraphics[trim={0 0 40px 0}, clip, height=1.5cm]{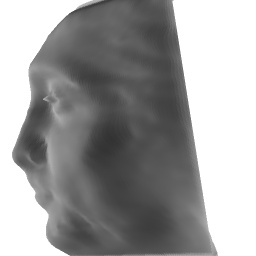}
  \includegraphics[trim={24px 0 0 0}, clip, height=1.5cm]{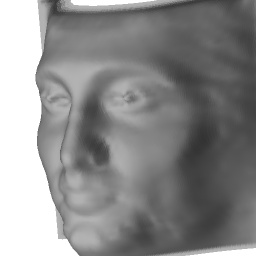}
  \includegraphics[height=1.5cm]{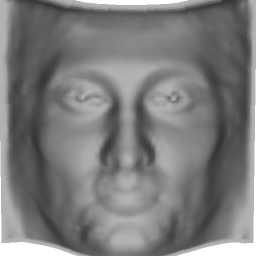}
  \includegraphics[height=1.5cm]{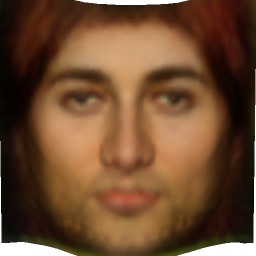}
  \includegraphics[trim={0 0 24px 0}, clip, height=1.5cm]{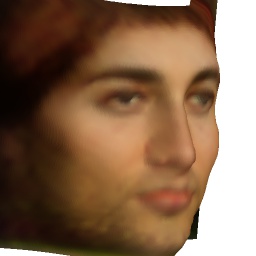}
  \includegraphics[trim={40px 0 0 0}, clip, height=1.5cm]{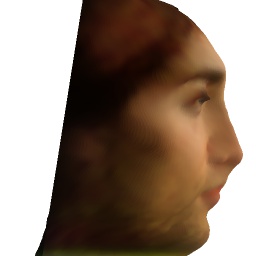}
\end{subfigure}
\\
\begin{subfigure}[b]{.2\linewidth}\centering
  \includegraphics[height=1.5cm]{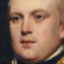}
\end{subfigure}
\begin{subfigure}[b]{.7\linewidth}\centering
  \includegraphics[height=1.5cm]{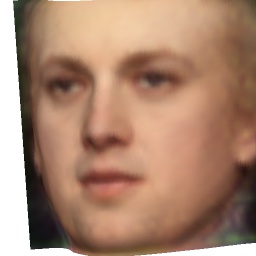}
  \includegraphics[height=1.5cm]{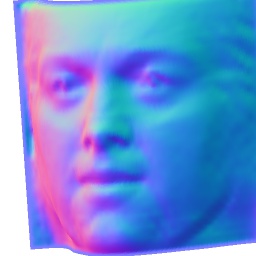}
  \includegraphics[trim={0 0 40px 0}, clip, height=1.5cm]{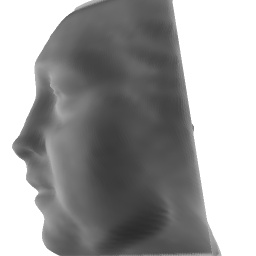}
  \includegraphics[trim={24px 0 0 0}, clip, height=1.5cm]{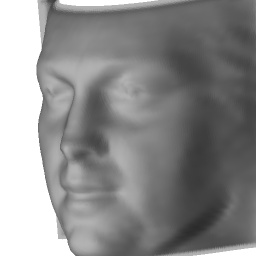}
  \includegraphics[height=1.5cm]{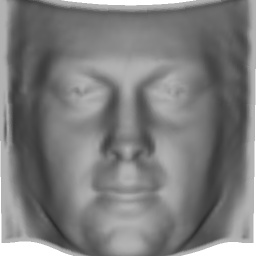}
  \includegraphics[height=1.5cm]{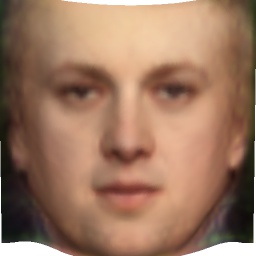}
  \includegraphics[trim={0 0 24px 0}, clip, height=1.5cm]{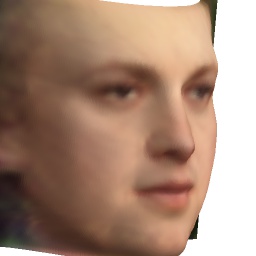}
  \includegraphics[trim={40px 0 0 0}, clip, height=1.5cm]{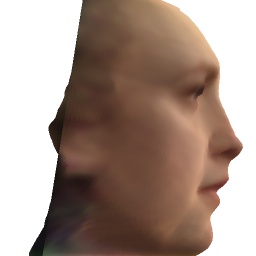}
\end{subfigure}
\\
\begin{subfigure}[b]{.2\linewidth}\centering
  \includegraphics[height=1.5cm]{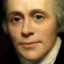}
\end{subfigure}
\begin{subfigure}[b]{.7\linewidth}\centering
  \includegraphics[height=1.5cm]{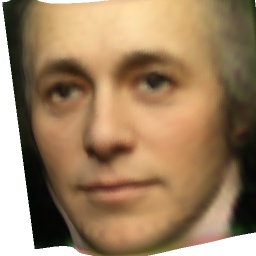}
  \includegraphics[height=1.5cm]{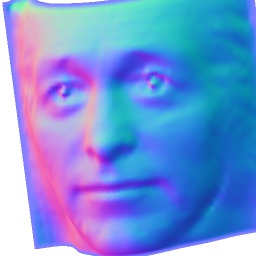}
  \includegraphics[trim={0 0 40px 0}, clip, height=1.5cm]{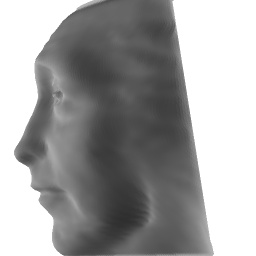}
  \includegraphics[trim={24px 0 0 0}, clip, height=1.5cm]{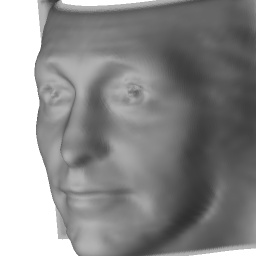}
  \includegraphics[height=1.5cm]{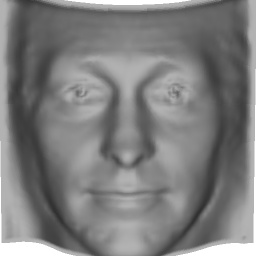}
  \includegraphics[height=1.5cm]{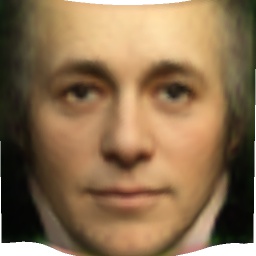}
  \includegraphics[trim={0 0 24px 0}, clip, height=1.5cm]{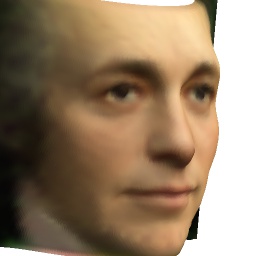}
  \includegraphics[trim={40px 0 0 0}, clip, height=1.5cm]{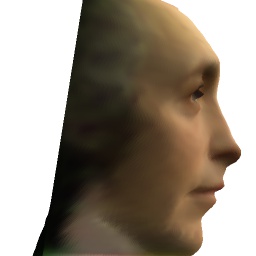}
\end{subfigure}
\\
\begin{subfigure}[b]{.2\linewidth}\centering
  \includegraphics[height=1.5cm]{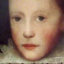}
\end{subfigure}
\begin{subfigure}[b]{.7\linewidth}\centering
  \includegraphics[height=1.5cm]{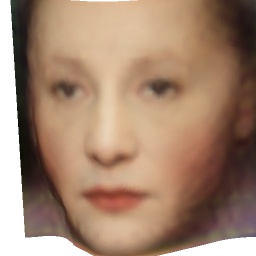}
  \includegraphics[height=1.5cm]{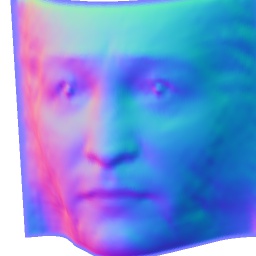}
  \includegraphics[trim={0 0 40px 0}, clip, height=1.5cm]{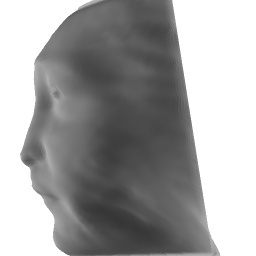}
  \includegraphics[trim={24px 0 0 0}, clip, height=1.5cm]{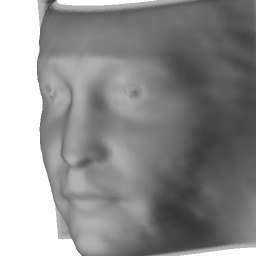}
  \includegraphics[height=1.5cm]{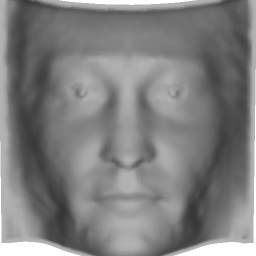}
  \includegraphics[height=1.5cm]{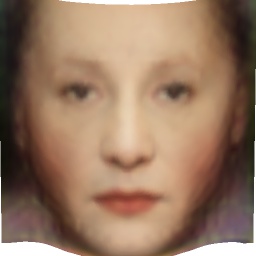}
  \includegraphics[trim={0 0 24px 0}, clip, height=1.5cm]{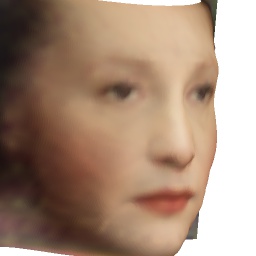}
  \includegraphics[trim={40px 0 0 0}, clip, height=1.5cm]{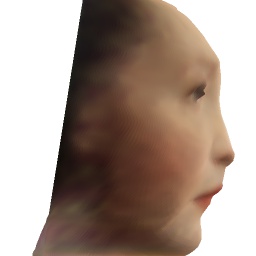}
\end{subfigure}
\\
\begin{subfigure}[b]{.2\linewidth}\centering
  \includegraphics[height=1.5cm]{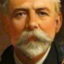}
\end{subfigure}
\begin{subfigure}[b]{.7\linewidth}\centering
  \includegraphics[height=1.5cm]{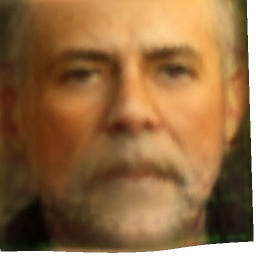}
  \includegraphics[height=1.5cm]{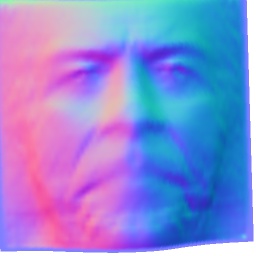}
  \includegraphics[trim={0 0 40px 0}, clip, height=1.5cm]{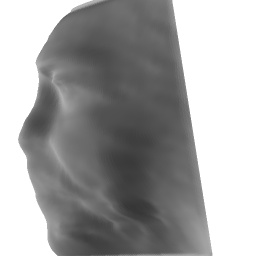}
  \includegraphics[trim={24px 0 0 0}, clip, height=1.5cm]{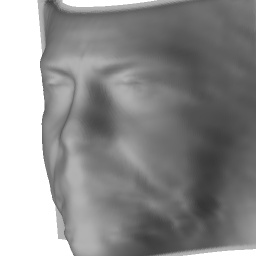}
  \includegraphics[height=1.5cm]{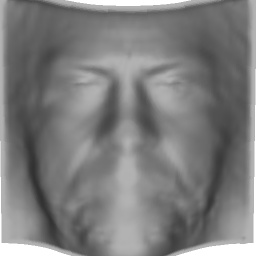}
  \includegraphics[height=1.5cm]{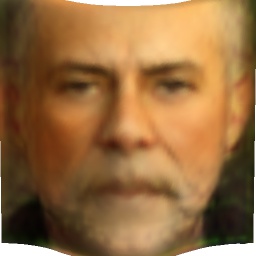}
  \includegraphics[trim={0 0 24px 0}, clip, height=1.5cm]{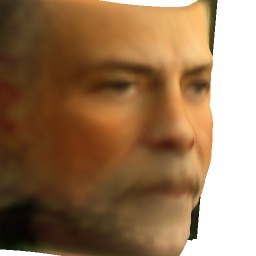}
  \includegraphics[trim={40px 0 0 0}, clip, height=1.5cm]{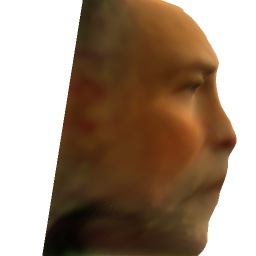}
\end{subfigure}
\\
\begin{subfigure}[b]{.2\linewidth}\centering
  \includegraphics[height=1.5cm]{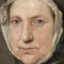}
\end{subfigure}
\begin{subfigure}[b]{.7\linewidth}\centering
  \includegraphics[height=1.5cm]{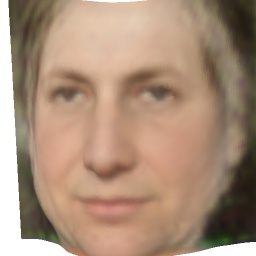}
  \includegraphics[height=1.5cm]{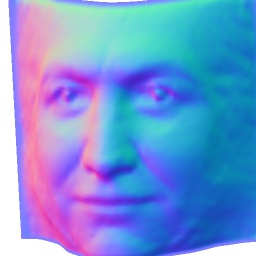}
  \includegraphics[trim={0 0 40px 0}, clip, height=1.5cm]{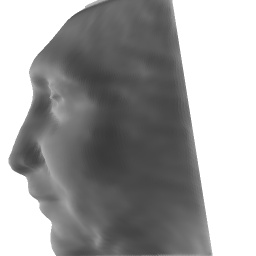}
  \includegraphics[trim={24px 0 0 0}, clip, height=1.5cm]{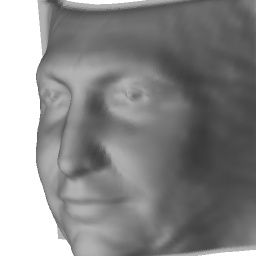}
  \includegraphics[height=1.5cm]{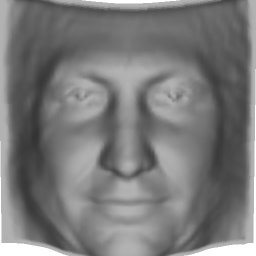}
  \includegraphics[height=1.5cm]{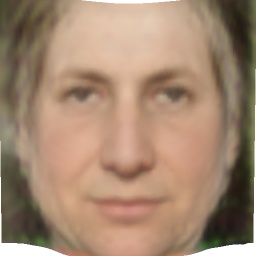}
  \includegraphics[trim={0 0 24px 0}, clip, height=1.5cm]{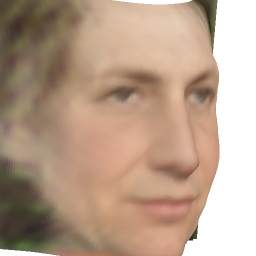}
  \includegraphics[trim={40px 0 0 0}, clip, height=1.5cm]{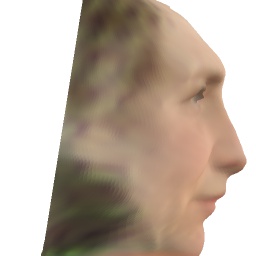}
\end{subfigure}
\\
\begin{subfigure}[b]{.2\linewidth}\centering
  \includegraphics[height=1.5cm]{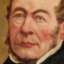}
\end{subfigure}
\begin{subfigure}[b]{.7\linewidth}\centering
  \includegraphics[height=1.5cm]{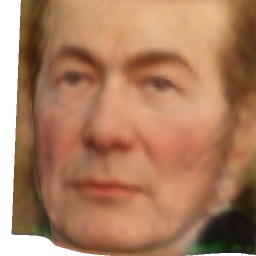}
  \includegraphics[height=1.5cm]{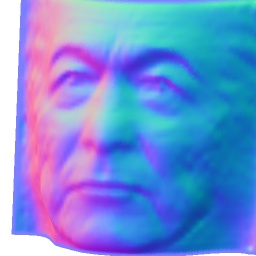}
  \includegraphics[trim={0 0 40px 0}, clip, height=1.5cm]{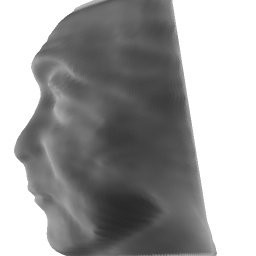}
  \includegraphics[trim={24px 0 0 0}, clip, height=1.5cm]{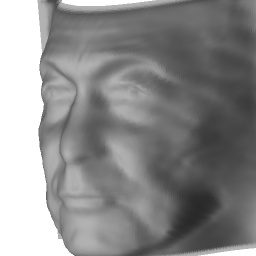}
  \includegraphics[height=1.5cm]{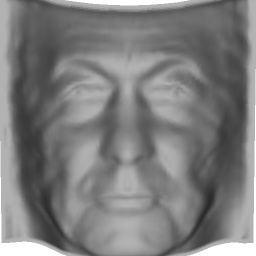}
  \includegraphics[height=1.5cm]{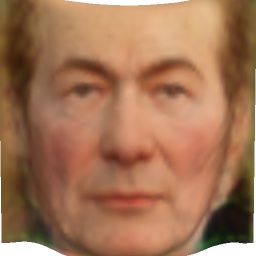}
  \includegraphics[trim={0 0 24px 0}, clip, height=1.5cm]{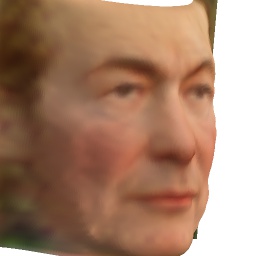}
  \includegraphics[trim={40px 0 0 0}, clip, height=1.5cm]{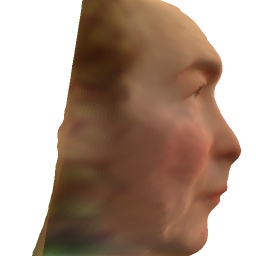}
\end{subfigure}
\\
\begin{subfigure}[b]{.2\linewidth}\centering
  \includegraphics[height=1.5cm]{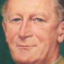}
\end{subfigure}
\begin{subfigure}[b]{.7\linewidth}\centering
  \includegraphics[height=1.5cm]{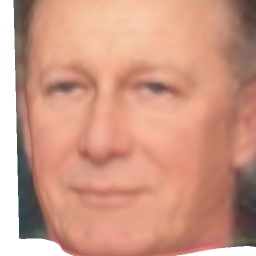}
  \includegraphics[height=1.5cm]{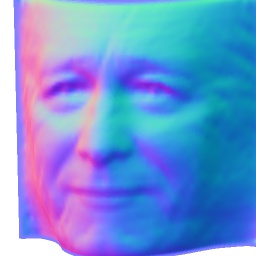}
  \includegraphics[trim={0 0 40px 0}, clip, height=1.5cm]{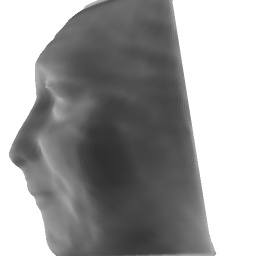}
  \includegraphics[trim={24px 0 0 0}, clip, height=1.5cm]{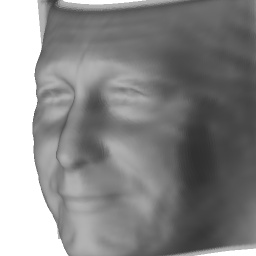}
  \includegraphics[height=1.5cm]{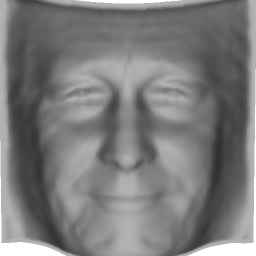}
  \includegraphics[height=1.5cm]{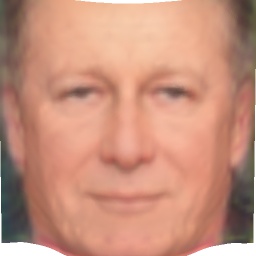}
  \includegraphics[trim={0 0 24px 0}, clip, height=1.5cm]{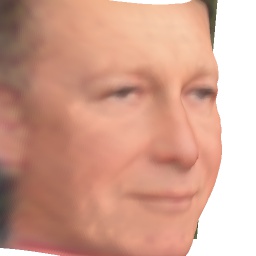}
  \includegraphics[trim={40px 0 0 0}, clip, height=1.5cm]{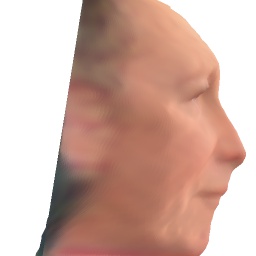}
\end{subfigure}
\\
\begin{subfigure}[b]{.2\linewidth}\centering
  \includegraphics[height=1.5cm]{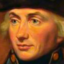}
\end{subfigure}
\begin{subfigure}[b]{.7\linewidth}\centering
  \includegraphics[height=1.5cm]{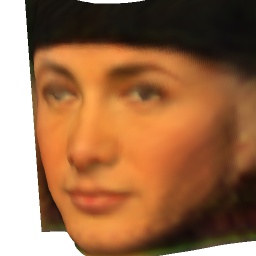}
  \includegraphics[height=1.5cm]{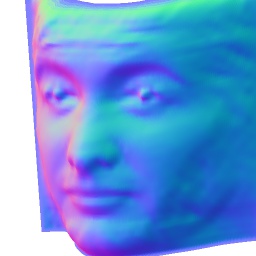}
  \includegraphics[trim={0 0 40px 0}, clip, height=1.5cm]{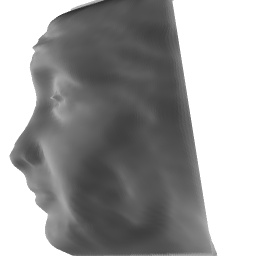}
  \includegraphics[trim={24px 0 0 0}, clip, height=1.5cm]{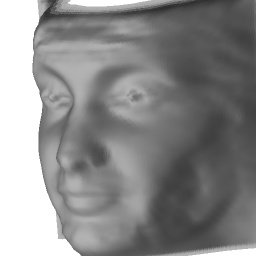}
  \includegraphics[height=1.5cm]{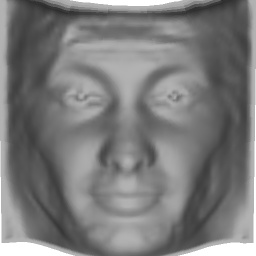}
  \includegraphics[height=1.5cm]{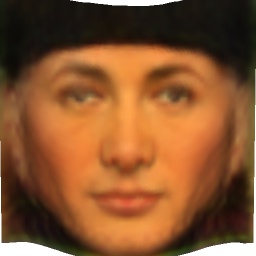}
  \includegraphics[trim={0 0 24px 0}, clip, height=1.5cm]{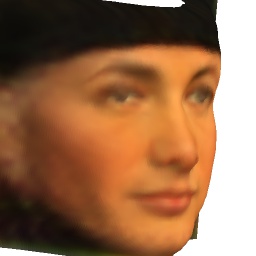}
  \includegraphics[trim={40px 0 0 0}, clip, height=1.5cm]{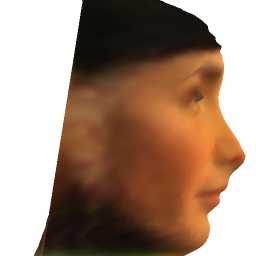}
\end{subfigure}
\\
\begin{subfigure}[b]{.2\linewidth}\centering
  \includegraphics[height=1.5cm]{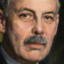}
\end{subfigure}
\begin{subfigure}[b]{.7\linewidth}\centering
  \includegraphics[height=1.5cm]{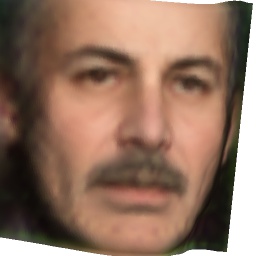}
  \includegraphics[height=1.5cm]{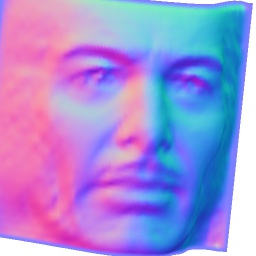}
  \includegraphics[trim={0 0 40px 0}, clip, height=1.5cm]{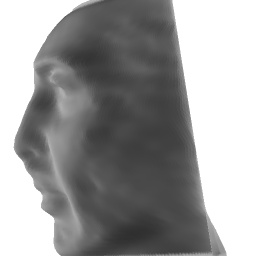}
  \includegraphics[trim={24px 0 0 0}, clip, height=1.5cm]{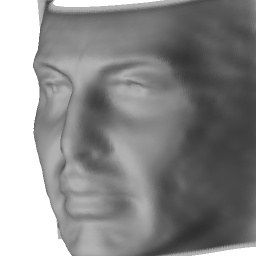}
  \includegraphics[height=1.5cm]{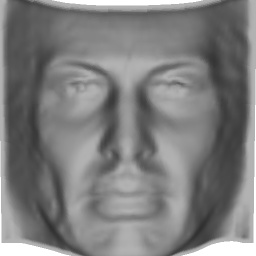}
  \includegraphics[height=1.5cm]{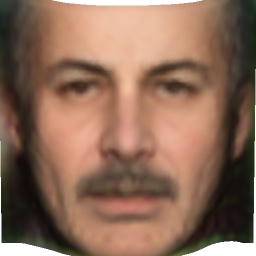}
  \includegraphics[trim={0 0 24px 0}, clip, height=1.5cm]{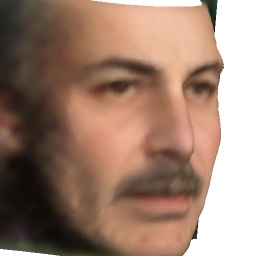}
  \includegraphics[trim={40px 0 0 0}, clip, height=1.5cm]{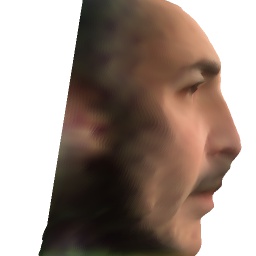}
\end{subfigure}
\\
\begin{subfigure}[b]{.2\linewidth}\centering
  \includegraphics[height=1.5cm]{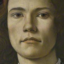}
\end{subfigure}
\begin{subfigure}[b]{.7\linewidth}\centering
  \includegraphics[height=1.5cm]{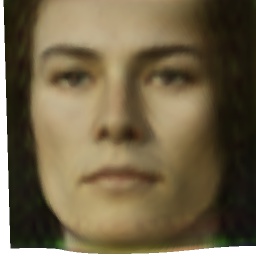}
  \includegraphics[height=1.5cm]{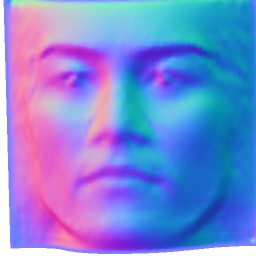}
  \includegraphics[trim={0 0 40px 0}, clip, height=1.5cm]{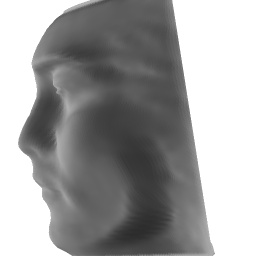}
  \includegraphics[trim={24px 0 0 0}, clip, height=1.5cm]{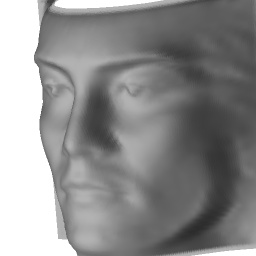}
  \includegraphics[height=1.5cm]{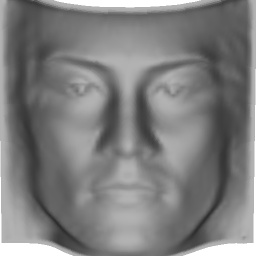}
  \includegraphics[height=1.5cm]{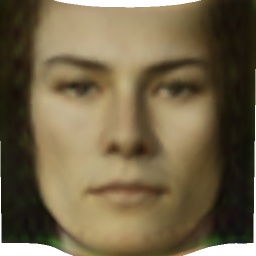}
  \includegraphics[trim={0 0 24px 0}, clip, height=1.5cm]{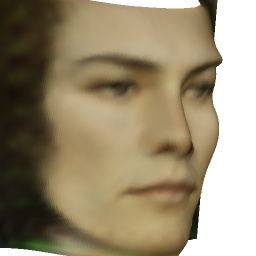}
  \includegraphics[trim={40px 0 0 0}, clip, height=1.5cm]{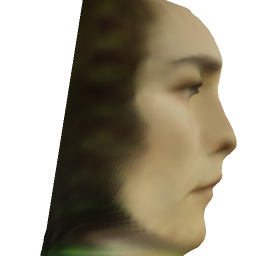}
\end{subfigure}
\\
\begin{subfigure}[b]{.2\linewidth}\centering
  \includegraphics[height=1.5cm]{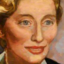}
\end{subfigure}
\begin{subfigure}[b]{.7\linewidth}\centering
  \includegraphics[height=1.5cm]{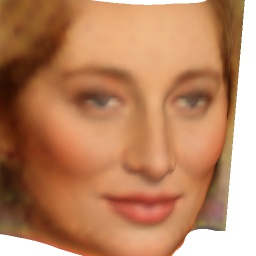}
  \includegraphics[height=1.5cm]{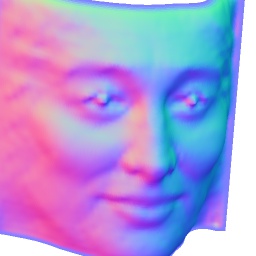}
  \includegraphics[trim={0 0 40px 0}, clip, height=1.5cm]{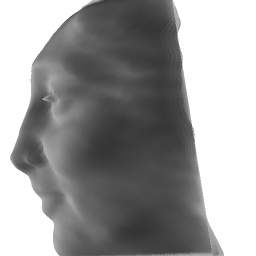}
  \includegraphics[trim={24px 0 0 0}, clip, height=1.5cm]{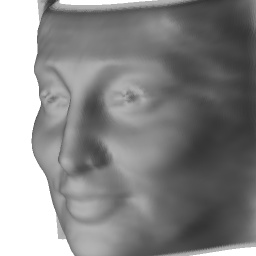}
  \includegraphics[height=1.5cm]{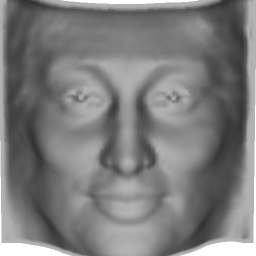}
  \includegraphics[height=1.5cm]{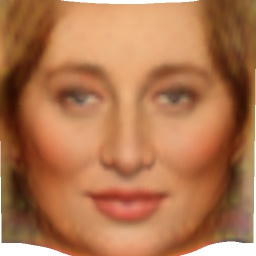}
  \includegraphics[trim={0 0 24px 0}, clip, height=1.5cm]{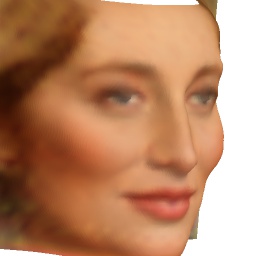}
  \includegraphics[trim={40px 0 0 0}, clip, height=1.5cm]{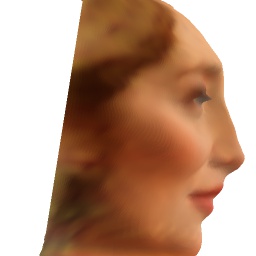}
\end{subfigure}
\\
\begin{subfigure}[b]{.2\linewidth}\centering
  \includegraphics[height=1.5cm]{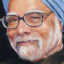}
  \caption{Input}
\end{subfigure}
\begin{subfigure}[b]{.7\linewidth}\centering
  \includegraphics[height=1.5cm]{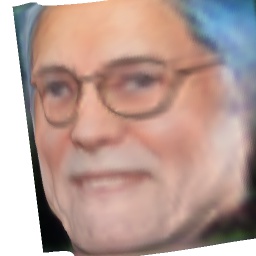}
  \includegraphics[height=1.5cm]{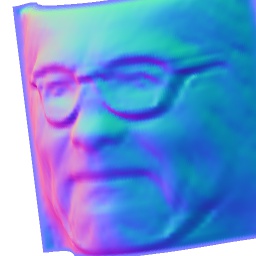}
  \includegraphics[trim={0 0 40px 0}, clip, height=1.5cm]{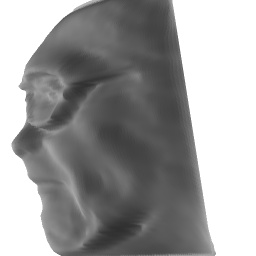}
  \includegraphics[trim={24px 0 0 0}, clip, height=1.5cm]{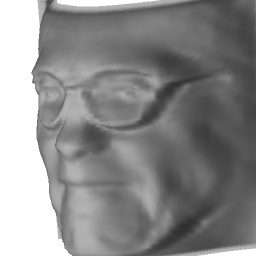}
  \includegraphics[height=1.5cm]{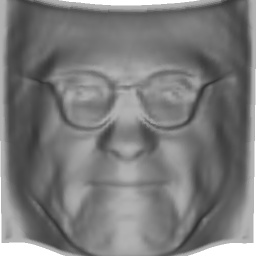}
  \includegraphics[height=1.5cm]{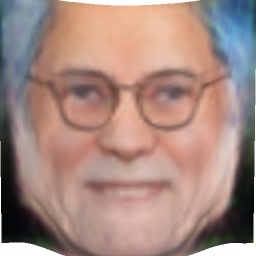}
  \includegraphics[trim={0 0 24px 0}, clip, height=1.5cm]{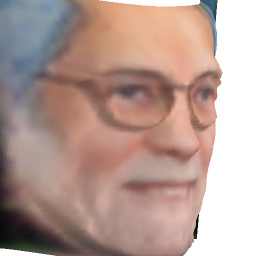}
  \includegraphics[trim={40px 0 0 0}, clip, height=1.5cm]{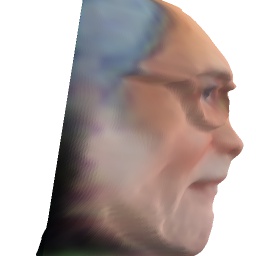}
  \caption{Reconstruction}
\end{subfigure}

\caption{\textbf{Reconstruction of face paintings.}}\label{fig:sup-paint}
\end{figure*}

%% file: supmat/fig-abstract-face.tex
\begin{figure*}[t]\centering
\captionsetup[subfigure]{justification=centering,labelformat=empty,labelsep=colon,aboveskip=2pt}
\begin{subfigure}[b]{.2\linewidth}\centering
  \includegraphics[height=1.5cm]{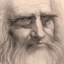}
\end{subfigure}
\begin{subfigure}[b]{.7\linewidth}\centering
  \includegraphics[height=1.5cm]{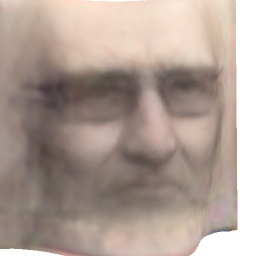}
  \includegraphics[height=1.5cm]{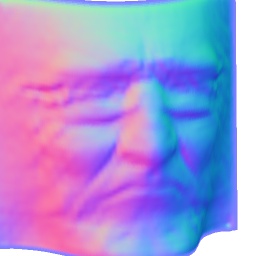}
  \includegraphics[trim={0 0 40px 0}, clip, height=1.5cm]{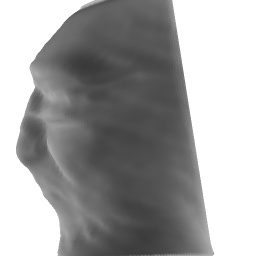}
  \includegraphics[trim={24px 0 0 0}, clip, height=1.5cm]{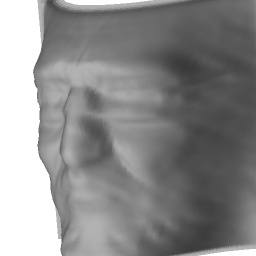}
  \includegraphics[height=1.5cm]{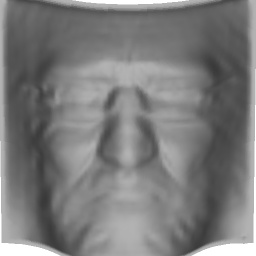}
  \includegraphics[height=1.5cm]{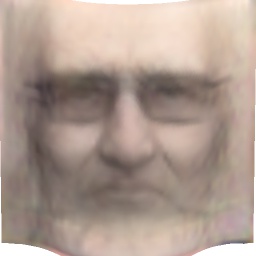}
  \includegraphics[trim={0 0 24px 0}, clip, height=1.5cm]{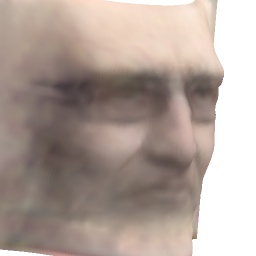}
  \includegraphics[trim={40px 0 0 0}, clip, height=1.5cm]{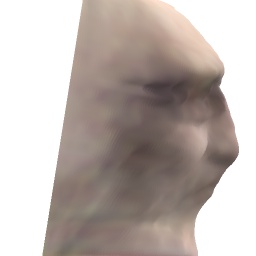}
\end{subfigure}
\\
\begin{subfigure}[b]{.2\linewidth}\centering
  \includegraphics[height=1.5cm]{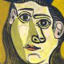}
\end{subfigure}
\begin{subfigure}[b]{.7\linewidth}\centering
  \includegraphics[height=1.5cm]{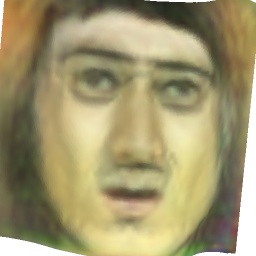}
  \includegraphics[height=1.5cm]{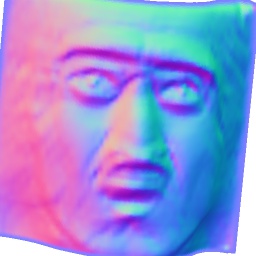}
  \includegraphics[trim={0 0 40px 0}, clip, height=1.5cm]{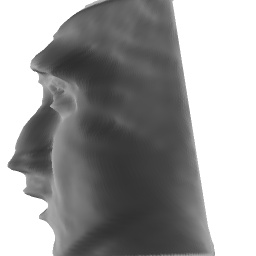}
  \includegraphics[trim={24px 0 0 0}, clip, height=1.5cm]{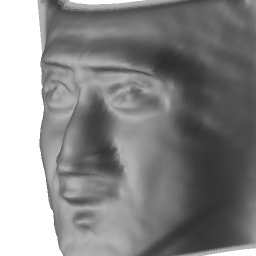}
  \includegraphics[height=1.5cm]{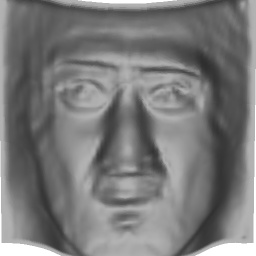}
  \includegraphics[height=1.5cm]{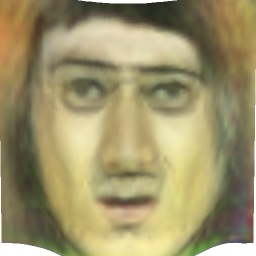}
  \includegraphics[trim={0 0 24px 0}, clip, height=1.5cm]{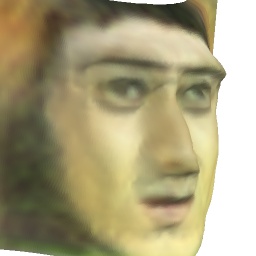}
  \includegraphics[trim={40px 0 0 0}, clip, height=1.5cm]{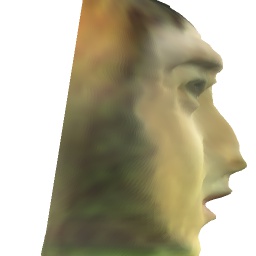}
\end{subfigure}
\\
\begin{subfigure}[b]{.2\linewidth}\centering
  \includegraphics[height=1.5cm]{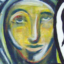}
\end{subfigure}
\begin{subfigure}[b]{.7\linewidth}\centering
  \includegraphics[height=1.5cm]{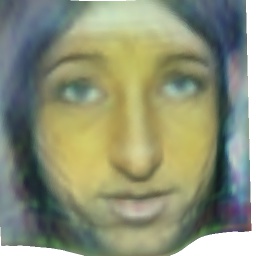}
  \includegraphics[height=1.5cm]{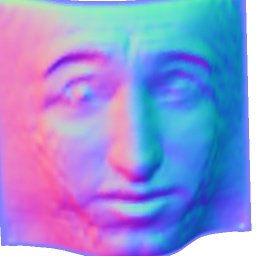}
  \includegraphics[trim={0 0 40px 0}, clip, height=1.5cm]{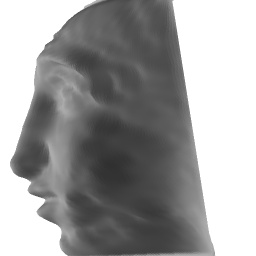}
  \includegraphics[trim={24px 0 0 0}, clip, height=1.5cm]{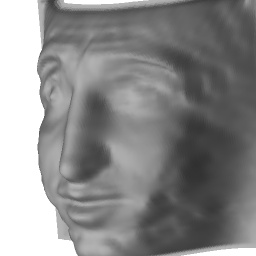}
  \includegraphics[height=1.5cm]{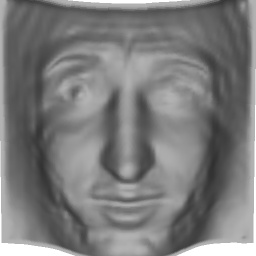}
  \includegraphics[height=1.5cm]{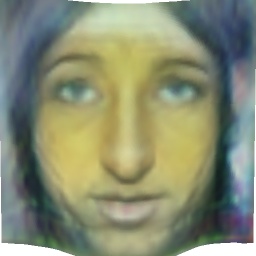}
  \includegraphics[trim={0 0 24px 0}, clip, height=1.5cm]{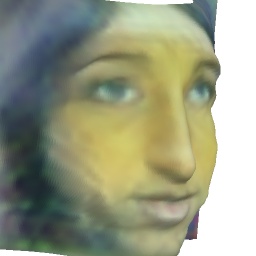}
  \includegraphics[trim={40px 0 0 0}, clip, height=1.5cm]{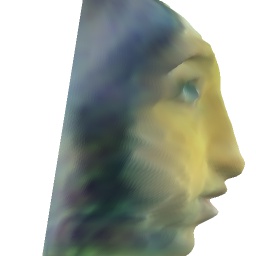}
\end{subfigure}
\\
\begin{subfigure}[b]{.2\linewidth}\centering
  \includegraphics[height=1.5cm]{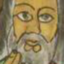}
\end{subfigure}
\begin{subfigure}[b]{.7\linewidth}\centering
  \includegraphics[height=1.5cm]{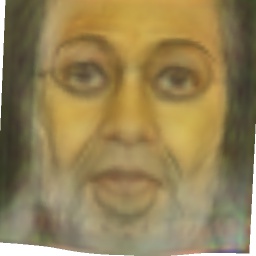}
  \includegraphics[height=1.5cm]{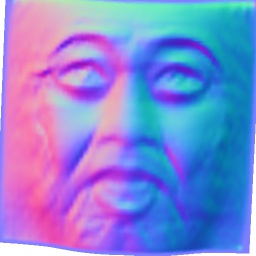}
  \includegraphics[trim={0 0 40px 0}, clip, height=1.5cm]{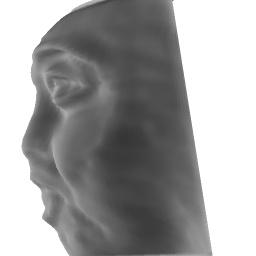}
  \includegraphics[trim={24px 0 0 0}, clip, height=1.5cm]{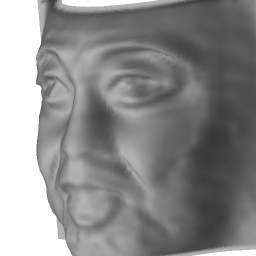}
  \includegraphics[height=1.5cm]{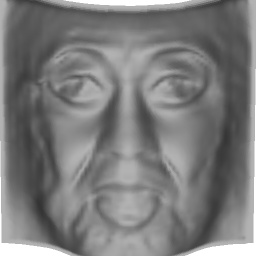}
  \includegraphics[height=1.5cm]{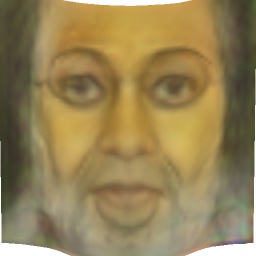}
  \includegraphics[trim={0 0 24px 0}, clip, height=1.5cm]{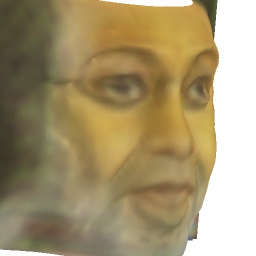}
  \includegraphics[trim={40px 0 0 0}, clip, height=1.5cm]{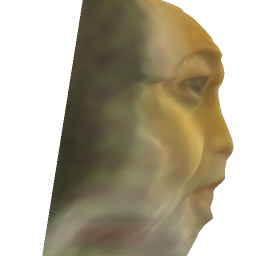}
\end{subfigure}
\\
\begin{subfigure}[b]{.2\linewidth}\centering
  \includegraphics[height=1.5cm]{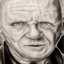}
\end{subfigure}
\begin{subfigure}[b]{.7\linewidth}\centering
  \includegraphics[height=1.5cm]{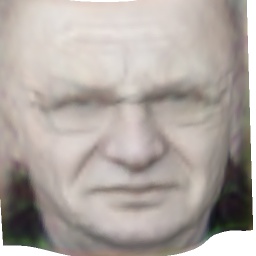}
  \includegraphics[height=1.5cm]{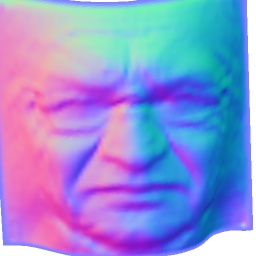}
  \includegraphics[trim={0 0 40px 0}, clip, height=1.5cm]{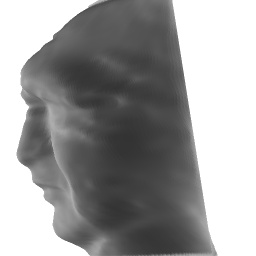}
  \includegraphics[trim={24px 0 0 0}, clip, height=1.5cm]{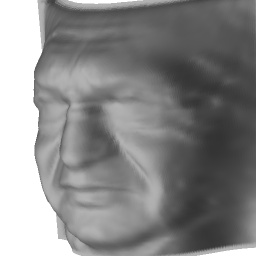}
  \includegraphics[height=1.5cm]{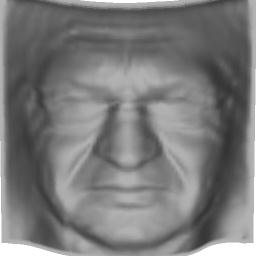}
  \includegraphics[height=1.5cm]{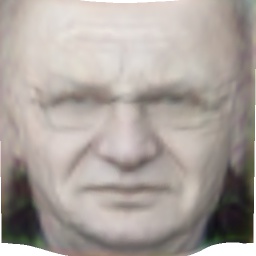}
  \includegraphics[trim={0 0 24px 0}, clip, height=1.5cm]{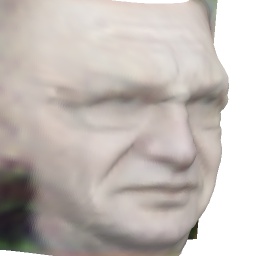}
  \includegraphics[trim={40px 0 0 0}, clip, height=1.5cm]{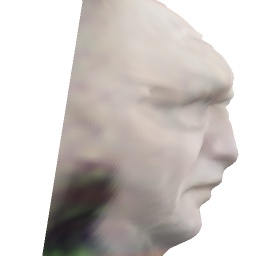}
\end{subfigure}
\\
\begin{subfigure}[b]{.2\linewidth}\centering
  \includegraphics[height=1.5cm]{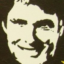}
\end{subfigure}
\begin{subfigure}[b]{.7\linewidth}\centering
  \includegraphics[height=1.5cm]{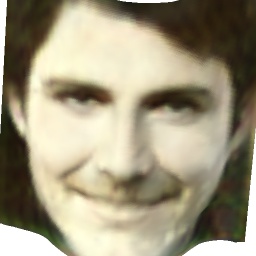}
  \includegraphics[height=1.5cm]{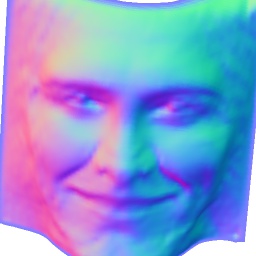}
  \includegraphics[trim={0 0 40px 0}, clip, height=1.5cm]{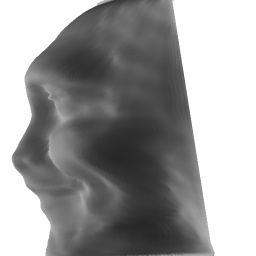}
  \includegraphics[trim={24px 0 0 0}, clip, height=1.5cm]{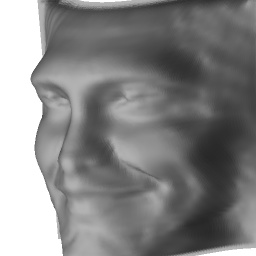}
  \includegraphics[height=1.5cm]{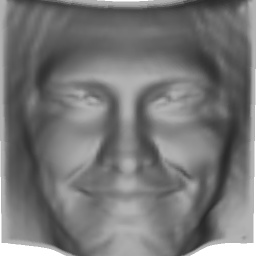}
  \includegraphics[height=1.5cm]{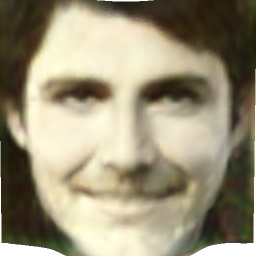}
  \includegraphics[trim={0 0 24px 0}, clip, height=1.5cm]{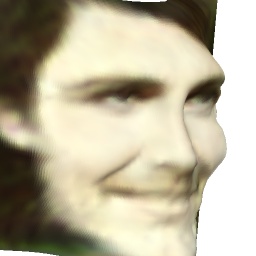}
  \includegraphics[trim={40px 0 0 0}, clip, height=1.5cm]{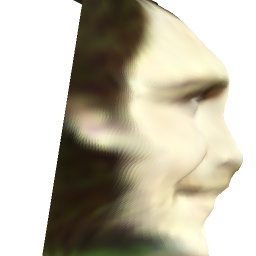}
\end{subfigure}
\\
\begin{subfigure}[b]{.2\linewidth}\centering
  \includegraphics[height=1.5cm]{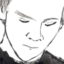}
\end{subfigure}
\begin{subfigure}[b]{.7\linewidth}\centering
  \includegraphics[height=1.5cm]{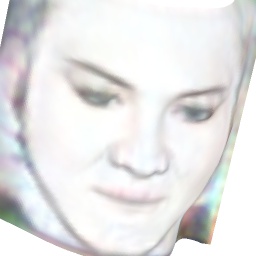}
  \includegraphics[height=1.5cm]{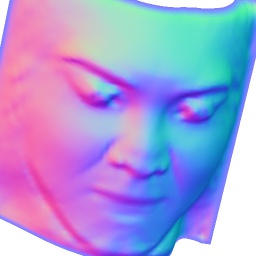}
  \includegraphics[trim={0 0 40px 0}, clip, height=1.5cm]{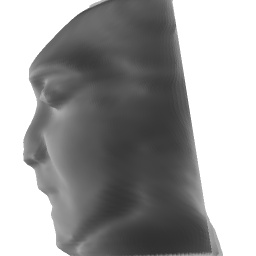}
  \includegraphics[trim={24px 0 0 0}, clip, height=1.5cm]{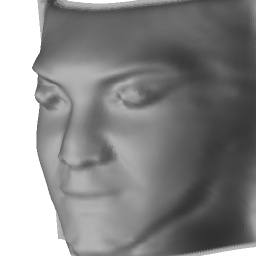}
  \includegraphics[height=1.5cm]{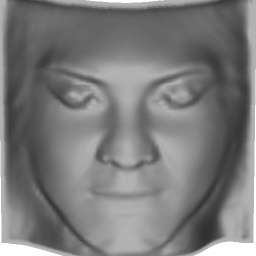}
  \includegraphics[height=1.5cm]{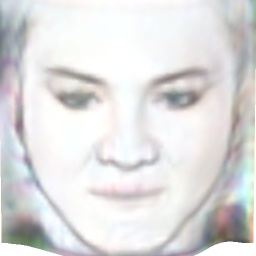}
  \includegraphics[trim={0 0 24px 0}, clip, height=1.5cm]{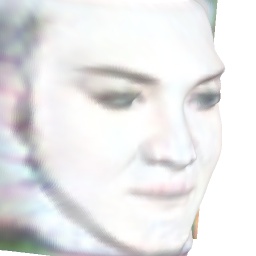}
  \includegraphics[trim={40px 0 0 0}, clip, height=1.5cm]{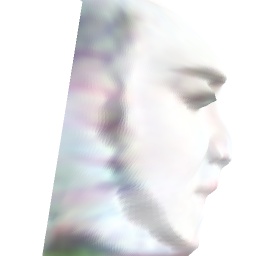}
\end{subfigure}
\\
\begin{subfigure}[b]{.2\linewidth}\centering
  \includegraphics[height=1.5cm]{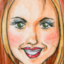}
\end{subfigure}
\begin{subfigure}[b]{.7\linewidth}\centering
  \includegraphics[height=1.5cm]{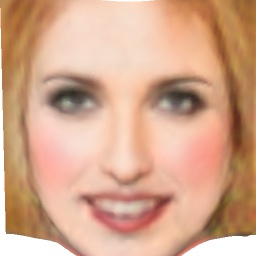}
  \includegraphics[height=1.5cm]{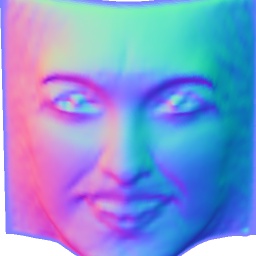}
  \includegraphics[trim={0 0 40px 0}, clip, height=1.5cm]{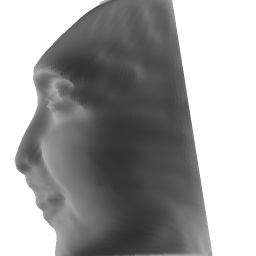}
  \includegraphics[trim={24px 0 0 0}, clip, height=1.5cm]{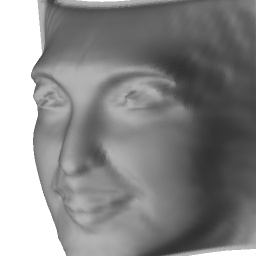}
  \includegraphics[height=1.5cm]{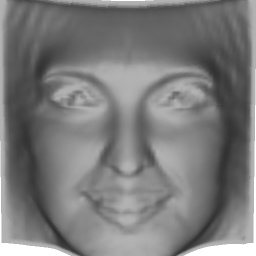}
  \includegraphics[height=1.5cm]{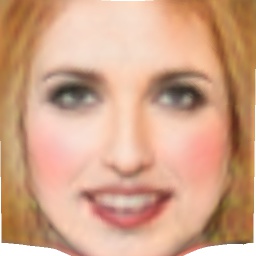}
  \includegraphics[trim={0 0 24px 0}, clip, height=1.5cm]{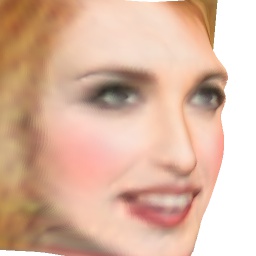}
  \includegraphics[trim={40px 0 0 0}, clip, height=1.5cm]{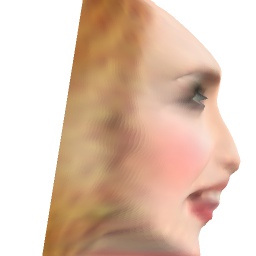}
\end{subfigure}
\\
\begin{subfigure}[b]{.2\linewidth}\centering
  \includegraphics[height=1.5cm]{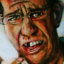}
\end{subfigure}
\begin{subfigure}[b]{.7\linewidth}\centering
  \includegraphics[height=1.5cm]{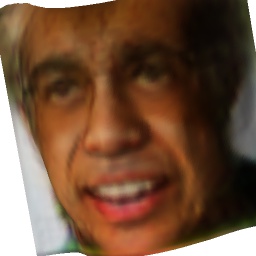}
  \includegraphics[height=1.5cm]{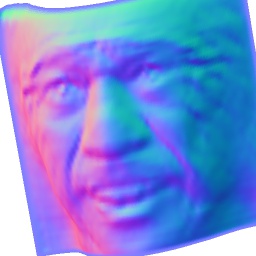}
  \includegraphics[trim={0 0 40px 0}, clip, height=1.5cm]{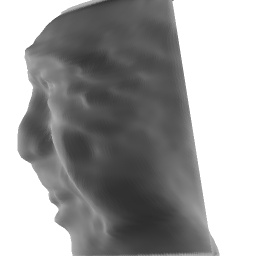}
  \includegraphics[trim={24px 0 0 0}, clip, height=1.5cm]{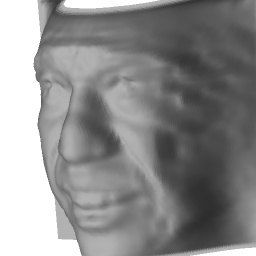}
  \includegraphics[height=1.5cm]{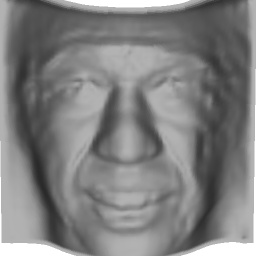}
  \includegraphics[height=1.5cm]{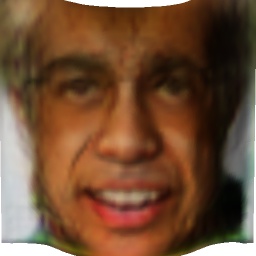}
  \includegraphics[trim={0 0 24px 0}, clip, height=1.5cm]{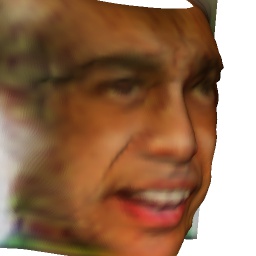}
  \includegraphics[trim={40px 0 0 0}, clip, height=1.5cm]{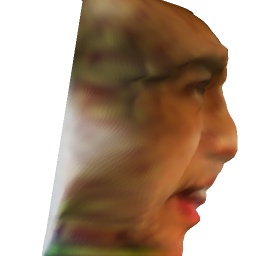}
\end{subfigure}
\\
\begin{subfigure}[b]{.2\linewidth}\centering
  \includegraphics[height=1.5cm]{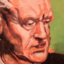}
\end{subfigure}
\begin{subfigure}[b]{.7\linewidth}\centering
  \includegraphics[height=1.5cm]{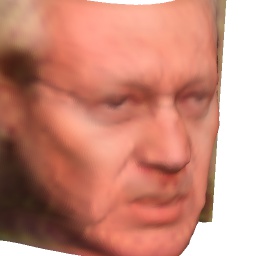}
  \includegraphics[height=1.5cm]{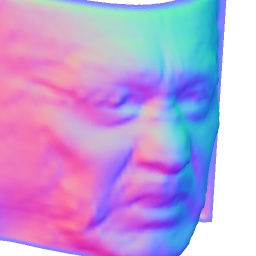}
  \includegraphics[trim={0 0 40px 0}, clip, height=1.5cm]{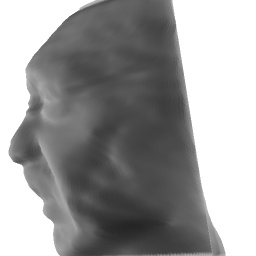}
  \includegraphics[trim={24px 0 0 0}, clip, height=1.5cm]{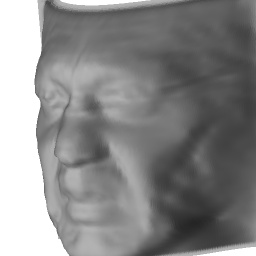}
  \includegraphics[height=1.5cm]{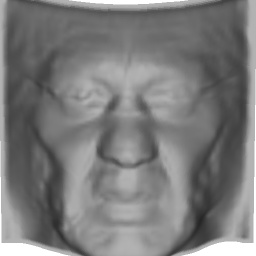}
  \includegraphics[height=1.5cm]{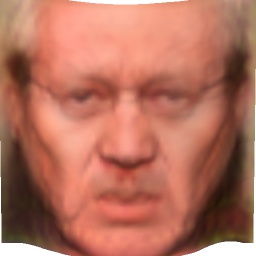}
  \includegraphics[trim={0 0 24px 0}, clip, height=1.5cm]{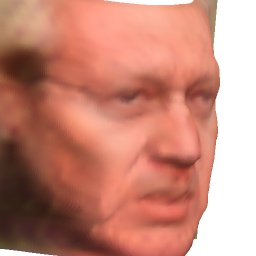}
  \includegraphics[trim={40px 0 0 0}, clip, height=1.5cm]{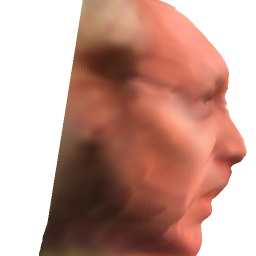}
\end{subfigure}
\\
\begin{subfigure}[b]{.2\linewidth}\centering
  \includegraphics[height=1.5cm]{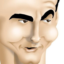}
\end{subfigure}
\begin{subfigure}[b]{.7\linewidth}\centering
  \includegraphics[height=1.5cm]{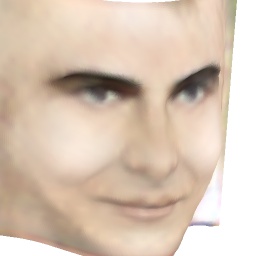}
  \includegraphics[height=1.5cm]{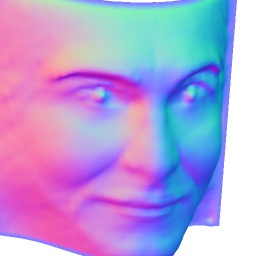}
  \includegraphics[trim={0 0 40px 0}, clip, height=1.5cm]{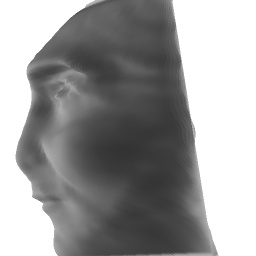}
  \includegraphics[trim={24px 0 0 0}, clip, height=1.5cm]{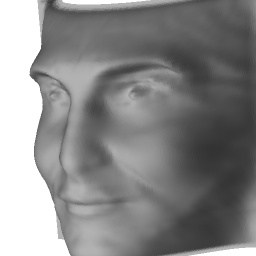}
  \includegraphics[height=1.5cm]{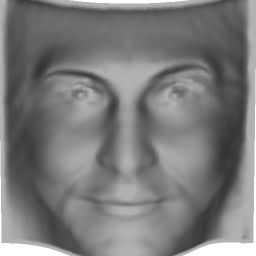}
  \includegraphics[height=1.5cm]{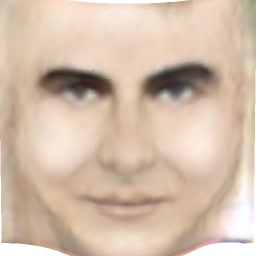}
  \includegraphics[trim={0 0 24px 0}, clip, height=1.5cm]{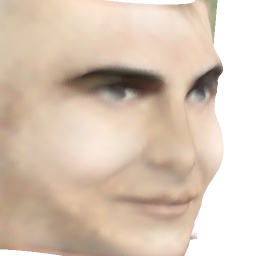}
  \includegraphics[trim={40px 0 0 0}, clip, height=1.5cm]{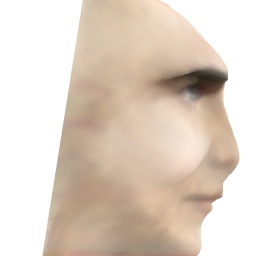}
\end{subfigure}
\\
\begin{subfigure}[b]{.2\linewidth}\centering
  \includegraphics[height=1.5cm]{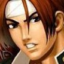}
\end{subfigure}
\begin{subfigure}[b]{.7\linewidth}\centering
  \includegraphics[height=1.5cm]{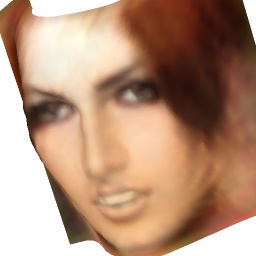}
  \includegraphics[height=1.5cm]{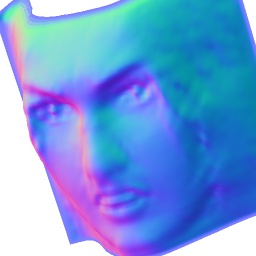}
  \includegraphics[trim={0 0 40px 0}, clip, height=1.5cm]{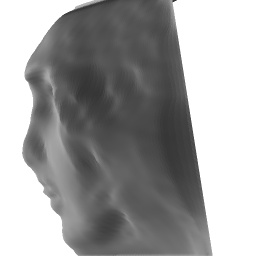}
  \includegraphics[trim={24px 0 0 0}, clip, height=1.5cm]{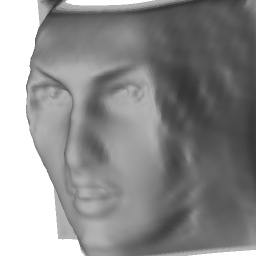}
  \includegraphics[height=1.5cm]{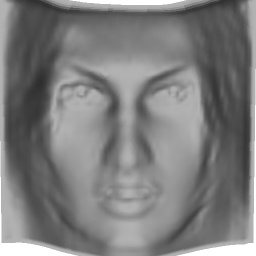}
  \includegraphics[height=1.5cm]{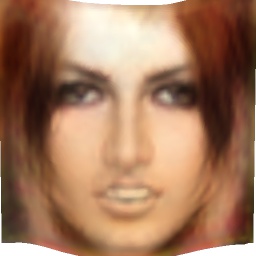}
  \includegraphics[trim={0 0 24px 0}, clip, height=1.5cm]{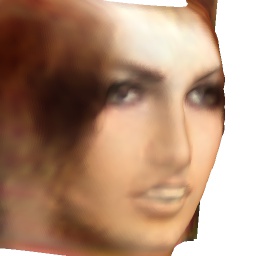}
  \includegraphics[trim={40px 0 0 0}, clip, height=1.5cm]{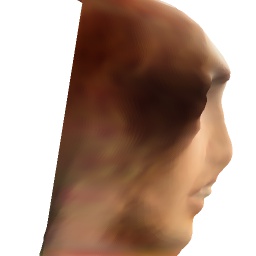}
\end{subfigure}
\\
\begin{subfigure}[b]{.2\linewidth}\centering
  \includegraphics[height=1.5cm]{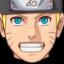}
\end{subfigure}
\begin{subfigure}[b]{.7\linewidth}\centering
  \includegraphics[height=1.5cm]{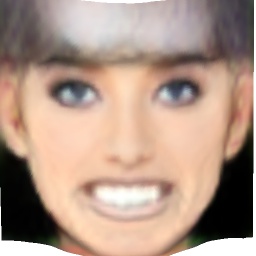}
  \includegraphics[height=1.5cm]{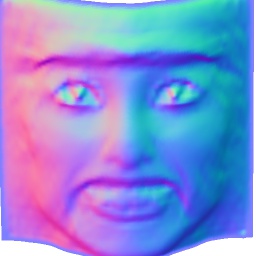}
  \includegraphics[trim={0 0 40px 0}, clip, height=1.5cm]{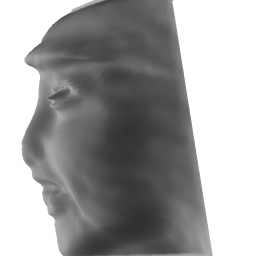}
  \includegraphics[trim={24px 0 0 0}, clip, height=1.5cm]{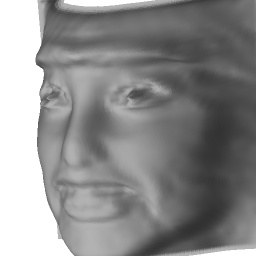}
  \includegraphics[height=1.5cm]{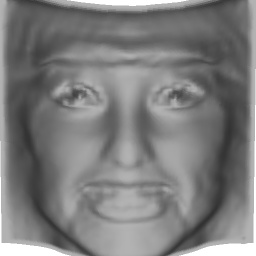}
  \includegraphics[height=1.5cm]{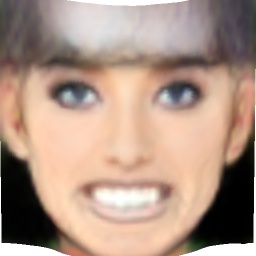}
  \includegraphics[trim={0 0 24px 0}, clip, height=1.5cm]{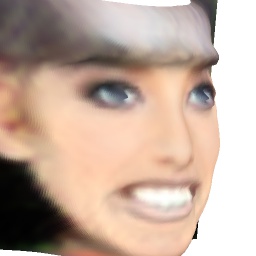}
  \includegraphics[trim={40px 0 0 0}, clip, height=1.5cm]{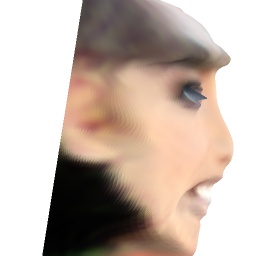}
\end{subfigure}
\\
\begin{subfigure}[b]{.2\linewidth}\centering
  \includegraphics[height=1.5cm]{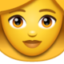}
  \caption{Input}
\end{subfigure}
\begin{subfigure}[b]{.7\linewidth}\centering
  \includegraphics[height=1.5cm]{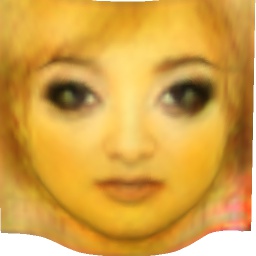}
  \includegraphics[height=1.5cm]{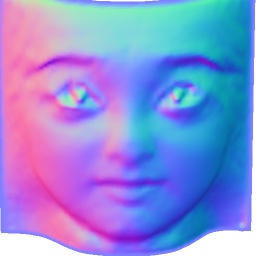}
  \includegraphics[trim={0 0 40px 0}, clip, height=1.5cm]{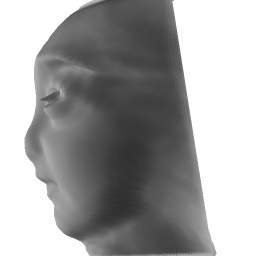}
  \includegraphics[trim={24px 0 0 0}, clip, height=1.5cm]{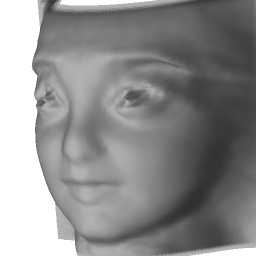}
  \includegraphics[height=1.5cm]{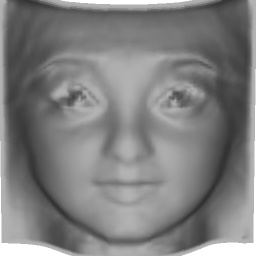}
  \includegraphics[height=1.5cm]{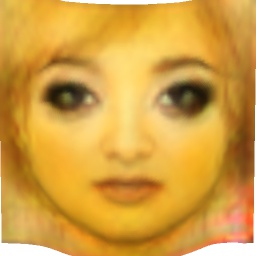}
  \includegraphics[trim={0 0 24px 0}, clip, height=1.5cm]{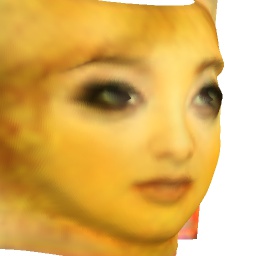}
  \includegraphics[trim={40px 0 0 0}, clip, height=1.5cm]{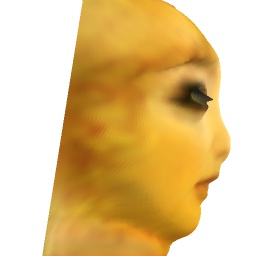}
  \caption{Reconstruction}
\end{subfigure}

\caption{\textbf{Reconstruction of abstract faces.}}\label{fig:sup-abstract-face}
\end{figure*}

%% file: supmat/fig-cat.tex
\begin{figure*}[t]\centering
\captionsetup[subfigure]{justification=centering,labelformat=empty,labelsep=colon,aboveskip=2pt}
\begin{subfigure}[b]{.2\linewidth}\centering
  \includegraphics[height=1.5cm]{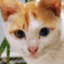}
\end{subfigure}
\begin{subfigure}[b]{.7\linewidth}\centering
  \includegraphics[height=1.5cm]{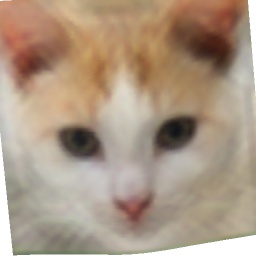}
  \includegraphics[height=1.5cm]{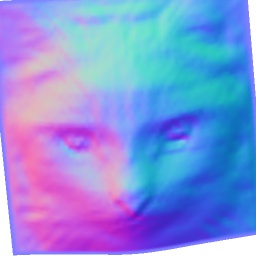}
  \includegraphics[trim={0 0 40px 0}, clip, height=1.5cm]{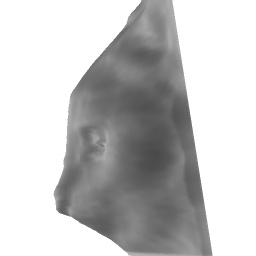}
  \includegraphics[trim={24px 0 0 0}, clip, height=1.5cm]{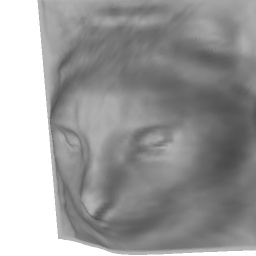}
  \includegraphics[height=1.5cm]{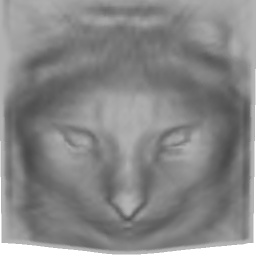}
  \includegraphics[height=1.5cm]{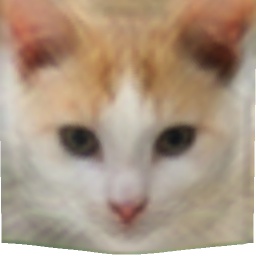}
  \includegraphics[trim={0 0 24px 0}, clip, height=1.5cm]{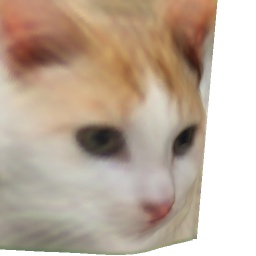}
  \includegraphics[trim={40px 0 0 0}, clip, height=1.5cm]{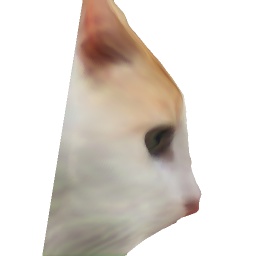}
\end{subfigure}
\\
\begin{subfigure}[b]{.2\linewidth}\centering
  \includegraphics[height=1.5cm]{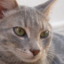}
\end{subfigure}
\begin{subfigure}[b]{.7\linewidth}\centering
  \includegraphics[height=1.5cm]{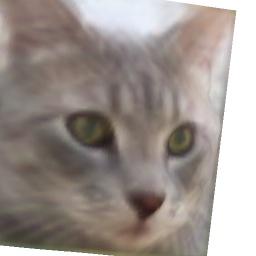}
  \includegraphics[height=1.5cm]{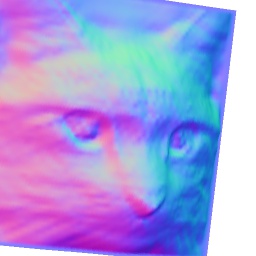}
  \includegraphics[trim={0 0 40px 0}, clip, height=1.5cm]{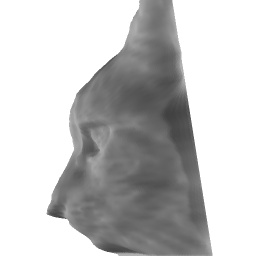}
  \includegraphics[trim={24px 0 0 0}, clip, height=1.5cm]{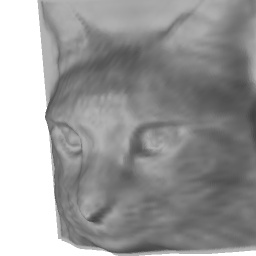}
  \includegraphics[height=1.5cm]{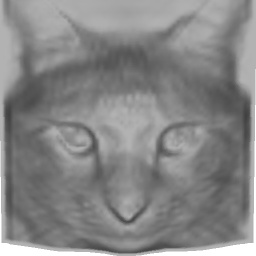}
  \includegraphics[height=1.5cm]{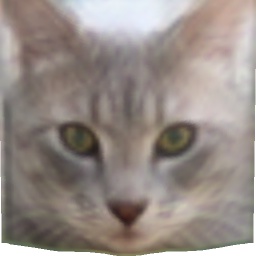}
  \includegraphics[trim={0 0 24px 0}, clip, height=1.5cm]{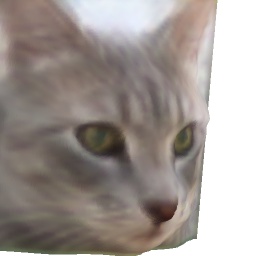}
  \includegraphics[trim={40px 0 0 0}, clip, height=1.5cm]{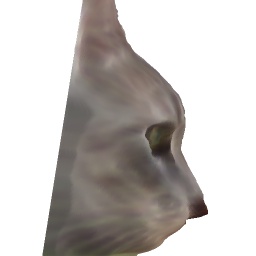}
\end{subfigure}
\\
\begin{subfigure}[b]{.2\linewidth}\centering
  \includegraphics[height=1.5cm]{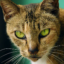}
\end{subfigure}
\begin{subfigure}[b]{.7\linewidth}\centering
  \includegraphics[height=1.5cm]{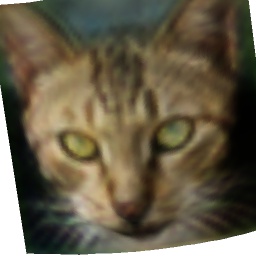}
  \includegraphics[height=1.5cm]{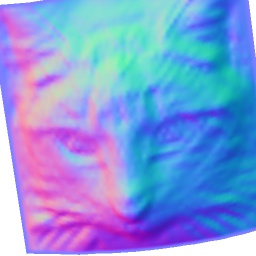}
  \includegraphics[trim={0 0 40px 0}, clip, height=1.5cm]{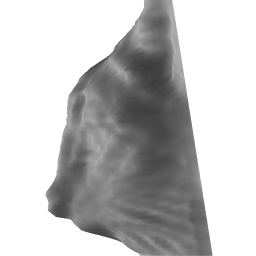}
  \includegraphics[trim={24px 0 0 0}, clip, height=1.5cm]{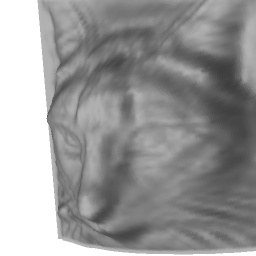}
  \includegraphics[height=1.5cm]{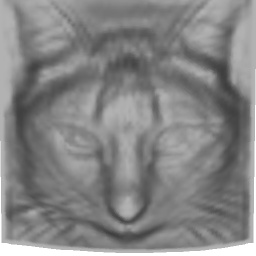}
  \includegraphics[height=1.5cm]{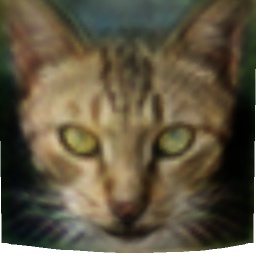}
  \includegraphics[trim={0 0 24px 0}, clip, height=1.5cm]{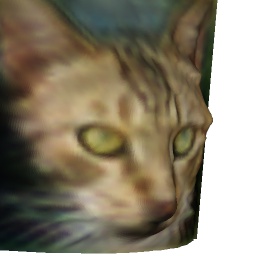}
  \includegraphics[trim={40px 0 0 0}, clip, height=1.5cm]{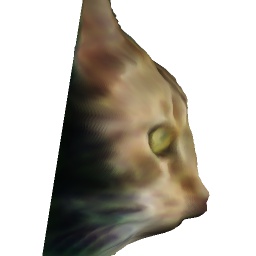}
\end{subfigure}
\\
\begin{subfigure}[b]{.2\linewidth}\centering
  \includegraphics[height=1.5cm]{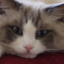}
\end{subfigure}
\begin{subfigure}[b]{.7\linewidth}\centering
  \includegraphics[height=1.5cm]{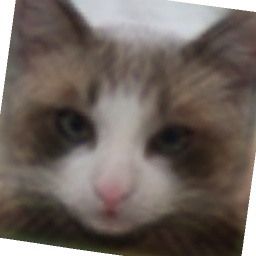}
  \includegraphics[height=1.5cm]{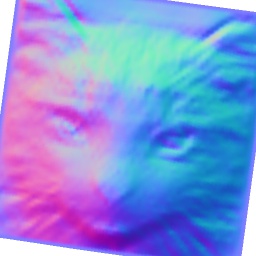}
  \includegraphics[trim={0 0 40px 0}, clip, height=1.5cm]{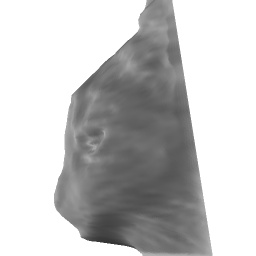}
  \includegraphics[trim={24px 0 0 0}, clip, height=1.5cm]{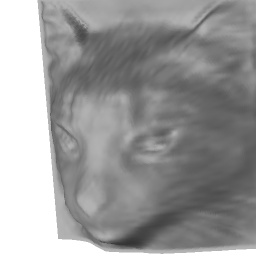}
  \includegraphics[height=1.5cm]{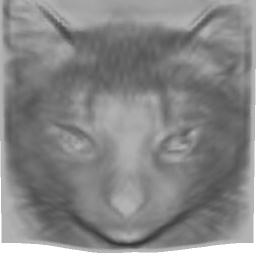}
  \includegraphics[height=1.5cm]{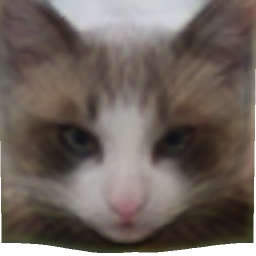}
  \includegraphics[trim={0 0 24px 0}, clip, height=1.5cm]{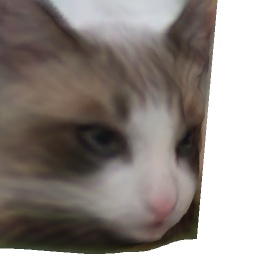}
  \includegraphics[trim={40px 0 0 0}, clip, height=1.5cm]{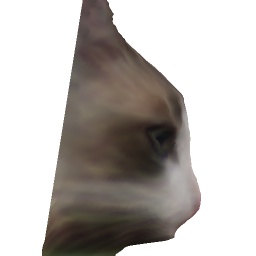}
\end{subfigure}
\\
\begin{subfigure}[b]{.2\linewidth}\centering
  \includegraphics[height=1.5cm]{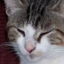}
\end{subfigure}
\begin{subfigure}[b]{.7\linewidth}\centering
  \includegraphics[height=1.5cm]{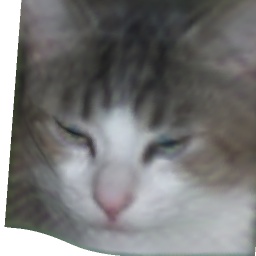}
  \includegraphics[height=1.5cm]{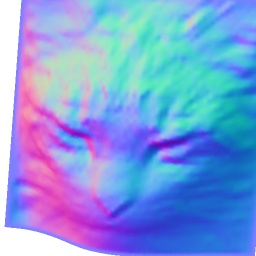}
  \includegraphics[trim={0 0 40px 0}, clip, height=1.5cm]{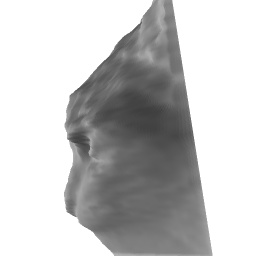}
  \includegraphics[trim={24px 0 0 0}, clip, height=1.5cm]{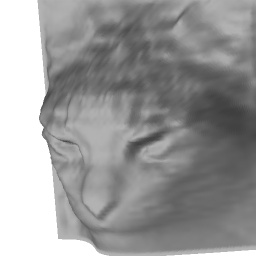}
  \includegraphics[height=1.5cm]{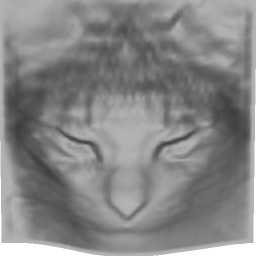}
  \includegraphics[height=1.5cm]{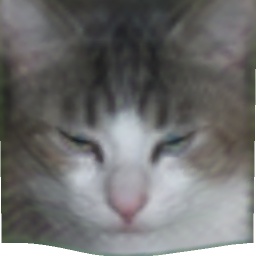}
  \includegraphics[trim={0 0 24px 0}, clip, height=1.5cm]{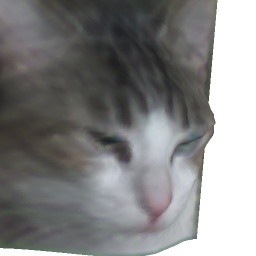}
  \includegraphics[trim={40px 0 0 0}, clip, height=1.5cm]{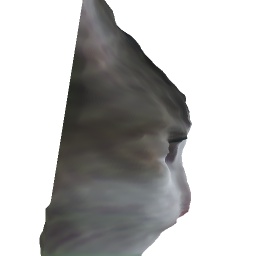}
\end{subfigure}
\\
\begin{subfigure}[b]{.2\linewidth}\centering
  \includegraphics[height=1.5cm]{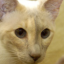}
\end{subfigure}
\begin{subfigure}[b]{.7\linewidth}\centering
  \includegraphics[height=1.5cm]{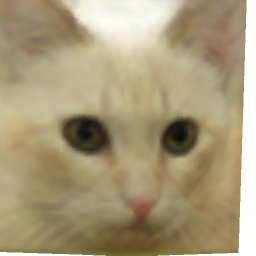}
  \includegraphics[height=1.5cm]{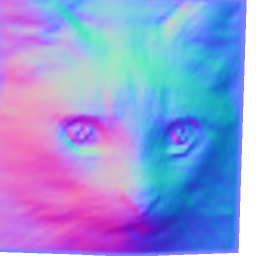}
  \includegraphics[trim={0 0 40px 0}, clip, height=1.5cm]{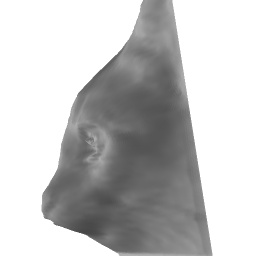}
  \includegraphics[trim={24px 0 0 0}, clip, height=1.5cm]{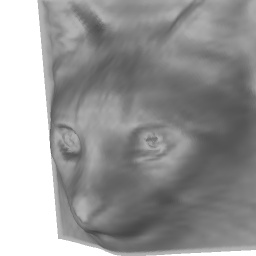}
  \includegraphics[height=1.5cm]{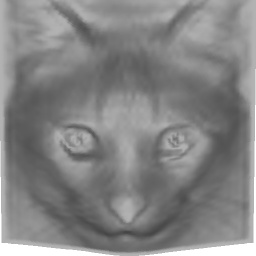}
  \includegraphics[height=1.5cm]{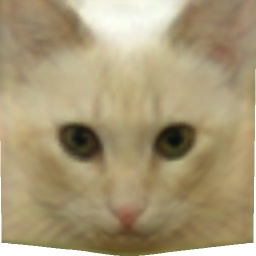}
  \includegraphics[trim={0 0 24px 0}, clip, height=1.5cm]{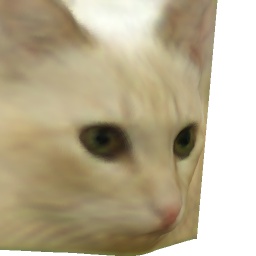}
  \includegraphics[trim={40px 0 0 0}, clip, height=1.5cm]{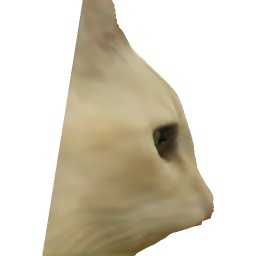}
\end{subfigure}
\\
\begin{subfigure}[b]{.2\linewidth}\centering
  \includegraphics[height=1.5cm]{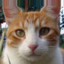}
\end{subfigure}
\begin{subfigure}[b]{.7\linewidth}\centering
  \includegraphics[height=1.5cm]{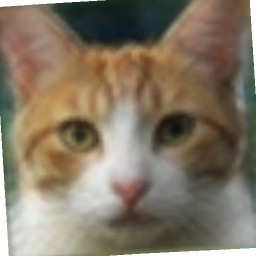}
  \includegraphics[height=1.5cm]{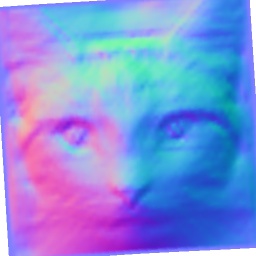}
  \includegraphics[trim={0 0 40px 0}, clip, height=1.5cm]{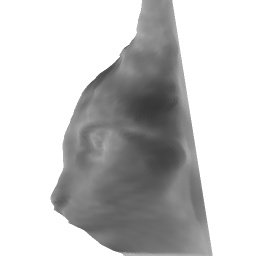}
  \includegraphics[trim={24px 0 0 0}, clip, height=1.5cm]{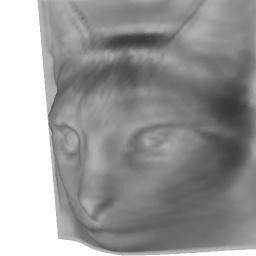}
  \includegraphics[height=1.5cm]{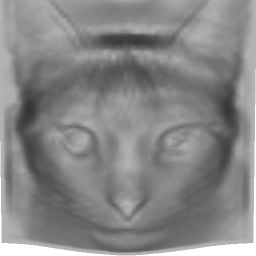}
  \includegraphics[height=1.5cm]{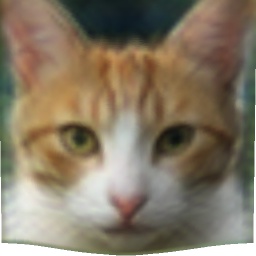}
  \includegraphics[trim={0 0 24px 0}, clip, height=1.5cm]{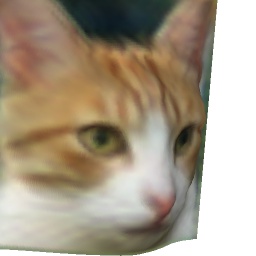}
  \includegraphics[trim={40px 0 0 0}, clip, height=1.5cm]{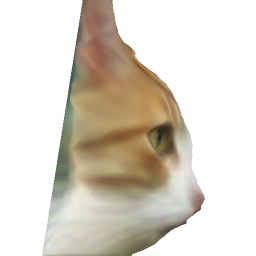}
\end{subfigure}
\\
\begin{subfigure}[b]{.2\linewidth}\centering
  \includegraphics[height=1.5cm]{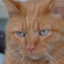}
\end{subfigure}
\begin{subfigure}[b]{.7\linewidth}\centering
  \includegraphics[height=1.5cm]{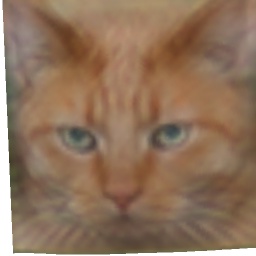}
  \includegraphics[height=1.5cm]{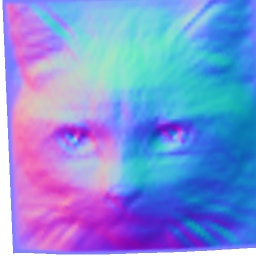}
  \includegraphics[trim={0 0 40px 0}, clip, height=1.5cm]{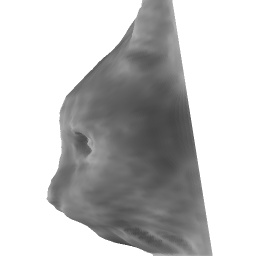}
  \includegraphics[trim={24px 0 0 0}, clip, height=1.5cm]{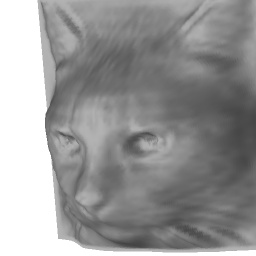}
  \includegraphics[height=1.5cm]{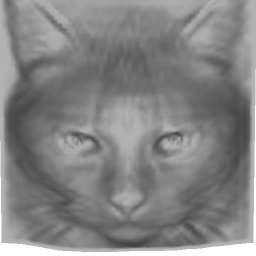}
  \includegraphics[height=1.5cm]{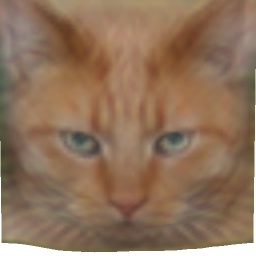}
  \includegraphics[trim={0 0 24px 0}, clip, height=1.5cm]{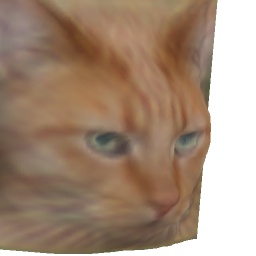}
  \includegraphics[trim={40px 0 0 0}, clip, height=1.5cm]{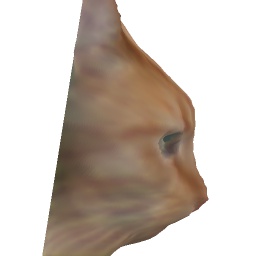}
\end{subfigure}
\\
\begin{subfigure}[b]{.2\linewidth}\centering
  \includegraphics[height=1.5cm]{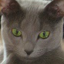}
\end{subfigure}
\begin{subfigure}[b]{.7\linewidth}\centering
  \includegraphics[height=1.5cm]{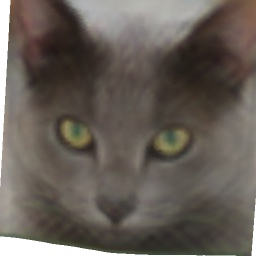}
  \includegraphics[height=1.5cm]{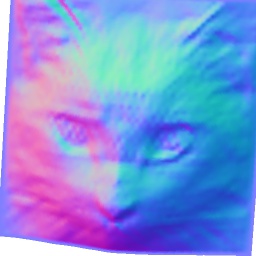}
  \includegraphics[trim={0 0 40px 0}, clip, height=1.5cm]{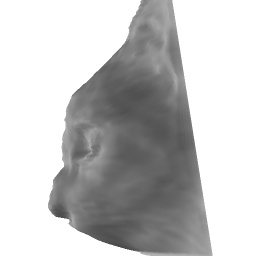}
  \includegraphics[trim={24px 0 0 0}, clip, height=1.5cm]{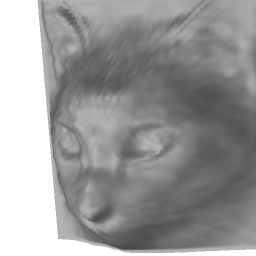}
  \includegraphics[height=1.5cm]{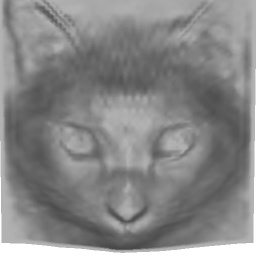}
  \includegraphics[height=1.5cm]{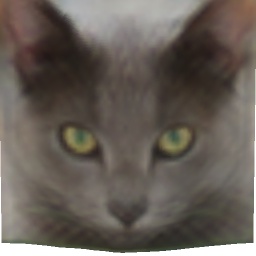}
  \includegraphics[trim={0 0 24px 0}, clip, height=1.5cm]{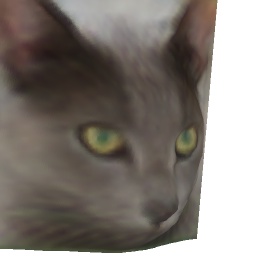}
  \includegraphics[trim={40px 0 0 0}, clip, height=1.5cm]{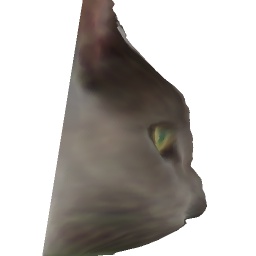}
\end{subfigure}
\\
\begin{subfigure}[b]{.2\linewidth}\centering
  \includegraphics[height=1.5cm]{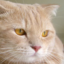}
\end{subfigure}
\begin{subfigure}[b]{.7\linewidth}\centering
  \includegraphics[height=1.5cm]{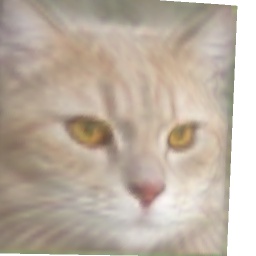}
  \includegraphics[height=1.5cm]{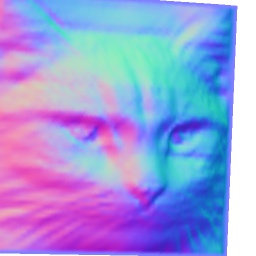}
  \includegraphics[trim={0 0 40px 0}, clip, height=1.5cm]{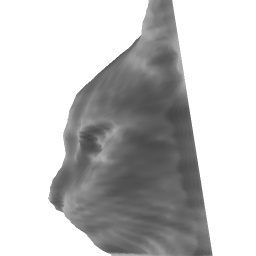}
  \includegraphics[trim={24px 0 0 0}, clip, height=1.5cm]{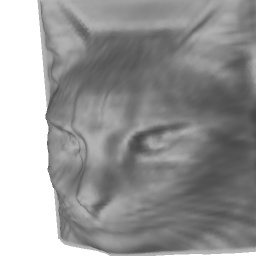}
  \includegraphics[height=1.5cm]{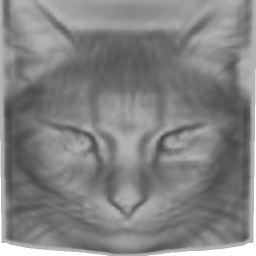}
  \includegraphics[height=1.5cm]{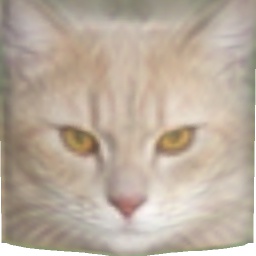}
  \includegraphics[trim={0 0 24px 0}, clip, height=1.5cm]{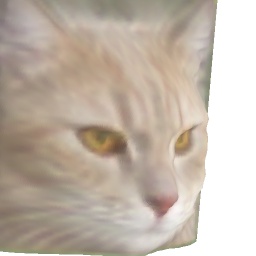}
  \includegraphics[trim={40px 0 0 0}, clip, height=1.5cm]{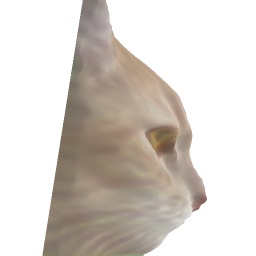}
\end{subfigure}
\\
\begin{subfigure}[b]{.2\linewidth}\centering
  \includegraphics[height=1.5cm]{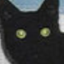}
\end{subfigure}
\begin{subfigure}[b]{.7\linewidth}\centering
  \includegraphics[height=1.5cm]{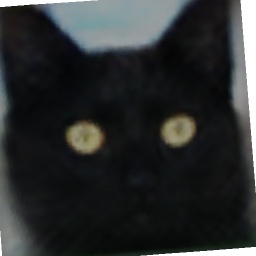}
  \includegraphics[height=1.5cm]{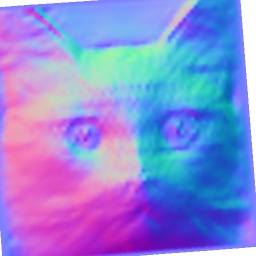}
  \includegraphics[trim={0 0 40px 0}, clip, height=1.5cm]{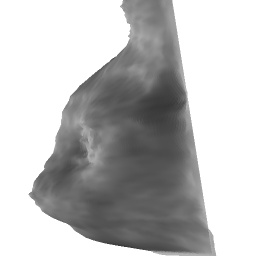}
  \includegraphics[trim={24px 0 0 0}, clip, height=1.5cm]{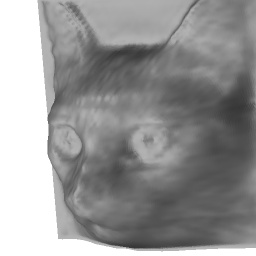}
  \includegraphics[height=1.5cm]{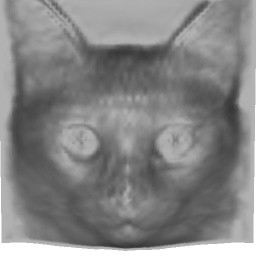}
  \includegraphics[height=1.5cm]{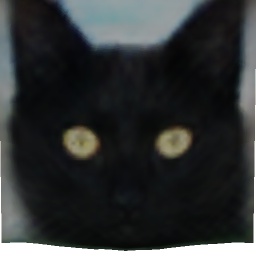}
  \includegraphics[trim={0 0 24px 0}, clip, height=1.5cm]{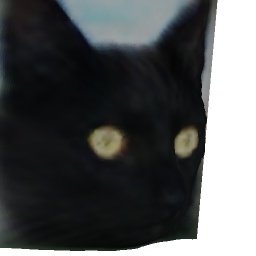}
  \includegraphics[trim={40px 0 0 0}, clip, height=1.5cm]{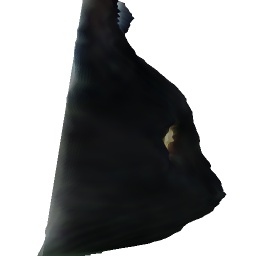}
\end{subfigure}
\\
\begin{subfigure}[b]{.2\linewidth}\centering
  \includegraphics[height=1.5cm]{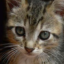}
\end{subfigure}
\begin{subfigure}[b]{.7\linewidth}\centering
  \includegraphics[height=1.5cm]{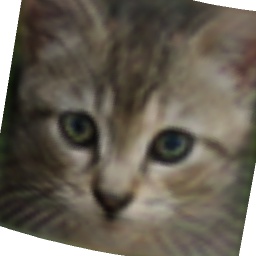}
  \includegraphics[height=1.5cm]{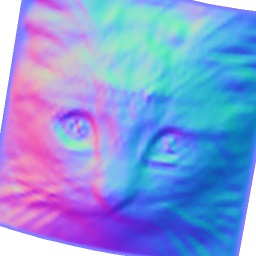}
  \includegraphics[trim={0 0 40px 0}, clip, height=1.5cm]{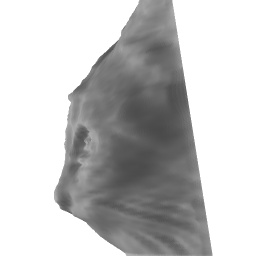}
  \includegraphics[trim={24px 0 0 0}, clip, height=1.5cm]{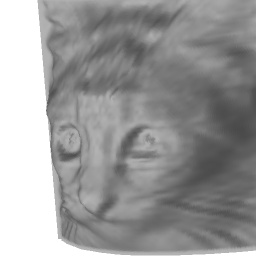}
  \includegraphics[height=1.5cm]{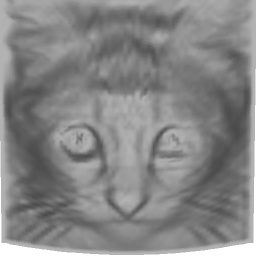}
  \includegraphics[height=1.5cm]{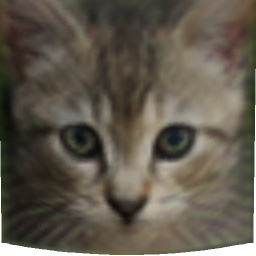}
  \includegraphics[trim={0 0 24px 0}, clip, height=1.5cm]{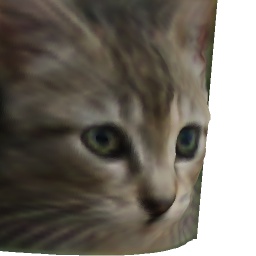}
  \includegraphics[trim={40px 0 0 0}, clip, height=1.5cm]{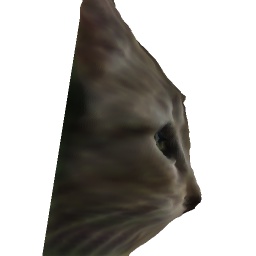}
\end{subfigure}
\\
\begin{subfigure}[b]{.2\linewidth}\centering
  \includegraphics[height=1.5cm]{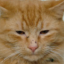}
\end{subfigure}
\begin{subfigure}[b]{.7\linewidth}\centering
  \includegraphics[height=1.5cm]{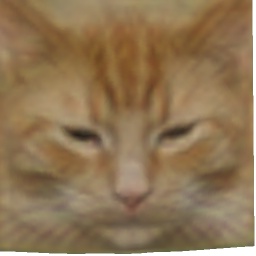}
  \includegraphics[height=1.5cm]{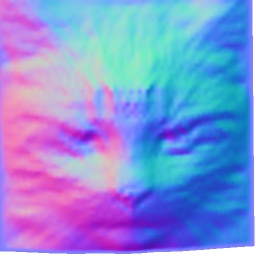}
  \includegraphics[trim={0 0 40px 0}, clip, height=1.5cm]{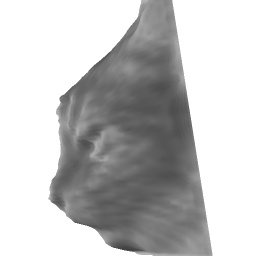}
  \includegraphics[trim={24px 0 0 0}, clip, height=1.5cm]{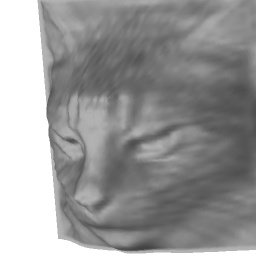}
  \includegraphics[height=1.5cm]{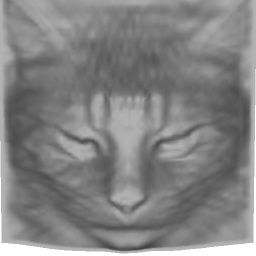}
  \includegraphics[height=1.5cm]{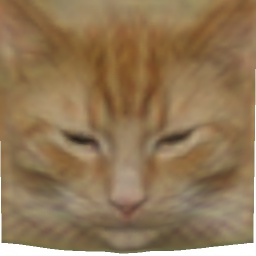}
  \includegraphics[trim={0 0 24px 0}, clip, height=1.5cm]{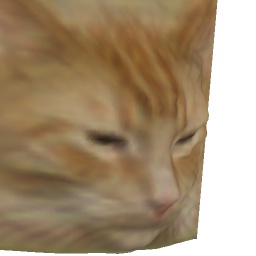}
  \includegraphics[trim={40px 0 0 0}, clip, height=1.5cm]{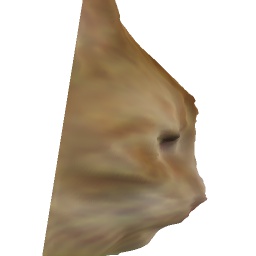}
\end{subfigure}
\\
\begin{subfigure}[b]{.2\linewidth}\centering
  \includegraphics[height=1.5cm]{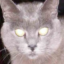}
  \caption{Input}
\end{subfigure}
\begin{subfigure}[b]{.7\linewidth}\centering
  \includegraphics[height=1.5cm]{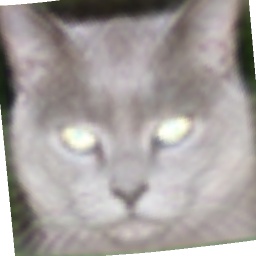}
  \includegraphics[height=1.5cm]{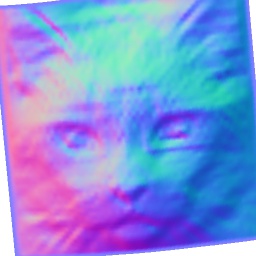}
  \includegraphics[trim={0 0 40px 0}, clip, height=1.5cm]{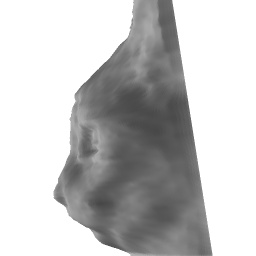}
  \includegraphics[trim={24px 0 0 0}, clip, height=1.5cm]{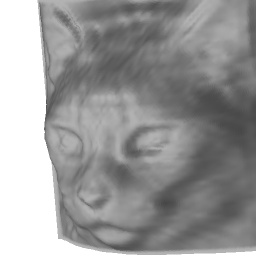}
  \includegraphics[height=1.5cm]{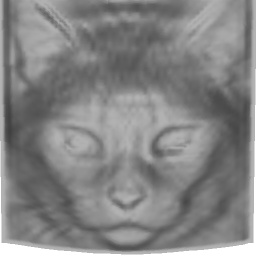}
  \includegraphics[height=1.5cm]{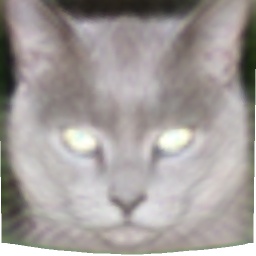}
  \includegraphics[trim={0 0 24px 0}, clip, height=1.5cm]{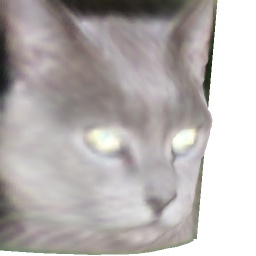}
  \includegraphics[trim={40px 0 0 0}, clip, height=1.5cm]{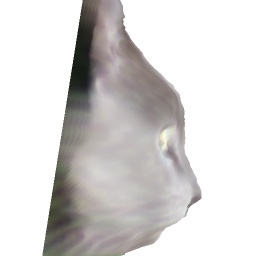}
  \caption{Reconstruction}
\end{subfigure}

\caption{\textbf{Reconstruction of cat faces.}}\label{fig:sup-cat}
\end{figure*}

%% file: supmat/fig-abstract-cat.tex
\begin{figure*}[t]\centering
\captionsetup[subfigure]{justification=centering,labelformat=empty,labelsep=colon,aboveskip=2pt}
\begin{subfigure}[b]{.2\linewidth}\centering
  \includegraphics[height=1.5cm]{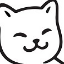}
\end{subfigure}
\begin{subfigure}[b]{.7\linewidth}\centering
  \includegraphics[height=1.5cm]{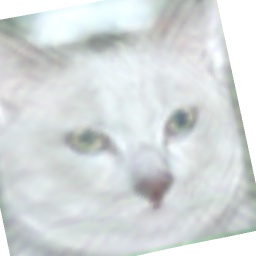}
  \includegraphics[height=1.5cm]{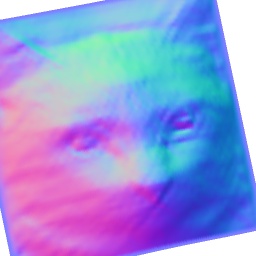}
  \includegraphics[trim={0 0 40px 0}, clip, height=1.5cm]{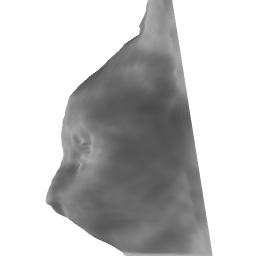}
  \includegraphics[trim={24px 0 0 0}, clip, height=1.5cm]{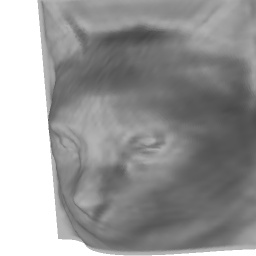}
  \includegraphics[height=1.5cm]{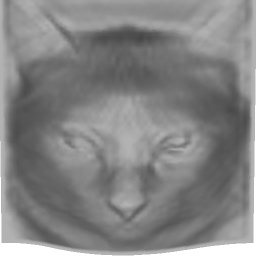}
  \includegraphics[height=1.5cm]{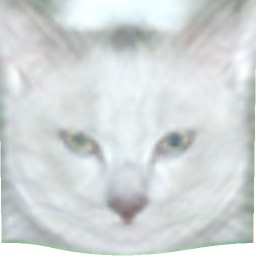}
  \includegraphics[trim={0 0 24px 0}, clip, height=1.5cm]{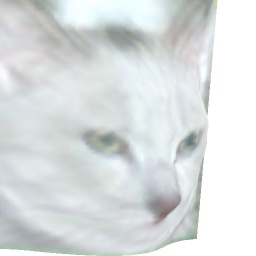}
  \includegraphics[trim={40px 0 0 0}, clip, height=1.5cm]{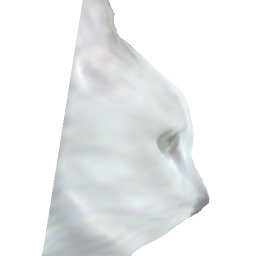}
\end{subfigure}
\\
\begin{subfigure}[b]{.2\linewidth}\centering
  \includegraphics[height=1.5cm]{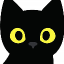}
\end{subfigure}
\begin{subfigure}[b]{.7\linewidth}\centering
  \includegraphics[height=1.5cm]{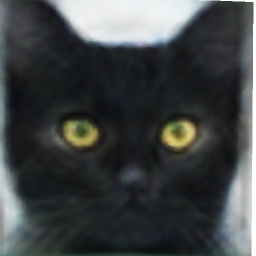}
  \includegraphics[height=1.5cm]{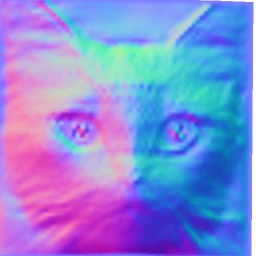}
  \includegraphics[trim={0 0 40px 0}, clip, height=1.5cm]{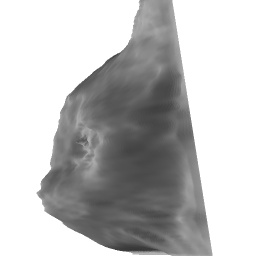}
  \includegraphics[trim={24px 0 0 0}, clip, height=1.5cm]{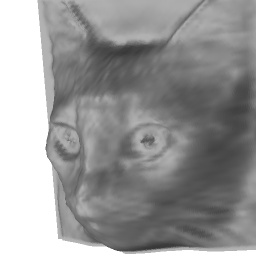}
  \includegraphics[height=1.5cm]{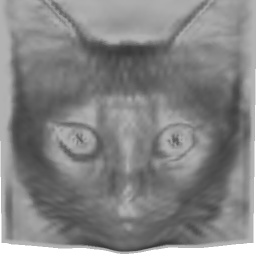}
  \includegraphics[height=1.5cm]{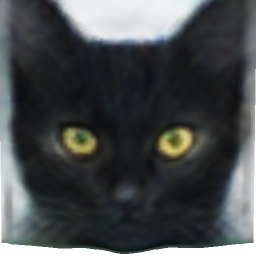}
  \includegraphics[trim={0 0 24px 0}, clip, height=1.5cm]{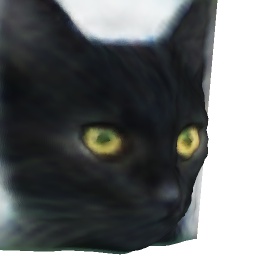}
  \includegraphics[trim={40px 0 0 0}, clip, height=1.5cm]{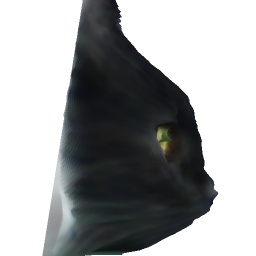}
\end{subfigure}
\\
\begin{subfigure}[b]{.2\linewidth}\centering
  \includegraphics[height=1.5cm]{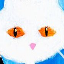}
\end{subfigure}
\begin{subfigure}[b]{.7\linewidth}\centering
  \includegraphics[height=1.5cm]{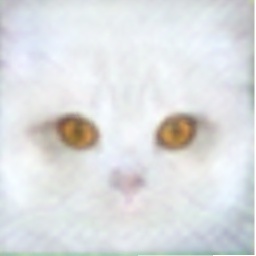}
  \includegraphics[height=1.5cm]{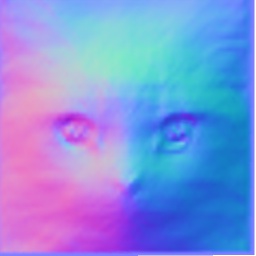}
  \includegraphics[trim={0 0 40px 0}, clip, height=1.5cm]{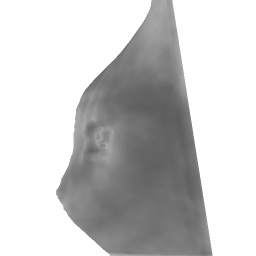}
  \includegraphics[trim={24px 0 0 0}, clip, height=1.5cm]{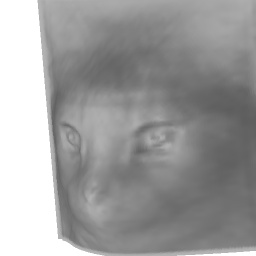}
  \includegraphics[height=1.5cm]{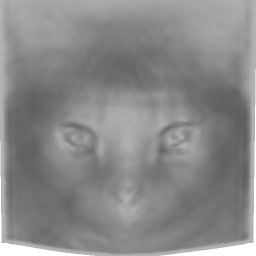}
  \includegraphics[height=1.5cm]{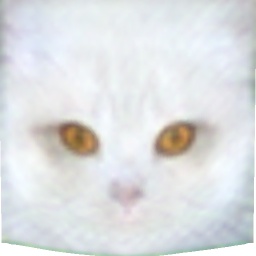}
  \includegraphics[trim={0 0 24px 0}, clip, height=1.5cm]{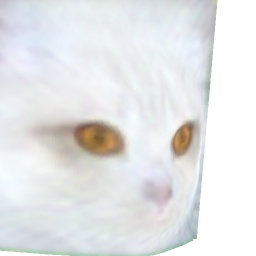}
  \includegraphics[trim={40px 0 0 0}, clip, height=1.5cm]{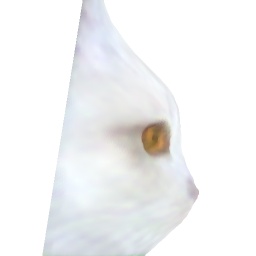}
\end{subfigure}
\\
\begin{subfigure}[b]{.2\linewidth}\centering
  \includegraphics[height=1.5cm]{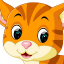}
\end{subfigure}
\begin{subfigure}[b]{.7\linewidth}\centering
  \includegraphics[height=1.5cm]{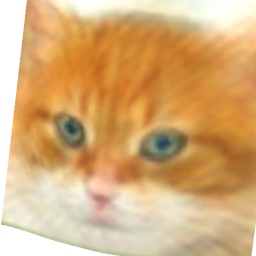}
  \includegraphics[height=1.5cm]{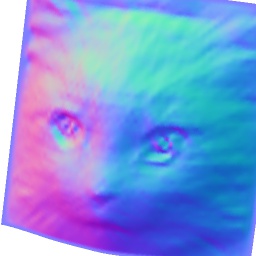}
  \includegraphics[trim={0 0 40px 0}, clip, height=1.5cm]{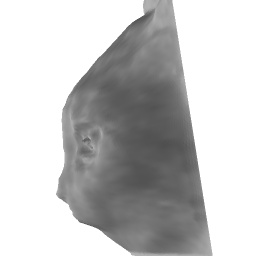}
  \includegraphics[trim={24px 0 0 0}, clip, height=1.5cm]{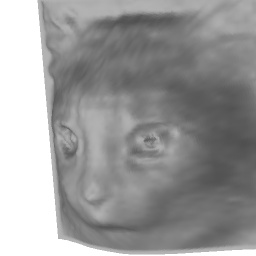}
  \includegraphics[height=1.5cm]{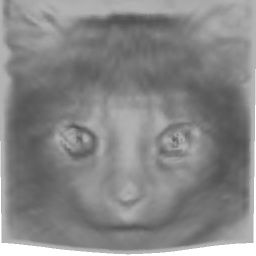}
  \includegraphics[height=1.5cm]{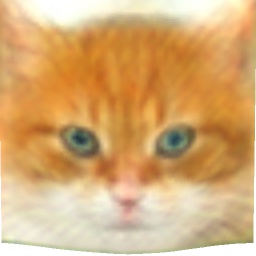}
  \includegraphics[trim={0 0 24px 0}, clip, height=1.5cm]{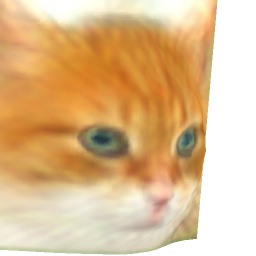}
  \includegraphics[trim={40px 0 0 0}, clip, height=1.5cm]{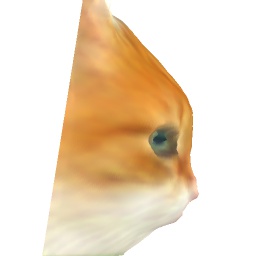}
\end{subfigure}
\\
\begin{subfigure}[b]{.2\linewidth}\centering
  \includegraphics[height=1.5cm]{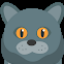}
\end{subfigure}
\begin{subfigure}[b]{.7\linewidth}\centering
  \includegraphics[height=1.5cm]{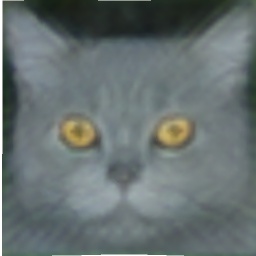}
  \includegraphics[height=1.5cm]{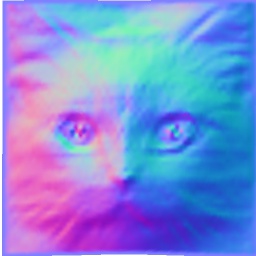}
  \includegraphics[trim={0 0 40px 0}, clip, height=1.5cm]{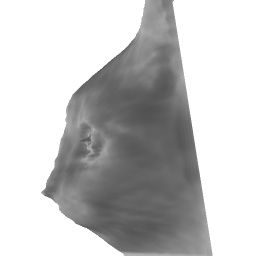}
  \includegraphics[trim={24px 0 0 0}, clip, height=1.5cm]{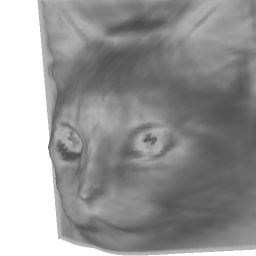}
  \includegraphics[height=1.5cm]{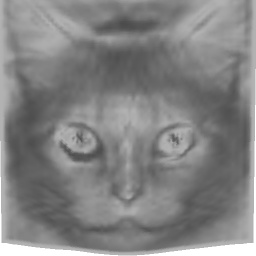}
  \includegraphics[height=1.5cm]{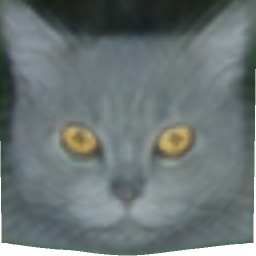}
  \includegraphics[trim={0 0 24px 0}, clip, height=1.5cm]{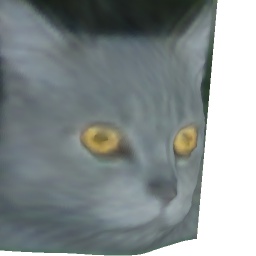}
  \includegraphics[trim={40px 0 0 0}, clip, height=1.5cm]{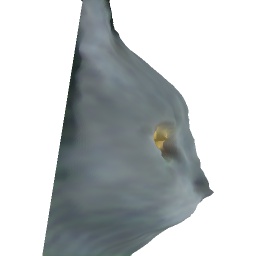}
\end{subfigure}
\\
\begin{subfigure}[b]{.2\linewidth}\centering
  \includegraphics[height=1.5cm]{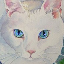}
\end{subfigure}
\begin{subfigure}[b]{.7\linewidth}\centering
  \includegraphics[height=1.5cm]{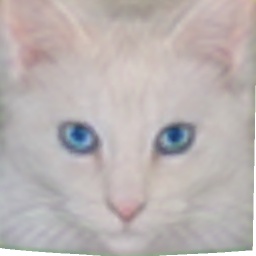}
  \includegraphics[height=1.5cm]{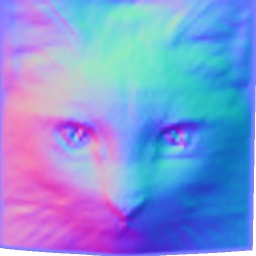}
  \includegraphics[trim={0 0 40px 0}, clip, height=1.5cm]{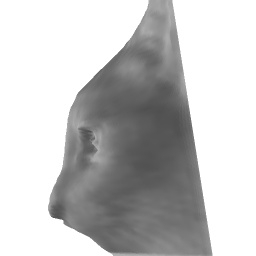}
  \includegraphics[trim={24px 0 0 0}, clip, height=1.5cm]{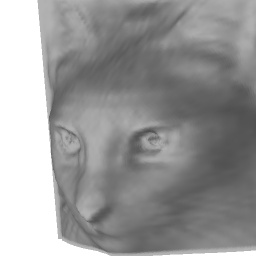}
  \includegraphics[height=1.5cm]{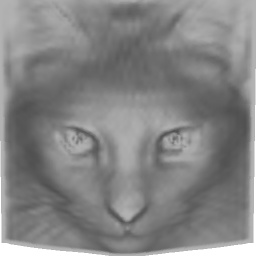}
  \includegraphics[height=1.5cm]{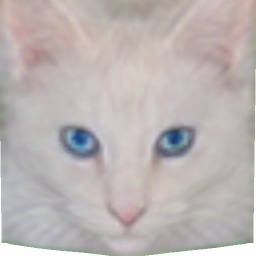}
  \includegraphics[trim={0 0 24px 0}, clip, height=1.5cm]{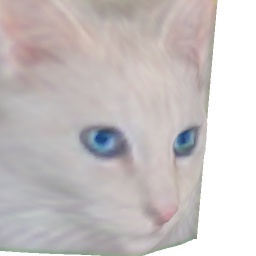}
  \includegraphics[trim={40px 0 0 0}, clip, height=1.5cm]{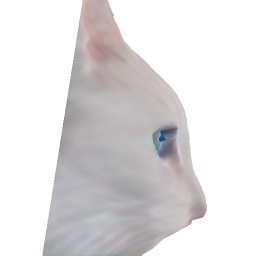}
\end{subfigure}
\\
\begin{subfigure}[b]{.2\linewidth}\centering
  \includegraphics[height=1.5cm]{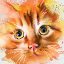}
  \caption{Input}
\end{subfigure}
\begin{subfigure}[b]{.7\linewidth}\centering
  \includegraphics[height=1.5cm]{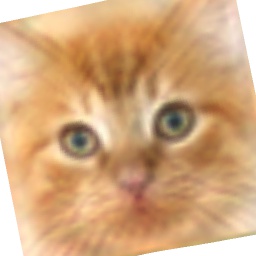}
  \includegraphics[height=1.5cm]{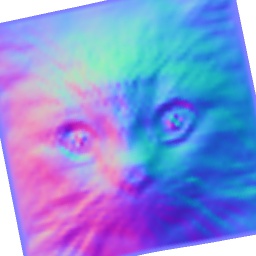}
  \includegraphics[trim={0 0 40px 0}, clip, height=1.5cm]{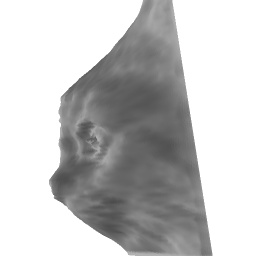}
  \includegraphics[trim={24px 0 0 0}, clip, height=1.5cm]{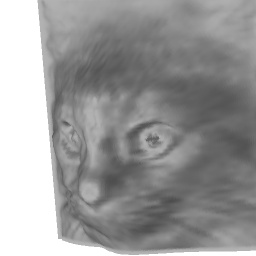}
  \includegraphics[height=1.5cm]{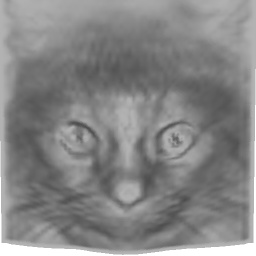}
  \includegraphics[height=1.5cm]{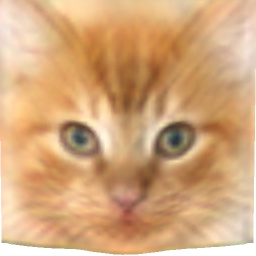}
  \includegraphics[trim={0 0 24px 0}, clip, height=1.5cm]{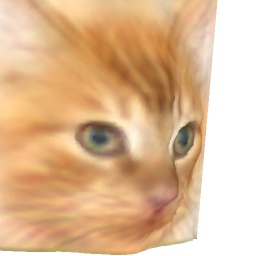}
  \includegraphics[trim={40px 0 0 0}, clip, height=1.5cm]{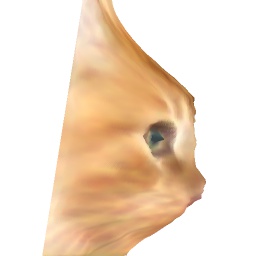}
  \caption{Reconstruction}
\end{subfigure}

\caption{\textbf{Reconstruction of abstract cats.}}\label{fig:sup-abstract-cat}
\end{figure*}

%% file: supmat/fig-car.tex
\begin{figure*}[t]\centering
\captionsetup[subfigure]{justification=centering,labelformat=empty,labelsep=colon,aboveskip=2pt}
\begin{subfigure}[b]{.2\linewidth}\centering
  \includegraphics[height=1.5cm]{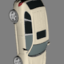}
\end{subfigure}
\begin{subfigure}[b]{.7\linewidth}\centering
  \includegraphics[height=1.5cm]{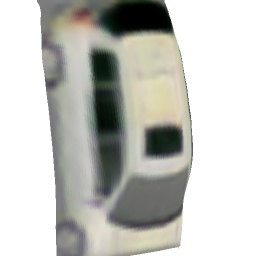}
  \includegraphics[height=1.5cm]{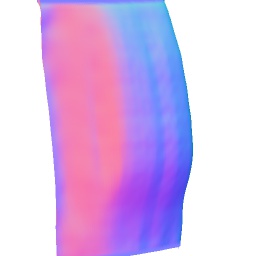}
  \includegraphics[trim={0 0 40px 0}, clip, height=1.5cm]{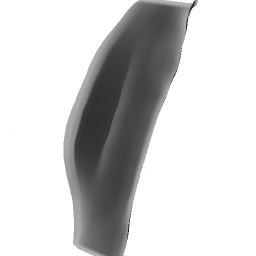}
  \includegraphics[trim={24px 0 0 0}, clip, height=1.5cm]{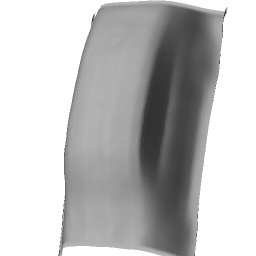}
  \includegraphics[height=1.5cm]{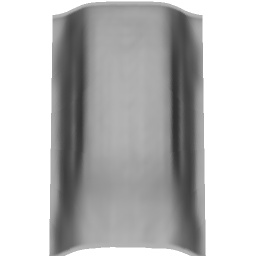}
  \includegraphics[height=1.5cm]{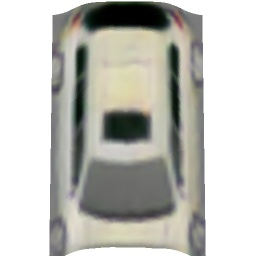}
  \includegraphics[trim={0 0 24px 0}, clip, height=1.5cm]{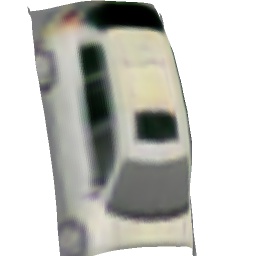}
  \includegraphics[trim={40px 0 0 0}, clip, height=1.5cm]{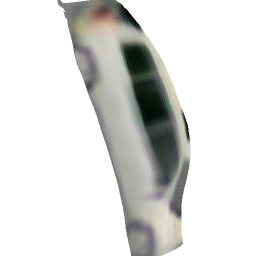}
\end{subfigure}
\\
\begin{subfigure}[b]{.2\linewidth}\centering
  \includegraphics[height=1.5cm]{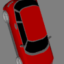}
\end{subfigure}
\begin{subfigure}[b]{.7\linewidth}\centering
  \includegraphics[height=1.5cm]{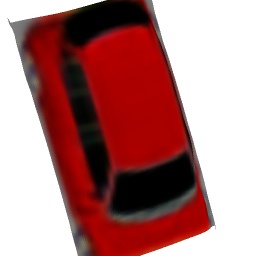}
  \includegraphics[height=1.5cm]{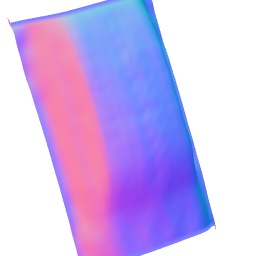}
  \includegraphics[trim={0 0 40px 0}, clip, height=1.5cm]{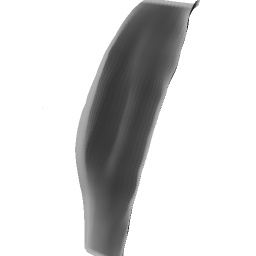}
  \includegraphics[trim={24px 0 0 0}, clip, height=1.5cm]{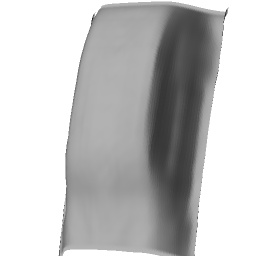}
  \includegraphics[height=1.5cm]{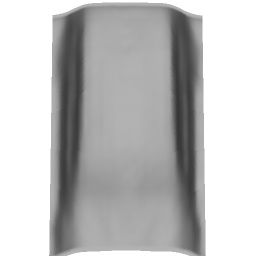}
  \includegraphics[height=1.5cm]{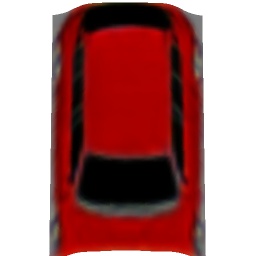}
  \includegraphics[trim={0 0 24px 0}, clip, height=1.5cm]{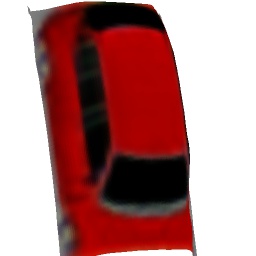}
  \includegraphics[trim={40px 0 0 0}, clip, height=1.5cm]{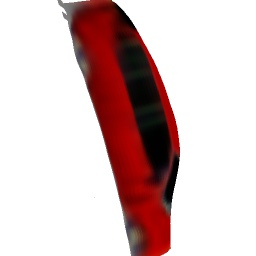}
\end{subfigure}
\\
\begin{subfigure}[b]{.2\linewidth}\centering
  \includegraphics[height=1.5cm]{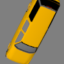}
\end{subfigure}
\begin{subfigure}[b]{.7\linewidth}\centering
  \includegraphics[height=1.5cm]{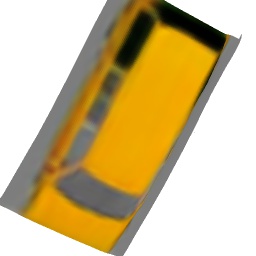}
  \includegraphics[height=1.5cm]{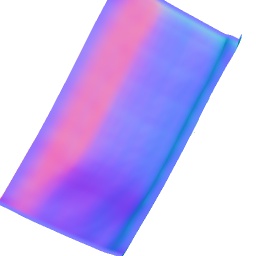}
  \includegraphics[trim={0 0 40px 0}, clip, height=1.5cm]{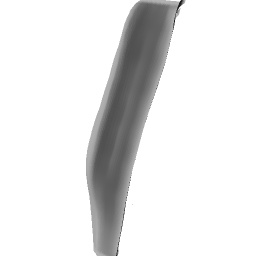}
  \includegraphics[trim={24px 0 0 0}, clip, height=1.5cm]{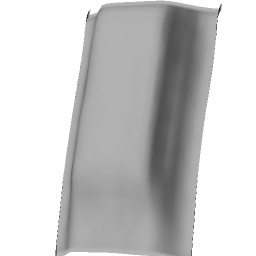}
  \includegraphics[height=1.5cm]{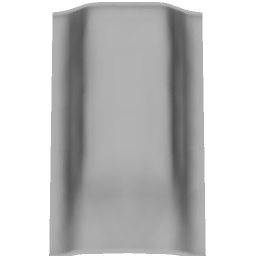}
  \includegraphics[height=1.5cm]{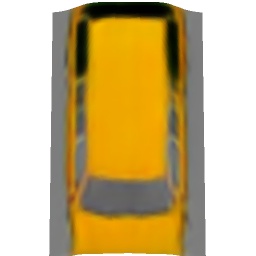}
  \includegraphics[trim={0 0 24px 0}, clip, height=1.5cm]{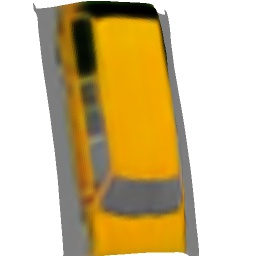}
  \includegraphics[trim={40px 0 0 0}, clip, height=1.5cm]{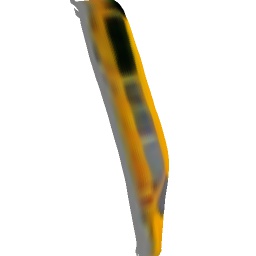}
\end{subfigure}
\\
\begin{subfigure}[b]{.2\linewidth}\centering
  \includegraphics[height=1.5cm]{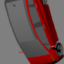}
\end{subfigure}
\begin{subfigure}[b]{.7\linewidth}\centering
  \includegraphics[height=1.5cm]{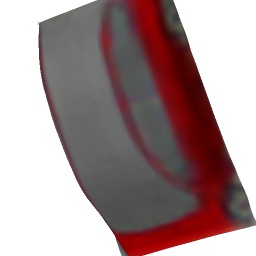}
  \includegraphics[height=1.5cm]{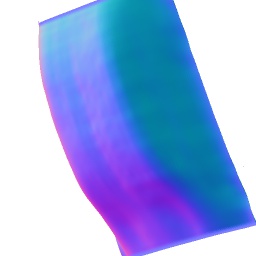}
  \includegraphics[trim={0 0 40px 0}, clip, height=1.5cm]{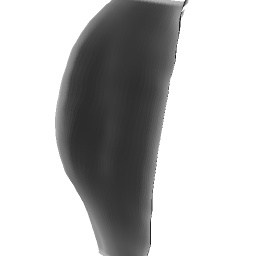}
  \includegraphics[trim={24px 0 0 0}, clip, height=1.5cm]{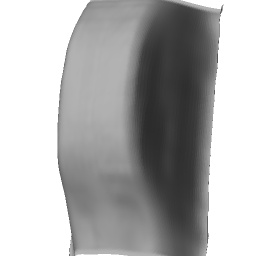}
  \includegraphics[height=1.5cm]{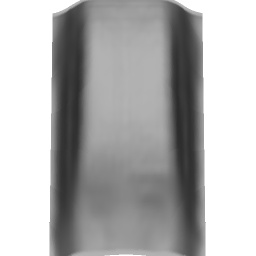}
  \includegraphics[height=1.5cm]{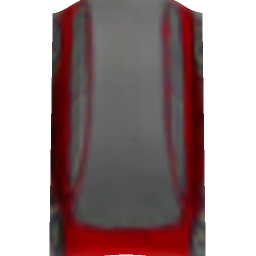}
  \includegraphics[trim={0 0 24px 0}, clip, height=1.5cm]{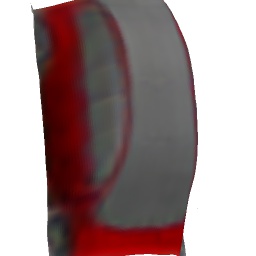}
  \includegraphics[trim={40px 0 0 0}, clip, height=1.5cm]{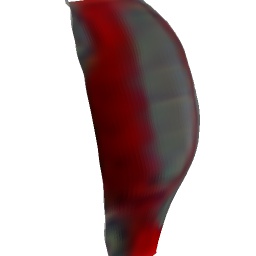}
\end{subfigure}
\\
\begin{subfigure}[b]{.2\linewidth}\centering
  \includegraphics[height=1.5cm]{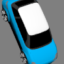}
  \caption{Input}
\end{subfigure}
\begin{subfigure}[b]{.7\linewidth}\centering
  \includegraphics[height=1.5cm]{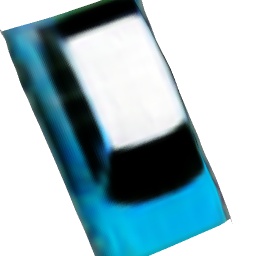}
  \includegraphics[height=1.5cm]{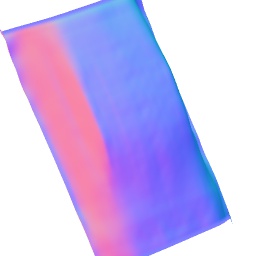}
  \includegraphics[trim={0 0 40px 0}, clip, height=1.5cm]{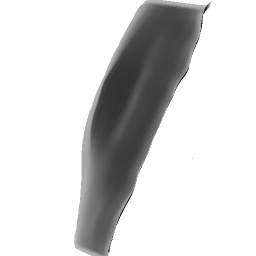}
  \includegraphics[trim={24px 0 0 0}, clip, height=1.5cm]{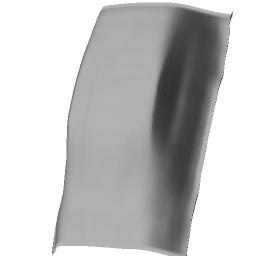}
  \includegraphics[height=1.5cm]{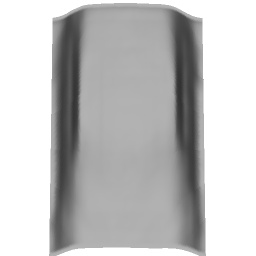}
  \includegraphics[height=1.5cm]{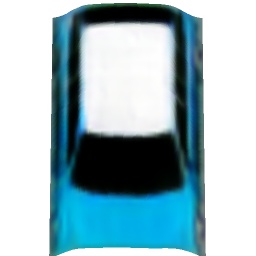}
  \includegraphics[trim={0 0 24px 0}, clip, height=1.5cm]{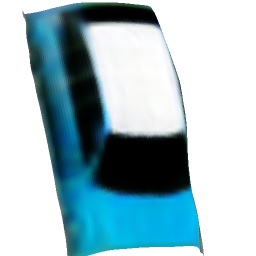}
  \includegraphics[trim={40px 0 0 0}, clip, height=1.5cm]{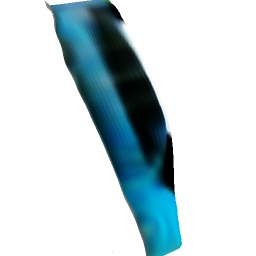}
  \caption{Reconstruction}
\end{subfigure}

\caption{\textbf{Reconstruction of synthetic cars.}}\label{fig:sup-car}
\end{figure*}